\PassOptionsToPackage{table}{xcolor}
\documentclass{article}

\usepackage[T1]{fontenc}
\usepackage{iclr2027_conference,times}

\usepackage{amsmath,amsfonts,bm}

\def\eqref#1{equation~\ref{#1}}

\def\plaineqref#1{\ref{#1}}

\def\1{\bm{1}}

\DeclareMathAlphabet{\mathsfit}{\encodingdefault}{\sfdefault}{m}{sl}
\SetMathAlphabet{\mathsfit}{bold}{\encodingdefault}{\sfdefault}{bx}{n}

\usepackage{amsmath,amssymb,mathtools}
\usepackage{booktabs,array,tabularx,multirow,placeins}
\usepackage{xcolor}
\usepackage{longtable}
\usepackage{float,needspace}
\usepackage{microtype}
\usepackage{graphicx}
\usepackage{hyperref}
\usepackage{xurl}
\definecolor{ReferenceBlue}{HTML}{36566F}
\definecolor{CitationBrown}{HTML}{A96832}
\definecolor{EquationWine}{HTML}{7A263A}
\definecolor{JumpRed}{HTML}{FF0000}
\definecolor{ResourceBlue}{HTML}{0066CC}
\definecolor{HomeLinkBlue}{HTML}{1A0DAB}
\definecolor{TableHeaderTint}{HTML}{EFEFED}
\newcommand{\tableheadrule}{\specialrule{0.35pt}{2pt}{1.1pt}\specialrule{0.35pt}{0pt}{2pt}}
\newcommand{\tableentryrule}{\noalign{\vskip2pt\hbox to\linewidth{\color{black!28}\leaders\hbox{\rule{2pt}{0.25pt}\hskip2pt}\hfill}\vskip2pt}}
\hypersetup{colorlinks=true,linkcolor=JumpRed,citecolor=CitationBrown,
  urlcolor=ResourceBlue,filecolor=ResourceBlue,pdfborder={0 0 0}}
\renewcommand{\plaineqref}[1]{\begingroup\hypersetup{linkcolor=JumpRed}\ref{#1}\endgroup}
\let\originaleqref\eqref
\renewcommand{\eqref}[1]{\begingroup\hypersetup{linkcolor=JumpRed}\originaleqref{#1}\endgroup}

\newcommand{\sprii}{SPRII}

\newcommand{\Lalign}{\mathcal{L}_{\mathrm{align}}}
\newcommand{\Lcross}{\mathcal{L}_{\mathrm{cross}}}
\newcommand{\zs}{z_{\mathrm{s}}}
\newcommand{\zp}{z_{\mathrm{p}}}

\newcommand{\appentry}[2]{%
  \par\addvspace{6pt}\noindent
  {\bfseries\hyperref[#1]{\makebox[2.2em][l]{\ref*{#1}}#2}%
  \hfill\hyperref[#1]{\pageref*{#1}}}\par}
\newcommand{\appsubentry}[2]{%
  \noindent\hspace*{1.2em}%
  \hyperref[#1]{\makebox[2.8em][l]{\ref*{#1}}#2}%
  \nobreak\dotfill\hyperref[#1]{\makebox[1.8em][r]{\pageref*{#1}}}\par}

\title{\upshape Shaping Persistent Representations\\from Independent Interactions}
\hypersetup{pdftitle={Shaping Persistent Representations from Independent Interactions},
  pdfauthor={Ji Dai, Quan Fang, Junyu Gao, Rongfeng Guo, Haoyan Rong, YipingHuang, Yongxi Li},pdfsubject={SPRII: reusable world context, formation, use and conditional value}}

\author{%
\textbf{Ji Dai\textsuperscript{1}\hspace{4pt}%
Quan Fang\textsuperscript{1}\hspace{4pt}%
Junyu Gao\textsuperscript{2}\hspace{4pt}%
Rongfeng Guo\textsuperscript{3}\hspace{4pt}%
Haoyan Rong\textsuperscript{4}\hspace{4pt}%
YipingHuang\textsuperscript{1}\hspace{4pt}%
Yongxi Li\textsuperscript{2}}\\[3pt]
{\normalfont\small\textsuperscript{1}Beijing University of Posts and Telecommunications}\\
{\normalfont\small\textsuperscript{2}Institute of Automation, Chinese Academy of Sciences}\\
{\normalfont\small\textsuperscript{3}Shenzhen University \quad \textsuperscript{4}Tsinghua University}}

\iclrfinalcopy
\makeatletter
\def\@maketitle{\vbox{\hsize\textwidth\centering
  {\LARGE\upshape\@title\par}
  \def\And{\end{tabular}\hfil\linebreak[0]\hfil
    \begin{tabular}[t]{l}\bf\rule{\z@}{24pt}\ignorespaces}
  \def\AND{\end{tabular}\hfil\linebreak[4]\hfil
    \begin{tabular}[t]{l}\bf\rule{\z@}{24pt}\ignorespaces}
  \ifx\@author\@empty\else
    \begin{tabular}[t]{@{}c@{}}\rule{\z@}{24pt}\@author\end{tabular}\par
  \fi
  \vskip 0.3in minus 0.1in}}
\makeatother
\renewcommand{\headrulewidth}{0pt}
\definecolor{MethodTint}{HTML}{EEE5D7}
\definecolor{KeyRowTint}{HTML}{EEE5D7}
\newcommand{\takeaway}[1]{\par\noindent\textbf{Takeaway.} #1\par\smallskip}
\newcommand{\takeawaybox}[1]{\par\addvspace{5pt}\noindent\begingroup
\setlength{\fboxsep}{6pt}\setlength{\fboxrule}{0.4pt}
\fcolorbox{ReferenceBlue!55}{KeyRowTint!35}{\parbox{\dimexpr\linewidth-2\fboxsep-2\fboxrule-2pt\relax}{\small\textbf{Takeaway.} #1}}\endgroup\par\addvspace{6pt}}
\makeatletter
\newcommand{\sdev}[1]{\ensuremath{\,\text{\fontsize{\dimexpr\f@size pt*7/10\relax}{\f@baselineskip}\selectfont$\pm\,#1$}}}
\newcommand{\msd}[2]{\ensuremath{#1\sdev{#2}}}
\makeatother
\newenvironment{papertablenotes}{\par\vspace{3pt}\begin{minipage}{\linewidth}\footnotesize\raggedright\setlength{\parskip}{0pt}}{\end{minipage}\par}
\newcommand{\tablenote}[2]{\par\noindent\hangindent=1em\hangafter=1\makebox[1em][l]{\textsuperscript{#1}}#2\par}
\floatstyle{ruled}
\newfloat{algorithm}{tbp}{loa}
\floatname{algorithm}{Algorithm}
\renewcommand{\topfraction}{0.92}
\renewcommand{\textfraction}{0.08}
\newcommand{\SPRIIfigone}{%
\begin{figure}[!t]
\centering
\includegraphics[width=\linewidth]{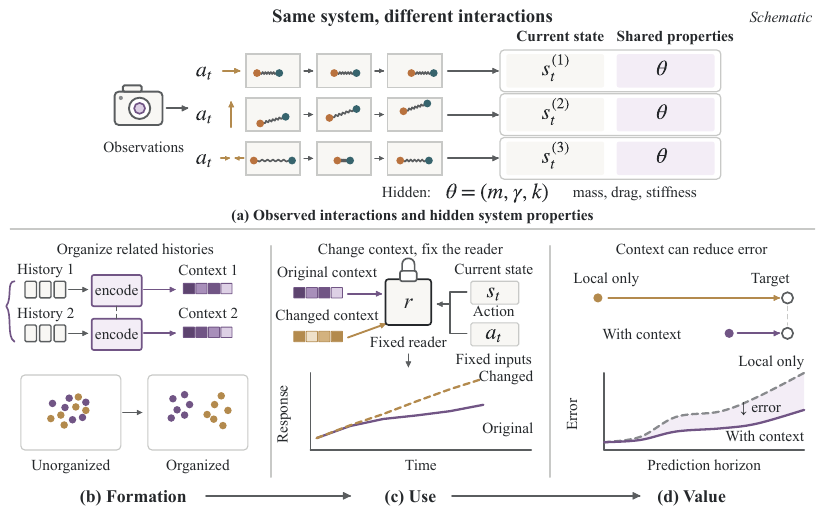}
\caption{\textbf{From related interactions to useful persistent information.}
All panels are schematic. (a) Interactions differ in state $s_t$ and action
$a_t$ while sharing mass, drag and stiffness, $\theta=(m,\gamma,k)$.
(b) Formation organizes persistent information. (c) Use tests its effect
on a fixed reader's predictions. (d) Value tests whether it lowers task error.
The linked questions distinguish information formation, predictive use
and task benefit; success at one stage does not guarantee the next.}
\label{fig:overview}
\end{figure}}

\newcommand{\SPRIIfigtwo}{%
\begin{figure}[!t]
\centering
\includegraphics[width=\linewidth]{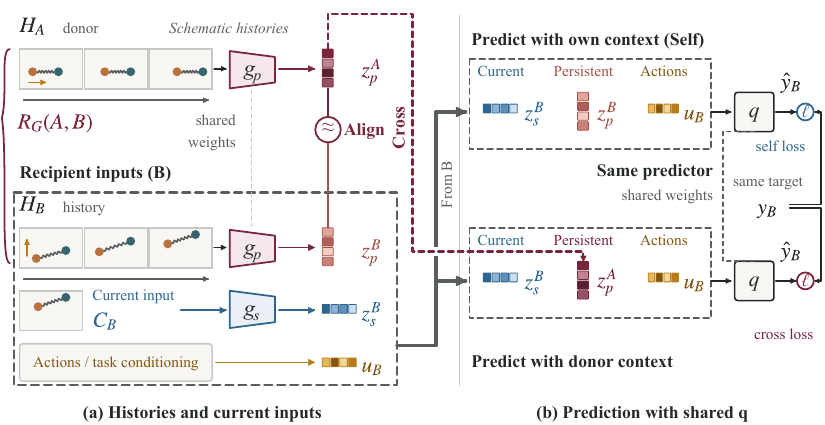}
\caption{\textbf{Relation-supervised learning of persistent context.}
Thick arrows carry recipient inputs (a) into both prediction branches (b).
(a) Align regularizes contexts paired by $\mathcal R_G$.
(b) Both calls share predictor $q$, current input $C_B$ and conditioning $u_B$;
Cross replaces only $z_p^B$ with related donor context $z_p^A$.
$y_B$ is the loss target. Both components augment the native objective.
Trajectories and tokens are schematic; Figure~\ref{fig:controlled-architecture}
shows the JEPA implementation.}
\label{fig:architecture}
\end{figure}}

\newcommand{\SPRIIfigthree}{%
\begin{figure}[!htbp]
\centering
\includegraphics[width=\linewidth]{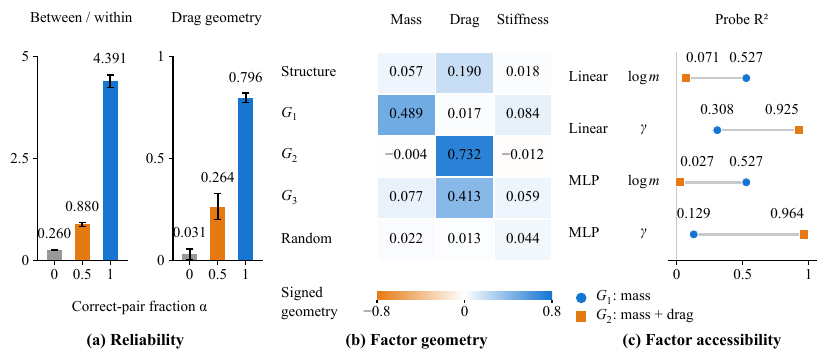}
\caption{\textbf{Relations shape the strength and content of persistent representations.}
(a) Pair reliability strengthens separation and drag geometry; whiskers: source SD.
(b) Partial geometry. (c) Linear/MLP probe $R^2$.
$G_1=m$, $G_2=(m,\gamma)$, $G_3=(m,\gamma,k)$.
Means cover three sources. Geometry uses 400 confirmation systems;
probes use 400 validation systems (Appendix~\ref{app:selectivity}).}
\label{fig:relation-mechanism}
\end{figure}}

\newcommand{\SPRIIfigfour}{%
\begin{figure}[!htbp]
\centering
\includegraphics[width=\linewidth]{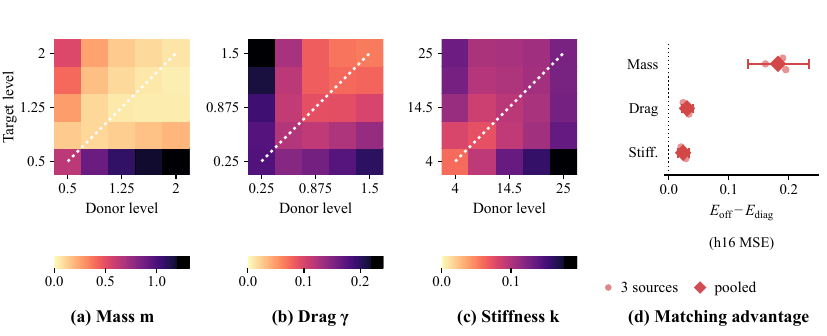}
\caption{\textbf{Donor physics structures fixed predictive responses.}
(a)--(c) Donor--target landscapes of h16 standardized-increment MSE,
averaged over 64 development systems and three Align + Cross sources; color scales differ.
Only donor context changes the fixed reader's prediction.
(d) Mean off-diagonal minus diagonal error; positive favors matching.
Points: sources; diamonds: means; intervals: pointwise 95\% paired-system CIs,
conditional on fitted models (Appendix~\ref{app:landscape}).}
\label{fig:functional-landscape}
\end{figure}}

\newcommand{\SPRIIfigfive}{%
\begin{figure}[!htbp]\centering
\includegraphics[width=\linewidth]{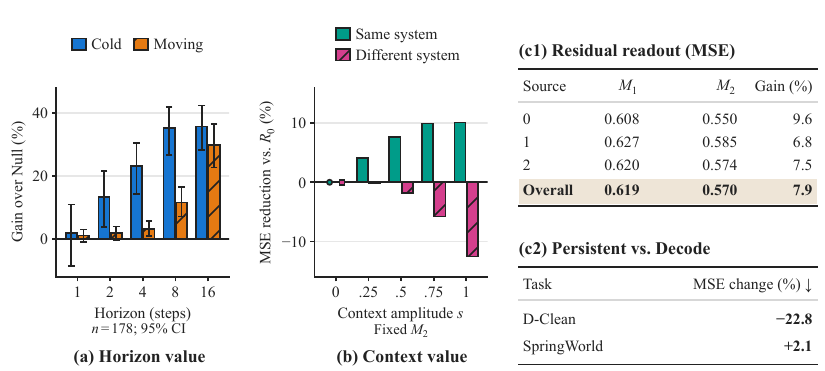}
\caption{\textbf{Persistent context has prediction- and readout-dependent value.}
(a) SpringWorld gain over Null readers as horizon increases; whiskers:
pointwise 95\% system-bootstrap intervals over 178 systems.
(b) Scaling context in fixed $M_2$ changes MSE relative to recipient-only
$R_0$; $s=0$ gives $R_0$.
(c1) Query-only $M_1$ versus context-conditioned $M_2$: source means over
three readers. (c2) MSE change from Decode to Persistent (\%); negative favors
direct context. In (c2), endpoints are D-Clean h32 and SpringWorld cold-start h16
(Appendix~\ref{app:conditional-value}).}
\label{fig:information-use}
\end{figure}}

\begin{document}
\addtocontents{toc}{\protect\setcounter{tocdepth}{-1}}

\maketitle

\begin{abstract}
World models learn environment dynamics from interaction experience.
These dynamics depend on the current state and actions, as well as on
properties that persist across interactions. Yet standard predictive
training can reduce error using local evidence alone, without organizing
persistent information into reusable context. We introduce \textbf{SPRII},
a training principle that uses relations between interactions as weak
supervision for persistent context while retaining the learner's native
objective. For example, different trajectories of the same system share
persistent properties even when their states and actions differ. SPRII uses
such relations to guide context learning without numerical property labels.
Two composable components encourage contexts from related interactions to
agree (Align) and use one interaction's context to predict another's
future (Cross). Our analysis distinguishes three linked questions:
what persistent information is accessible in the learned context
(\textbf{Formation}), how that context influences a fixed predictor
(\textbf{Use}), and whether it reduces task error (Value).
Success at one stage does not guarantee success at the next.
Controlled experiments show that more reliable relations improve
representation organization, but adding a shared-property constraint can
reduce access to a property that remains shared. Context substitutions
change predictions at fixed model weights, while the benefit from history
depends on prediction horizon and readout. Evaluations span thirteen
settings, including controlled physical systems, public dynamics tasks,
robotic and tactile data, and partner interaction, across multiple learner
families. Relative to the corresponding baselines, SPRII yields average
gains of over 10\% in downstream task performance and over 15\% in
persistent-property readout.
The project page is available~at~%
{\hypersetup{urlcolor=blue}%
\href{https://persistent-learning-review.netlify.app/interactive.html}{\texttt{SPRII}}}.

\end{abstract}

\section{Introduction}
\label{sec:introduction}
World models learn environment dynamics from interaction experience to
predict future states and support decisions
\citep{hafner2019planet,hafner2020dreamer,zhou2025dinowm}.
In many systems, future evolution depends on both the current interaction
and properties that remain stable across interactions. Consider two
spring-connected balls driven by external forces:
$s_{t+1}=f(s_t,a_t;\theta)$. The state $s_t$ and action $a_t$ describe the
current configuration and applied force, while $\theta$ contains system
properties such as mass $m$, drag $\gamma$ and stiffness $k$. Different trajectories of the
same system vary in their states and actions but share $\theta$.
These are \emph{persistent properties}; an \emph{interaction} is a realized
episode, and a \emph{history} is its observed sequence of states and actions.

Learning persistent properties helps a model account for the system-specific
dynamics behind observed motion. Related histories can be especially useful when current observations leave
these dynamics ambiguous or when the prediction target is more sensitive to
them, as can occur at longer horizons. An object at rest reveals little
about how quickly it will slow after a push; an earlier interaction can
provide that information.

Context-conditioned models already use
experience to infer how dynamics differ across systems
\citep{lee2020cadm,shaj2022hiprssm,kirchmeyer2022coda}.
However, accurate prediction need not organize system information for reuse.
Strong signals from current states, actions and recent motion can reduce the
incentive to learn reusable context. Physical properties recoverable from
observations can remain difficult to read from predictive latents
\citep{tan2026latent}. Numerical property labels are often unavailable,
leaving no direct supervision for organizing this information.

\SPRIIfigone

Relations between interactions offer weak but often available supervision. We may know that two histories concern the same system, object
or partner without knowing its physical or behavioral parameters. Work on
function representations and relational dynamics learning demonstrates the
value of such relationships \citep{gondal2021fcrl,guo2022ria}.
Pairing histories with shared properties but different states and actions
directs learning toward information that persists across interactions.
The relation need not supply numerical property values; same-system identity
need not even specify which properties govern the shared dynamics.

We propose SPRII (Shaping Persistent Representations from Independent
Interactions), a training principle that uses these relations to supervise
\emph{persistent context}. The predictor combines this history-derived
context with the current input and actions. Two composable training
components encourage related contexts to agree while maintaining variation
(\emph{Align}), and use one interaction's context to predict the future of
another related interaction (\emph{Cross}). SPRII makes the relation between histories an explicit training choice.
Its components use agreement regularization \citep{bardes2022vicreg} and
cross-trajectory prediction \citep{chen2026nod}, retaining the native objective.
Either component, or both, can be applied through an existing history state
or a dedicated encoder (Figure~\ref{fig:architecture}).

Persistent information can be learned without being used, and used without
improving prediction.
We therefore introduce an analysis framework tracing three linked stages:
whether shared information is organized and accessible in the learned
context (\emph{Formation}), whether changing that context affects a fixed
predictor with current inputs and actions held fixed (\emph{Use}), and
whether its use reduces error for the specified task and readout
(\emph{Value}). Geometry and probes, donor substitutions, and task
comparisons examine successive links in this chain; success at one does
not guarantee success at the next. Figure~\ref{fig:overview} illustrates
the chain, with analytical counterparts in Appendix~\ref{app:theory}.

Controlled experiments reveal how these links depend on training and
prediction choices. More reliable relations strengthen cross-interaction
organization; changing which properties are shared redirects accessibility,
even for a property that remains shared. Replacing context steers a fixed reader's predictions along the substituted
physical properties. In SpringWorld, history gains increase with
horizon, and readout changes alter the benefit of frozen context
(Figures~\ref{fig:relation-mechanism}--\ref{fig:information-use}).

Evaluation spans thirteen settings and latent-prediction, contrastive,
recurrent state-space and neural-operator learners
(Appendix~\ref{app:guide}). Across five primary setting--learner comparisons
in spring dynamics (SpringWorld) and object-interaction prediction
(CoPhy; \citealp{baradel2020cophy}), error falls by 14.6\% on average
(range: 9.9--22.3\%) relative to the native learner or a history-conditioned
control without relation objectives (Table~\ref{tab:main-task}). On a controlled inertial system (D-Clean),
32-step error is 33.7\% lower than CaDM under a common readout
(Table~\ref{tab:extended-validation}). Recorded robotic and tactile data
and partner interaction extend the study beyond simulated physical systems.

Our contributions are:
\begin{itemize}
\item \textbf{A relation-supervised training principle.}
We introduce SPRII, which learns persistent context from related interactions
through composable Align and Cross components while retaining each learner's
native objective without requiring numerical property labels.
\item \textbf{A framework connecting information to task benefit.}
We distinguish Formation, Use and Value through operational tests and
analytical counterparts that separate accessible information, its influence
on a fixed predictor, and its contribution to task performance.
\item \textbf{Findings on what is learned and when it helps.}
Controlled studies show how relation reliability and shared properties shape
accessible information, how context influences fixed predictors' outputs,
and how horizon and readout shape the resulting task benefit.
\item \textbf{Generality across learners, relations and tasks.}
We apply SPRII across thirteen settings and multiple learner families,
spanning physical dynamics, robotics, tactile sensing, partner interaction
and control. These applications demonstrate how the same training principle
accommodates different tasks and forms of persistent information.
\end{itemize}

\section{A Relation-Supervised Training Principle}
\label{sec:method}
\subsection{Interactions, relations and persistent properties}
\label{sec:relations}
To predict a new spring trajectory (the recipient), a model can use an earlier
trajectory of the same system (the donor). Denote these interactions by $B$
and $A$. Recipient $B$ has history $H_B$, current input $C_B$, permitted task
conditioning $u_B$ and target $y_B$; donor $A$ supplies history $H_A$.
The relation $\mathcal R_G(A,B)$ identifies shared persistent information,
for example through system, task or partner identity. In controlled systems,
$G$ specifies the shared subset of $\theta$, such as mass while drag varies.

Training samples a recipient, then a related donor, defining pair distribution
$\Pi_G$; representation training uses relation membership rather than numerical factor labels.
Here, \emph{independent interactions} denotes separately realized episodes
or trajectories. The relation defines what is shared, while the sampling
rule selects the legal donor. Sharing an additional factor can therefore
change both the shared information and the selected pairs.
Appendix~\ref{app:relation-sampling} specifies each relation and sampling law.

\subsection{A persistent--current interface}
\label{sec:interface}
SPRII separates history-derived context from recipient-specific current information
through a shared history encoder $g_p$ and a current-input encoder $g_s$:
\begin{equation}
 z_p^A=g_p(H_A),\qquad z_p^B=g_p(H_B),\qquad z_s^B=g_s(C_B).
 \label{eq:factorization}
\end{equation}
Here $z_s$ denotes current-input features, such as image embeddings, rather
than the physical state itself. A dedicated encoder or a projection of an
existing history state implements $g_p$ in one forward pass. Context can
encode several shared conditions in distributed form; prediction uses it
directly, while evaluation probes assess its information content.
For action-conditioned prediction, $u_B$ contains the permitted future
actions, masks and horizon. The predictor $q$ combines these inputs
with a persistent context, whose source can be the recipient or a related donor.

\SPRIIfigtwo

\Needspace{7\baselineskip}
\subsection{Align and Cross}
\label{sec:objectives}
\paragraph{Consistency across related histories.}
Align brings related contexts together while preserving variation.
We implement $\Lalign$ with VICReg \citep{bardes2022vicreg}: paired-context
agreement, per-dimension variance and a cross-dimension covariance penalty.
The relation determines which pairs receive this pressure
(Appendix~\ref{app:training-update}).

\paragraph{Prediction with context from another interaction.}
Cross trains prediction with related donor context. Self and
cross-interaction paths share the recipient's current inputs:
\begin{equation}
\begin{aligned}
 \hat y_B^{\rm self}&=q(z_s^B,z_p^B,u_B),\\
 \hat y_B^{A\to B}&=q(z_s^B,z_p^A,u_B).
\end{aligned}
\label{eq:prediction-paths}
\end{equation}
Switching the context source from $H_B$ to $H_A$ gives the Cross objective:
\begin{equation}
 \Lcross=\mathbb E_{(A,B)\sim\Pi_G}
 \left[\ell_{\rm task}\!\left(\hat y_B^{A\to B},y_B\right)\right].
 \label{eq:cross-prediction}
\end{equation}
The recipient target supplies supervision. Because current inputs may already
suffice, we test reliance on context directly through the interventions in
Section~\ref{sec:use-results}.

\paragraph{Joint training.}
The two operations augment the native objective,
\begin{equation}
 \mathcal L=\mathcal L_{\rm base}
       +\lambda_{\rm p}\Lalign+\lambda_{\rm x}\Lcross,
 \label{eq:sprii-objective}
\end{equation}
where $\mathcal L_{\rm base}$ retains the learner's prediction loss and
regularizers. In our NOD and Overcooked instantiations, the native objective
already uses other interactions, so we add Align. The training recipes in
Appendix~\ref{app:method-interfaces} use Align, Cross or both, with operations
and weights fixed before evaluation.

\subsection{Applying the principle to existing learners}
In JEPA, an observation encoder and causal Transformer combine observation
and action histories into $z_p$; an MLP of the last observation embedding
produces $z_s$. The predictor uses both, future actions and a horizon embedding
to predict the recipient's future embedding. CPC and RSSM retain their native
contrastive and variational objectives; supervised dynamics and action learners
retain their task losses (Appendices~\ref{app:relation-sampling}--\ref{app:method-interfaces};
Figure~\ref{fig:controlled-architecture}).

At inference, SpringWorld encodes an independent same-system history without
gradient updates; other settings identify donors by system, task or partner.
SpringWorld's physical prediction and Use/Value tests use a task reader $r$
fitted to frozen encoder features. This reader is separate from $q$, which
predicts embeddings during source training.

\section{Experiments}
\label{sec:experiments}
\label{sec:results}
SpringWorld shows elastic-coupling dynamics in $128\times128$ images;
PokeWorld shows an actuated finger pushing an object. Both vary mass, drag
and stiffness across systems while holding them fixed within each system.
SpringWorld tests held-out prediction, Use and Value; PokeWorld varies
which factors pairs share. Public benchmarks and robotic, tactile,
multi-agent and control tasks test generality
(Appendices~\ref{app:guide},~\ref{app:data-splits}).

We first compare predictive performance, then examine Formation through geometry
and probes, Use through fixed-predictor context substitutions, and Value through
task-reader and horizon comparisons.
A \emph{source} is a trained representation model; a reader predicts from
its frozen features. Controls are the history-conditioned host (Native),
a split history interface without relation objectives (Structure), and
random pairing with the same relation objective (Random). Readers share targets and
permitted inputs within each comparison. Appendix~\ref{app:measurements}
specifies aggregation and selection; native embedding losses remain separate.

\subsection{Predictive performance}
\label{sec:task-results}
A 96-frame same-system history conditions a SpringWorld query with no preceding motion history;
frozen-source readers predict state increments over the next $0.8$ seconds.
On 256 sealed systems (held out until the final evaluation), all nine
source/reader pairs favor SPRII over Native and TDS
(Table~\ref{tab:main-task}A). CoPhy gains span three Collision learners and Balls RSSM, with budgets
matched within each learner (Table~\ref{tab:main-task}B).

\begingroup
\colorlet{MainTableTint}{KeyRowTint}

\begin{table}[H]
\centering
\small
\setlength{\tabcolsep}{4pt}
\caption{\textbf{Primary predictive performance on SpringWorld and CoPhy.}
Lower is better; bold marks the lowest comparable mean.\protect\hyperlink{main-table-note-1}{\textcolor{black}{\textsuperscript{1}}}}
\label{tab:main-task}\label{tab:application-overview}
\begin{tabularx}{\linewidth}{@{\hspace{3.5pt}}l*{3}{>{\centering\arraybackslash}X}@{\hspace{3.5pt}}}
\toprule
\multicolumn{4}{@{}l}{\textbf{A. SpringWorld: single-state queries, 256 sealed systems}}\\[3pt]
\textbf{Source recipe} & \textbf{MSE $\downarrow$} & \textbf{SPRII reduction} & \textbf{95\% CI}\\
\tableheadrule
Native JEPA & \msd{0.138}{0.002} & 22.3\% & [13.2, 30.7]\%\\
TDS & \msd{0.130}{0.013} & 17.5\% & [8.4, 25.9]\%\\
\rowcolor{MainTableTint}
\textbf{SPRII (Align + Cross)} & \msd{\mathbf{0.107}}{0.004} & \multicolumn{2}{c@{}}{Reference}\\
\end{tabularx}
\par\vspace{9pt}
\begin{tabularx}{\linewidth}{@{\hspace{3.5pt}}l*{4}{>{\centering\arraybackslash}X}@{\hspace{3.5pt}}}
\midrule
\multicolumn{5}{@{}l}{\textbf{B. CoPhy: physical trajectory MSE}}\\[2pt]
& \multicolumn{3}{c}{\textbf{Collision}} & \textbf{Balls4}\\
\cmidrule(lr){2-4}\cmidrule(l){5-5}
\textbf{Source recipe} & \textbf{CPC} & \textbf{RSSM} & \textbf{JEPA} & \textbf{RSSM}\\
\tableheadrule
Native & \msd{0.248}{0.008} & \msd{0.268}{0.007} & \msd{0.250}{0.004} & \msd{1.292}{0.037}\\
Structure & \msd{0.223}{0.006} & \msd{0.228}{0.005} & \msd{0.215}{0.007} & \msd{1.280}{0.031}\\
Random & \msd{0.221}{0.006} & \msd{0.222}{0.007} & \msd{0.265}{0.003} & \msd{1.327}{0.011}\\
\rowcolor{MainTableTint}
\textbf{SPRII (Cross)} & \msd{\mathbf{0.189}}{0.007} & \msd{\mathbf{0.202}}{0.007} & \msd{\mathbf{0.184}}{0.005} & \msd{\mathbf{1.154}}{0.025}\\
\bottomrule
\end{tabularx}

\end{table}
\footnotetext[1]{\hypertarget{main-table-note-1}{}Table~\ref{tab:main-task}: SpringWorld uses standardized state-increment MSE,
three sources $\times$ three readers, source mean $\pm$ SD and paired-system CIs.
CoPhy reports physical trajectory MSE (mean $\pm$ SD over three joint seeds), on 1,994 Collision
episodes and 1,000 Balls4 recipients (Appendices~\ref{app:task-complete},~\ref{app:measurements}).}

\endgroup

\Needspace{8\baselineskip}
SPRII lowers error relative to released NOD in all three Burgers distributions
(Table~\ref{tab:extended-validation}A). With matched additional updates,
SPRII retains a 4.1\% ID gain; continued NOD has up to 1.1\% lower OOD error
(Appendix~\ref{app:external-nod}).
Applied to an amortized variant of CoDA, SPRII lowers mean zero-step error
(Table~\ref{tab:extended-validation}A; Appendix~\ref{app:coda-sprii}).
Using a common reader design across the D-Clean methods, SPRII gives the lowest mean error
at all four prediction horizons, including 33.7\% lower error than CaDM
at 32 steps (Table~\ref{tab:extended-validation}B).

\begingroup
\colorlet{MainTableTint}{KeyRowTint}
\newcommand{\methodvenue}[2]{\hyperlink{cite.#1}{\textcolor{CitationBrown}{(#2)}}}
\begin{table}[H]
\centering
\small
\setlength{\tabcolsep}{4pt}
\caption{\textbf{Predictive comparisons on Burgers and D-Clean.}
Lower is better; bold marks the lowest mean within each comparable host or reader group.\protect\hyperlink{main-table-note-2}{\textcolor{black}{\textsuperscript{2}}}}
\label{tab:extended-validation}
\begin{tabularx}{\linewidth}{@{\hspace{3.5pt}}lXccc@{\hspace{3.5pt}}}
\toprule
\multicolumn{5}{@{}l}{\textbf{A. Burgers: native-trajectory / zero-step MSE ($\times10^{-3}$)}}\\[2pt]
\textbf{Host learner} & \textbf{Source recipe} & \textbf{ID} & \textbf{Viscous OOD} & \textbf{Inviscid OOD}\\
\tableheadrule
NOD \methodvenue{chen2026nod}{arXiv\textquotesingle26} & Released configuration & \msd{9.096}{0.544} & \msd{8.012}{0.815} & \msd{11.857}{0.628}\\
\rowcolor{MainTableTint}
& \textbf{+ SPRII} & \msd{\mathbf{8.999}}{0.798} & \msd{\mathbf{7.487}}{0.183} & \msd{\mathbf{11.510}}{0.360}\\
\addlinespace[3pt]
Amortized CoDA \methodvenue{kirchmeyer2022coda}{ICML\textquotesingle22} & Plain & \msd{3.483}{0.567} & \msd{6.390}{1.436} & \msd{5.883}{1.700}\\
\rowcolor{MainTableTint}
& \textbf{+ SPRII (Cross)} & \msd{\mathbf{3.250}}{0.729} & \msd{\mathbf{6.060}}{1.185} & \msd{\mathbf{5.501}}{1.496}\\
\end{tabularx}
\par\vspace{11pt}
\setlength{\tabcolsep}{2.5pt}
\begin{tabularx}{\linewidth}{@{\hspace{3.5pt}}l*{4}{>{\centering\arraybackslash}X}@{\hspace{3.5pt}}}
\midrule
\multicolumn{5}{@{}l}{\textbf{B. D-Clean: common-reader raw-state MSE ($\times10^{-4}$)}}\\[3pt]
\textbf{Method} & \textbf{h1} & \textbf{h4} & \textbf{h16} & \textbf{h32}\\
\tableheadrule
State-space TDS & $0.1034$ & $0.8009$ & $4.557$ & $13.43$\\
FCRL \methodvenue{gondal2021fcrl}{ICML\textquotesingle21} & \msd{0.1228}{0.0112} & \msd{0.7620}{0.0809} & \msd{4.833}{0.492} & \msd{13.34}{1.014}\\
CaDM \methodvenue{lee2020cadm}{ICML\textquotesingle20} & \msd{0.0332}{0.0018} & \msd{0.1474}{0.0081} & \msd{0.7711}{0.0081} & \msd{2.273}{0.0628}\\
DALI context \methodvenue{dali2025}{NeurIPS\textquotesingle25} & \msd{0.3201}{0.0357} & \msd{2.732}{0.3115} & \msd{17.26}{1.991} & \msd{45.92}{5.362}\\
\rowcolor{MainTableTint}
\textbf{SPRII} & \msd{\mathbf{0.0314}}{0.0029} & \msd{\mathbf{0.1166}}{0.0051} & \msd{\mathbf{0.5232}}{0.0117} & \msd{\mathbf{1.507}}{0.1355}\\
\bottomrule
\end{tabularx}

\end{table}
\footnotetext[2]{\hypertarget{main-table-note-2}{}Table~\ref{tab:extended-validation}: Burgers reports mean $\pm$ SD over three sources,
with 45/500/100 ID/viscous/inviscid queries. D-Clean uses 100 systems and
three sources $\times$ three readers; SD is across source means.
DALI is a context-component adaptation with 160,000 source updates;
other D-Clean sources use 20,000. This is not an equal-total-compute comparison.
Training and evaluation protocols are in Appendices~\ref{app:closest-baselines}--\ref{app:coda-sprii}.}
\endgroup

\FloatBarrier
\subsection{Formation: relations shape strength and accessible content}
\label{sec:formation-results}
PokeWorld permits controlled changes to pairing accuracy and to the
properties pairs share, with source architecture and observation budgets fixed.
\SPRIIfigthree
\Needspace{7\baselineskip}
\paragraph{Reliable relations strengthen organization.}
On 400 confirmation systems, raising the correct-pair fraction from $0$ to $1$
increases the between/within-system context-distance ratio from $0.260$ to
$4.39$. Partial drag geometry, measured by distance correlation controlling
the other factors, rises from $0.031$ to $0.796$
(Figure~\ref{fig:relation-mechanism}(a)). Observation budgets, donor/query
marginals and model settings are fixed.

\Needspace{6\baselineskip}
\paragraph{Changing the pairing rule redirects accessibility.}
Changing $G_1=m$ to $G_2=(m,\gamma)$ keeps mass shared but lowers its linear
probe $R^2$ from $0.527$ to $0.071$; drag rises from $0.308$ to $0.925$.
These probes use per-system context centroids
(Figure~\ref{fig:relation-mechanism}(c)). Nonlinear probes and all three
source replicas reproduce both directions, showing that changing the pairing
rule redirects factor accessibility.
\par\Needspace{9\baselineskip}
In the capacity-limited linear reference of Appendix~\ref{app:predictive-selection},
with code dimension $d$ smaller than history dimension $p$, the selected
persistent subspace is a top-$d$ eigenspace of
\[
M_G=K^\top K-\beta D_G,\qquad \beta\geq0,
\]
where $K$ maps history residuals to target residuals after removing information
explained by current inputs and conditioning, and $D_G$ is the
second-moment matrix of paired-history differences. This balances predictive
relevance against relation disagreement: changing the relation can redirect
the selected directions even when earlier factors remain shared.

\paragraph{Align organizes context; Cross adds benefit.}
SpringWorld's component study uses 178 systems for geometry and 100 development
systems (600 cases; one reader per source) for task error, separate from the
sealed evaluation in Table~\ref{tab:main-task}.
From Structure's separation ratio $0.240$ and reader error $0.172$, Align reaches
$2.05$ and $0.123$; Cross reaches $0.252$ and $0.168$; both reach $2.36$ and
$0.110$ (Appendix~\ref{app:spring-components}). Align supplies most of the
organization and task gain; adding Cross lowers error in all three sources.
In Collision, CPC, RSSM and JEPA use Cross to train the donor-context route.

\Needspace{4\baselineskip}
\subsection{Use: persistent context enters fixed predictive computation}
\label{sec:use-results}
We next substitute donor context to test its effect on a fixed predictor.
In SpringWorld, we intervene on a task reader fitted to frozen features,
separately from the source predictor $q$.
\SPRIIfigfour
\paragraph{Donor physics changes which dynamics a fixed predictor fits.}
In SpringWorld, we hold the source, fitted reader, recipient initial state
and future actions fixed, then vary one donor factor. Independently varying
the corresponding simulator target factor yields a $5\times5$ donor--target
landscape for each of 64 development systems. Simulator targets serve only
to compute prediction error. Matching donor and target physics lowers mean
16-step (h16) error for
mass, drag and stiffness in every source
(Figure~\ref{fig:functional-landscape}(a)--(d)). Donor context therefore
steers the fixed reader's predictions toward the corresponding physical dynamics.

\paragraph{Align strengthens the advantage of matching context.}
Using the same matching-advantage measure as Figure~\ref{fig:functional-landscape}(d),
which shows the joint recipe, we extend the comparison to Structure, Align and Cross.
The Align effect, averaged over the presence and absence of Cross, is positive
for all three factors. Each recipe uses its own source and fitted reader.
Appendix~\ref{app:landscape} gives complete effects and a decomposition
of the same readers' physical output responses.

\paragraph{Relation design has a predictive counterpart.}
On 144 PokeWorld validation systems, a fixed reader's predictions respond
more strongly along mass under $G_1$ and drag under $G_2$
(Appendix~\ref{app:poke-direction}).

\Needspace{4\baselineskip}
\subsection{Value: prediction conditions and the reader shape benefit}
\label{sec:value-results}
We vary the horizon and reader to test the task benefit of frozen context
(Appendix~\ref{app:specificity-theory}), keeping source features unchanged.
\SPRIIfigfive
\paragraph{History value changes with the prediction horizon.}
Null readers receive no history; Persistent readers receive learned context,
Decode readers receive decoded physical parameters, and Oracle readers receive
true parameters. SpringWorld's Cold and Moving queries both provide zero
observed transitions and differ in initial motion. Persistent-over-Null gain
rises from 2.0\% at h1 to 35.7\% at h16 for Cold, and from 1.2\% to 29.7\%
for Moving (Figure~\ref{fig:information-use}(a)). All comparisons use the same
178 systems and fitted grid; horizon changes the target while query evidence stays fixed.

\paragraph{Decodable parameters do not determine the best readout.}
Persistent and Decode readers receive the same donor context, either
directly or through decoded physical parameters, with a shared frozen query
encoder. In D-Clean, direct context lowers h32 error by 22.8\% relative to
Decode despite independently validated decoder $R^2>0.996$.
In SpringWorld, direct context instead has 2.1\% higher error than Decode
(Figure~\ref{fig:information-use}(c2)). The preferred context interface therefore
depends on the task, even when physical parameters are accurately decodable
(paired effects and readers: Appendix~\ref{app:horizon}).

\Needspace{4\baselineskip}
\paragraph{A residual readout tests the added value of context.}
We freeze source features and a fitted recipient-only base $R_0$, then fit
query-only ($M_1$) or context-conditioned ($M_2$) residual readers with common
targets and schedules. On SpringWorld development systems, $M_2$ lowers mean
MSE from $0.619$ to $0.570$ relative to $M_1$, a 7.9\% reduction
(Figure~\ref{fig:information-use}(c1)); all nine source/reader pairs improve
(Appendices~\ref{app:reader-interfaces},~\ref{app:realization-slot}).

With fixed $M_2$, scaling its persistent input by $s$ lowers error for
same-system context and raises it for different-system context
(Figure~\ref{fig:information-use}(b)).
Zero amplitude equals the frozen recipient-only base, distinct from the
separately fitted $M_1$. These substitutions show sensitivity to
system-specific content. Training and evaluating the same reader architecture
with mismatched context raises error in all nine source/reader pairs
(Appendix~\ref{app:reader-capacity-control}), confirming the importance of
correctly associated history.

The readout improves SpringWorld prediction, whereas $M_1$ has lower error
in all five evaluated PokeWorld $G_1/G_2$ runs
(Appendix~\ref{app:realization-slot}). Realized value therefore depends on
the reader and prediction problem.
Appendix~\ref{app:utility-theory}
formalizes available task value and fitted-reader realization. In pendulum control
(Appendix~\ref{app:pendulum-control}), context yields higher returns than CaDM
in four of five conditions.

\subsection{Generality}
\label{sec:generality}
Relations can also express task, object and partner identity. In offline
forecasting from recorded RH20T manipulation, task-matched context under
Align + Cross improves force and motion forecasts relative to specified
mismatched-donor controls
(Appendix~\ref{app:rh-specificity}). In real Baxter grasps, hardness- and
shape-based relations favor readout of the corresponding property
(Appendix~\ref{app:baxter}); in cooperative Overcooked play, same-partner
alignment strengthens partner-behavior readout
(Appendix~\ref{app:overcooked}). The broader CoPhy matrix is reported in Appendix~\ref{app:cophy}.
These settings change both the observations and what persists: physical
dynamics in simulation, object properties in touch, and behavior across
encounters with a partner. Each retains its own task, with relations as a common form of supervision.

\section{Related Work}
\label{sec:related-work}
Context-conditioned dynamics models infer environment factors from transitions
\citep{lee2020cadm,dali2025,shaj2022hiprssm}. NOD conditions prediction on
same-system trajectories \citep{chen2026nod}.
GG-ODE regularizes environment codes contrastively \citep{huang2023ggode};
RIA infers shared-environment relations through interventional prediction
\citep{guo2022ria}. FCRL and view-design studies examine how shared information
shapes representations \citep{gondal2021fcrl,tian2020goodviews,vonkugelgen2021contentstyle};
VICReg supplies Align's agreement objective \citep{bardes2022vicreg}.
These approaches establish ways to learn context or use histories in prediction.
SPRII treats relation reliability and shared properties as training choices,
then traces their effects through three complementary measurements:
geometry and probes of the learned context, donor substitutions through
fixed predictors, and task comparisons under different readouts. This
distinguishes the information made accessible by a relation from its use
and realized benefit (Appendix~\ref{app:extended-related-work}).

\Needspace{8\baselineskip}
\section{Conclusion}
\label{sec:conclusion}

SPRII uses relations between interactions to help world models organize
persistent information into reusable context. Its composable Align and Cross
components retain the learner's native objective and require no numerical
property labels. The Formation, Use and Value framework separates three
questions: what information is accessible in the context, how a fixed
predictor uses it, and whether that use improves task performance. Accurate
prediction alone does not establish all three.

The controlled studies show that more reliable relations improve
representation organization, but requiring additional shared properties can
reduce access to information that remains shared. Context substitutions
reveal sensitivity to persistent information at fixed model weights, while
horizon and readout comparisons show that its predictive benefit depends on
the task and how the context is used. These findings make relation design
and readout design complementary: one shapes the information available in
context, and the other determines how effectively it serves a task.

In practice, system, object or partner identities provide a starting point
for defining related histories; probes, context interventions and task
comparisons can then assess what those relations achieve. Future work could
improve the reliability of relations learned from interaction logs, extend
persistent context to properties that change over longer timescales, and
investigate its role in planning. These directions would extend the
evaluation to evolving environments and decision-making tasks.

\FloatBarrier
\clearpage
\begingroup
\hypersetup{urlcolor=black,linkcolor=black,citecolor=black}
\bibliography{sprii_restructured/references}
\bibliographystyle{iclr2027_conference}
\endgroup

\appendix
\raggedbottom
\clearpage
\addtocontents{toc}{\protect\setcounter{tocdepth}{2}}
\begingroup
\hypersetup{linkcolor=black}
\renewcommand{\contentsname}{Appendix Contents}
\fontsize{9}{10}\selectfont
\setlength{\parskip}{0pt}
\makeatletter
\renewcommand*{\l@section}[2]{%
  \ifnum\c@tocdepth>\m@ne
    \addpenalty{\@secpenalty}\addvspace{3.2pt}%
    \begingroup\fontsize{10}{12}\selectfont\bfseries
    \@dottedtocline{0}{0em}{1.6em}{#1}{#2}%
    \endgroup
  \fi}
\renewcommand*{\l@subsection}[2]{%
  \begingroup\fontsize{9}{10.5}\selectfont\bfseries
  \@dottedtocline{1}{1.5em}{2.3em}{#1}{#2}%
  \endgroup}
\makeatother
\tableofcontents
\endgroup
\clearpage
\renewcommand{\arraystretch}{1.13}
\renewcommand{\topfraction}{0.95}
\renewcommand{\bottomfraction}{0.90}
\renewcommand{\textfraction}{0.05}
\renewcommand{\floatpagefraction}{0.75}
\setcounter{topnumber}{3}
\setcounter{bottomnumber}{3}
\setcounter{totalnumber}{5}
\makeatletter
\setlength{\@fptop}{0pt}
\makeatother
\begingroup
\hypersetup{linkcolor=black}
\section{Findings, Practical Use, and Evaluation Guide}
\label{app:guide}
\subsection{Task and property gains at a glance}
\label{app:gains-overview}

\begin{figure}[H]
\centering
\includegraphics[width=\linewidth]{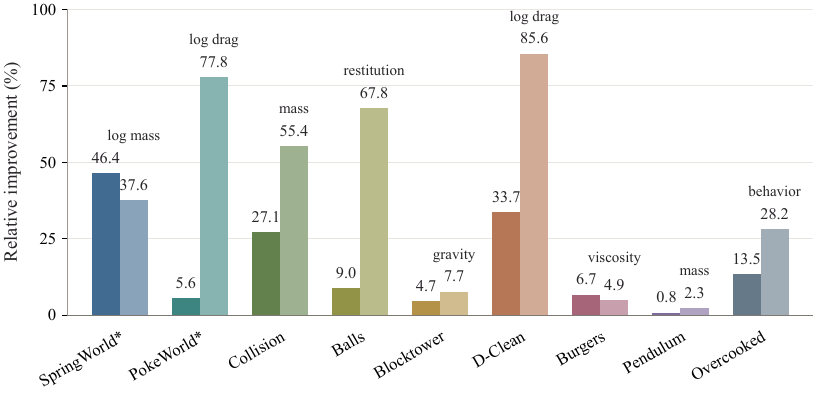}
\caption{\textbf{Selected task and property gains.}
\textbf{Left:} downstream task improvement. \textbf{Right:} improved readout
of the labeled property. Both are relative to the corresponding
baseline. Asterisks mark single-source examples;
Figure~\ref{fig:paired-gain-sources} gives comparison details.}
\label{fig:predictive-gains}
\end{figure}
\FloatBarrier

Figure~\ref{fig:predictive-gains} illustrates two useful outcomes of learning
across related encounters: better task performance and clearer access to
persistent properties. The examples span trajectory prediction, field
forecasting, control and cooperation with recurring partners. These gains
need not grow together: a property can become much easier to read while the
task improves only modestly. The practical lesson is to examine both what a
history representation contains and what a learner can do with it.

\subsection{Discussion: putting the principle to work}
\label{app:scope}

\paragraph{(1) Where the idea applies.}
Many learning problems involve repeated encounters with a system whose
immediate state changes more readily than some of its underlying properties.
A new trajectory may begin at a different position; an object may be
manipulated in a different way; a familiar partner may face a new situation.
Earlier encounters could still contain useful evidence about how each will
respond. The practical opportunity is therefore to connect experiences that
would otherwise be treated as separate examples. This does not require
repeating the same task or observing exactly the same conditions. It requires
a reason to expect some relevant information to persist across the encounters
being connected. Object identifiers, repeated experimental conditions, or
recurring partners may provide such a connection without numerical labels for
the properties themselves. In a new application, a useful starting point is
to identify what recurs, over what period it is reasonably stable, and which
earlier observations would be available when a new prediction is needed.

\paragraph{(2) How to put it into a model.}
A useful design begins by separating what the model must learn from the
current situation from what earlier experience could already tell it. The
current input describes the particular prediction being made, while a history
representation provides a route for information to carry across encounters.
An existing learner with a history state may already offer that route; another
learner could obtain it from a separate encoder. The task itself can remain
the organizing objective. Relation supervision encourages experience from one encounter to be
useful in another, without changing what the model is ultimately asked to predict. In practice, this also makes the organization of recorded
experience part of model design: a training example may be accompanied by
earlier, related experience as well as its immediate inputs. The corresponding
deployment arrangement should make clear where that history comes from and
when it becomes available.

\paragraph{(3) How it may be used.}
Consider a robot encountering the same object several times. A brief earlier
push or contact could provide context for predicting the outcome of a later
movement, even when the object's current appearance gives little indication
of its response. A predictive model could combine that experience with the
new state and proposed action; a subsequent planning system could then use
the resulting forecast when comparing actions. For a dynamics application,
trajectories recorded under the same physical conditions could provide context
for forecasting a new trajectory from different initial conditions. For
repeated cooperation, earlier exchanges with a partner could inform predictions
of that partner's behavior in a new situation. In these applications, history could be organized around an object, a
physical regime, or a partner. The same accumulated knowledge could then
support several predictions as tasks and immediate conditions change.
Experience could remain useful beyond the episode in which it was obtained. These applications also distinguish the reusable idea from a particular
benchmark: the historical observations, current inputs, and desired outputs
can differ across applications. The studies below examine specific instances
of this idea; the broader applications described here remain possibilities
whose value would need to be established in their own setting.

\subsection{Findings and broader implications}
\label{app:claim-map}
The studies trace how experience becomes reusable: what is learned,
how it enters a prediction, and when it helps. Table~\ref{tab:claim-map}
connects the observed patterns to broader ideas for designing learning systems.

\begin{table}[H]
\centering\small
\setlength{\tabcolsep}{4.5pt}
\renewcommand{\arraystretch}{1.14}
\newcommand{\findingpoint}{\hangindent=0.8em\hangafter=1\noindent\makebox[0.8em][l]{\textbullet}}
\caption{\textbf{Findings and possible design implications.}
The evidence column links each observation to its supporting experiments.}
\label{tab:claim-map}
\begin{tabularx}{\linewidth}{@{\hspace{3.5pt}}>{\raggedright\arraybackslash}p{.39\linewidth}>{\raggedright\arraybackslash}X>{\raggedright\arraybackslash}p{.13\linewidth}@{\hspace{3.5pt}}}
\toprule
\rowcolor{TableHeaderTint}
\textbf{Finding} & \textbf{Broader implication} & \textbf{Evidence}\\\tableheadrule
\findingpoint \textbf{More reliable relations} strengthen the organization of shared information.
& How experiences are linked may matter alongside how much data is collected.
& App.~\ref{app:poke-formation}\\\addlinespace[6pt]
\findingpoint \textbf{Sharing more properties} can make a still-shared property harder to read out.
& A useful relation may be one that emphasizes the information a task needs.
& App.~\ref{app:poke-formation}\\\addlinespace[6pt]
\findingpoint In SpringWorld, \textbf{alignment} drives much of the organization; \textbf{cross-prediction} adds further task gains.
& Learning what should stay consistent and learning how to put it to work are distinct goals that can reinforce each other.
& App.~\ref{app:spring-components}\\\addlinespace[6pt]
\findingpoint \textbf{Changing the supplied history} redirects the same predictor along the changed physical properties.
& A representation's role becomes clearer when we examine how decisions respond to it.
& App.~\ref{app:landscape}; \ref{app:poke-direction}\\\addlinespace[6pt]
\findingpoint In SpringWorld, history brings larger gains at \textbf{longer prediction horizons} under the same current-evidence budget.
& Past experience may be most useful \mbox{when it resolves uncertainty} \mbox{left by the present situation.}
& App.~\ref{app:horizon-conditions}\\\addlinespace[6pt]
\findingpoint \textbf{Direct context} works better in D-Clean; \textbf{decoded physical parameters} work better in SpringWorld.
& A more interpretable representation need not be the most useful interface for a task.
& App.~\ref{app:horizon}\\\addlinespace[6pt]
\findingpoint \textbf{Changing the readout} recovers gains from frozen SpringWorld representations, but not from every tested source.
& Progress may come from finding a better way to use existing knowledge, as well as from learning new information.
& App.~\ref{app:realization-slot}\\[3pt]
\bottomrule
\end{tabularx}
\end{table}
\FloatBarrier

\clearpage
\subsection{Evaluation settings}
\label{app:suite}
\emph{Evaluated instance} identifies the learner, enabled operations and
principal controls. A component study
compares the interface alone (Structure), Align, Cross and their combination;
application-specific choices are detailed in Appendix~\ref{app:method-interfaces}.

\begingroup
\fontsize{8.7}{9.7}\selectfont
\setlength{\parskip}{0pt}
\setlength{\tabcolsep}{3pt}
\setlength{\LTleft}{0pt}\setlength{\LTright}{0pt}
\setlength{\LTpre}{5pt}\setlength{\LTpost}{0pt}
\setlength{\LTcapwidth}{\linewidth}
\renewcommand{\arraystretch}{1.08}
\newcommand{\suitepicture}[1]{\par\vspace{4pt}\includegraphics[width=0.98\linewidth,height=54pt,keepaspectratio]{appendix_pipeline/figures/atlas_#1.pdf}}
\newcommand{\suitehead}[3]{\hypertarget{suite-#1}{}\hyperref[#2]{\textbf{#3}}}
\newcommand{\suiteev}[2]{\hyperref[#1]{#2}}
\newcommand{\suitephrase}[2]{\par\noindent\textbf{#1:}~#2}
\newcommand{\suitetarget}[1]{\suitephrase{Target}{#1}}
\newcommand{\suitegroup}[1]{\addlinespace[5pt]\multicolumn{5}{@{}l@{}}{\normalsize\bfseries #1}\\*[3pt]\cmidrule{1-5}\addlinespace[3pt]}
\begin{longtable}{@{\hspace{3.5pt}}>{\raggedright\arraybackslash}p{0.168\linewidth}>{\raggedright\arraybackslash}p{0.174\linewidth}>{\raggedright\arraybackslash}p{0.190\linewidth}>{\raggedright\arraybackslash}p{0.230\linewidth}>{\raggedright\arraybackslash}p{\dimexpr.238\linewidth-8\tabcolsep-7pt\relax}@{\hspace{3.5pt}}}
\caption{\textbf{Evaluation settings and their principal tests.}
Images illustrate the settings, not model predictions.\label{tab:suite-map}}\\
\toprule
\rowcolor{TableHeaderTint}
\textbf{Setting} & \textbf{Relation} & \textbf{Experience $\to$ prediction} & \textbf{Evaluated instance} & \textbf{Primary evidence}\\\tableheadrule
\endfirsthead
\caption[]{\textbf{Evaluation settings and their principal tests (continued).}}\\
\toprule
\rowcolor{TableHeaderTint}
\textbf{Setting} & \textbf{Relation} & \textbf{Experience $\to$ prediction} & \textbf{Evaluated instance} & \textbf{Primary evidence}\\\tableheadrule
\endhead
\multicolumn{5}{r@{}}{\textit{Continued on the next page.}}\\
\endfoot
\bottomrule
\endlastfoot
\suitegroup{Controlled environments}
\suitehead{spring}{app:spring-profile}{SpringWorld}\suitepicture{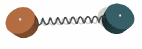}
& \suitephrase{Shared}{Mass $m$, drag $\gamma$, stiffness $k$.}\suitephrase{Varied}{Interaction history.}
& \suitephrase{Input}{Independent history; current query/actions.}\suitetarget{Future state increments.}
& \suitephrase{Learner}{JEPA.}\suitephrase{Instance}{Align + Cross.}\suitephrase{Controls}{Native; TDS.}\suitephrase{Components}{Structure; Align; Cross.}
& \suitephrase{Primary}{\suiteev{app:spring-sealed}{Sealed new-system prediction}.}\suitephrase{Further}{\suiteev{app:spring-components}{Geometry}; \suiteev{app:landscape}{Use}; \suiteev{app:conditional-value}{Value}.}\\[2pt]\midrule
\suitehead{poke}{app:poke-profile}{PokeWorld}\suitepicture{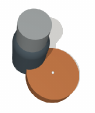}
& \suitephrase{Shared}{Selected factors: $m$; $(m,\gamma)$; $(m,\gamma,k)$.}\suitephrase{Varied}{Correct-pair fraction $\alpha$, separately.}
& \suitephrase{Input}{Independent image/action history; current query.}\suitetarget{Future observation embedding.}
& \suitephrase{Learner}{JEPA.}\suitephrase{Instances}{Structure; Align; Cross; Align + Cross.}\suitephrase{Controls}{Relation and objective composition.}
& \suitephrase{Primary}{\suiteev{app:poke-formation}{Formation: relation reliability and shared factors}.}\suitephrase{Further}{\suiteev{app:poke-formation}{Probes}; \suiteev{app:poke-fixed}{Use}; \suiteev{app:poke-value}{task readouts}.}\\[2pt]\midrule
\suitehead{dclean}{app:dclean-profile}{D-Clean}\suitepicture{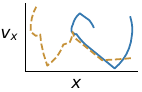}
& \suitephrase{Shared}{Drag $\gamma$.}\suitephrase{Varied}{Initial state; force sequence.}
& \suitephrase{Input}{Independent state/action history; recipient state.}\suitetarget{Future prediction.}
& \suitephrase{Learner}{JEPA.}\suitephrase{Source}{Align + Cross.}\suitephrase{Readouts}{Null; Persistent; Decode; Oracle.}\suitephrase{Comparisons}{TDS; NOD; FCRL; CaDM; DALI.}
& \suitephrase{Primary}{\suiteev{tab:dclean-full}{Formation}; \suiteev{app:horizon}{Value}.}\suitephrase{Further}{\suiteev{app:dclean-official}{Closest methods}.}\\[2pt]\midrule
\suitegroup{Public benchmarks and real data}
\suitehead{cophy_collision}{app:cophy-collision-profile}{CoPhy}\newline\textit{Collision}\suitepicture{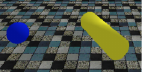}
& \suitephrase{Shared}{Matched objects; scene-specific rule.}\suitephrase{Varied}{Support interaction.}
& \suitephrase{Input}{Three support histories; three query frames.}\suitetarget{Future trajectory.}
& \suitephrase{Learners}{CPC; RSSM; JEPA.}\suitephrase{Instance}{Cross.}\suitephrase{Controls}{Native; Structure; Random.}\suitephrase{Extensions}{Other learners in \ref{app:cophy}.}
& \suitephrase{Primary}{\suiteev{app:cophy-collision-test}{Held-out prediction}.}\suitephrase{Further}{\suiteev{app:cophy}{Learner matrix}.}\\[2pt]\midrule
\suitehead{cophy_balls}{app:cophy-balls-profile}{CoPhy}\newline\textit{Balls}\suitepicture{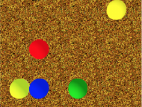}
& \suitephrase{Shared}{Matched objects; scene-specific rule.}\suitephrase{Varied}{Support interaction.}
& \suitephrase{Input}{Three support histories; three query frames.}\suitetarget{Ball trajectories.}
& \suitephrase{Instances}{JEPA, CPC, RSSM: Cross; CoPhyNet: Align + Cross.}\suitephrase{Controls}{Native; Random; Structure where tested.}
& \suitephrase{Primary}{\suiteev{app:balls-heldout}{Held-out RSSM prediction}.}\suitephrase{Further}{\suiteev{app:balls-use}{Memory probes and Use}; \suiteev{app:cophy}{learner matrix}.}\\[2pt]\midrule
\suitehead{cophy_blocktower}{app:cophy-blocktower-profile}{CoPhy}\newline\textit{Blocktower}\suitepicture{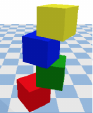}
& \suitephrase{Shared}{Matched objects; scene-specific rule.}\suitephrase{Varied}{Support interaction.}
& \suitephrase{Input}{Three support histories; three query frames.}\suitetarget{Multibody motion.}
& \suitephrase{Instances}{JEPA, CPC, RSSM: Cross; CoPhyNet: Align + Cross.}\suitephrase{Controls}{Native; Random; Structure where tested.}
& \suitephrase{Primary}{\suiteev{app:cophy}{Multibody prediction}.}\suitephrase{Further}{\suiteev{app:cophy-reader-design}{Reader design}.}\\[2pt]\midrule
\suitehead{nod_burgers}{app:nod-burgers-profile}{NOD}\newline\textit{Burgers}\suitepicture{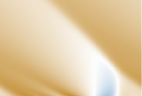}
& \suitephrase{Shared}{Viscosity $\nu$.}\suitephrase{Varied}{Initial field.}
& \suitephrase{Input}{Independent trajectory; recipient field.}\suitetarget{1D field evolution.}
& \suitephrase{Learner}{Released NOD.}\suitephrase{Instance}{Relation-trained stage; prediction-only continuation.}\suitephrase{Controls}{NOD; Random; continued NOD; GEPS/CoDA (50 adaptation steps).}
& \suitephrase{Primary}{\suiteev{app:external-nod}{Public operator prediction}.}\suitephrase{Further}{Viscosity probes; fixed-donor use.}\\[2pt]\midrule
\suitehead{nod_fhn}{app:nod-fhn-profile}{NOD}\newline\textit{FHN}\suitepicture{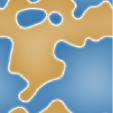}
& \suitephrase{Shared}{Reaction parameters $(k,\beta)$.}\suitephrase{Varied}{Initial field.}
& \suitephrase{Input}{Independent field histories; recipient field.}\suitetarget{2D field evolution.}
& \suitephrase{Learner}{NOD-Hier.}\suitephrase{Instance}{Align; prediction-only adaptation.}\suitephrase{Comparison}{Equal-history NOD.}
& \suitephrase{Primary}{\suiteev{app:external-fhn}{2D prediction; history aggregation}.}\suitephrase{Further}{Factor probes.}\\[2pt]\midrule
\suitehead{baxter}{app:baxter-profile}{Baxter}\suitepicture{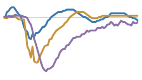}
& \suitephrase{Shared}{Hardness or shape.}\suitephrase{Varied}{The other factor.}
& \suitephrase{Input}{Separate tactile grasp.}\suitetarget{Future tactile representation.}
& \suitephrase{Learner}{Predictive encoder.}\suitephrase{Instance}{Align + Cross.}\suitephrase{Control}{Random relation.}
& \suitephrase{Primary}{\suiteev{app:baxter}{Tactile factor selectivity}.}\suitephrase{Further}{Hardness and shape readouts.}\\[2pt]\midrule
\suitehead{rh20t}{app:rh-profile}{RH20T}\suitepicture{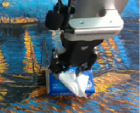}
& \suitephrase{Shared}{Task identity.}\suitephrase{Varied}{Recorded episode.}
& \suitephrase{Input}{Multimodal history; current inputs/actions.}\suitetarget{Force/torque; TCP motion.}
& \suitephrase{Learner}{Supervised predictor.}\suitephrase{Instances}{Cross; Align + Cross.}\suitephrase{Controls}{Structure; pairing; inputs.}
& \suitephrase{Primary}{\suiteev{app:rh20t}{Robot forecasting and task-specific donor use}.}\\[2pt]\midrule
\suitehead{swimmer}{app:swimmer-profile}{Swimmer}\suitepicture{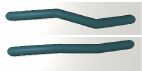}
& \suitephrase{Recipient}{Fixed system.}\suitephrase{Donor}{Own-system or wrong-system update.}
& \suitephrase{Input}{Persistent update; current query/action.}\suitetarget{Prediction through fixed weights.}
& \suitephrase{Learners}{Frozen JEPA/GRU.}\suitephrase{Intervention}{Update substitution.}\suitephrase{Training}{No new Align/Cross source fit.}
& \suitephrase{Primary}{\suiteev{app:fixed-weight}{Fixed-update use}.}\\[2pt]\midrule
\suitehead{overcooked}{app:overcooked-profile}{Overcooked}\suitepicture{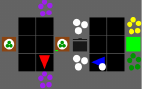}
& \suitephrase{Shared}{Fixed partner identity.}\suitephrase{Varied}{Disjoint episode.}
& \suitephrase{Input}{Prior episodes; current query.}\suitetarget{Ego-action prediction.}
& \suitephrase{Learner}{AD action learner.}\suitephrase{Instance}{Align only.}\suitephrase{Control}{Matched history slot; variance/covariance terms.}
& \suitephrase{Primary}{\suiteev{app:overcooked}{Partner-behavior readout}.}\suitephrase{Further}{Cooperation across history budgets.}\\[2pt]\midrule
\suitehead{pendulum}{app:pendulum-profile}{Pendulum}\suitepicture{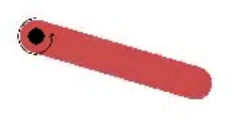}
& \suitephrase{Shared}{System dynamics.}\suitephrase{Varied}{Episodes; mass and length across systems.}
& \suitephrase{Input}{Transition history; current state.}\suitetarget{Dynamics for online control.}
& \suitephrase{Learner}{CaDM interface.}\suitephrase{Instance}{R7 relation recipe.}\suitephrase{Controls}{CaDM; Random relation.}
& \suitephrase{Primary}{\suiteev{app:pendulum-control}{Return and success}.}\suitephrase{Further}{Same-checkpoint factor probes and Use.}\\[2pt]
\end{longtable}
\endgroup
\FloatBarrier

\clearpage
\paragraph{How the paired gains are calculated.}
Figure~\ref{fig:predictive-gains} selects positive examples pairing task and
property measurements under the same model comparison and source set.
Error/cost gains are $100(1-E_{\rm SPRII}/E_{\rm ref})$; probe $R^2$ gains are
$100(R^2_{\rm SPRII}-R^2_{\rm ref})/(1-R^2_{\rm ref})$; Overcooked return gains
are $100(R_{\rm SPRII}-R_{\rm ref})/R_{\rm ref}$, using a positive reference.
Reported measurements are averaged across sources before taking ratios.
These examples include development results and do not estimate average
suite-wide benefit.

\begin{figure}[H]
\centering
\includegraphics[width=\linewidth]{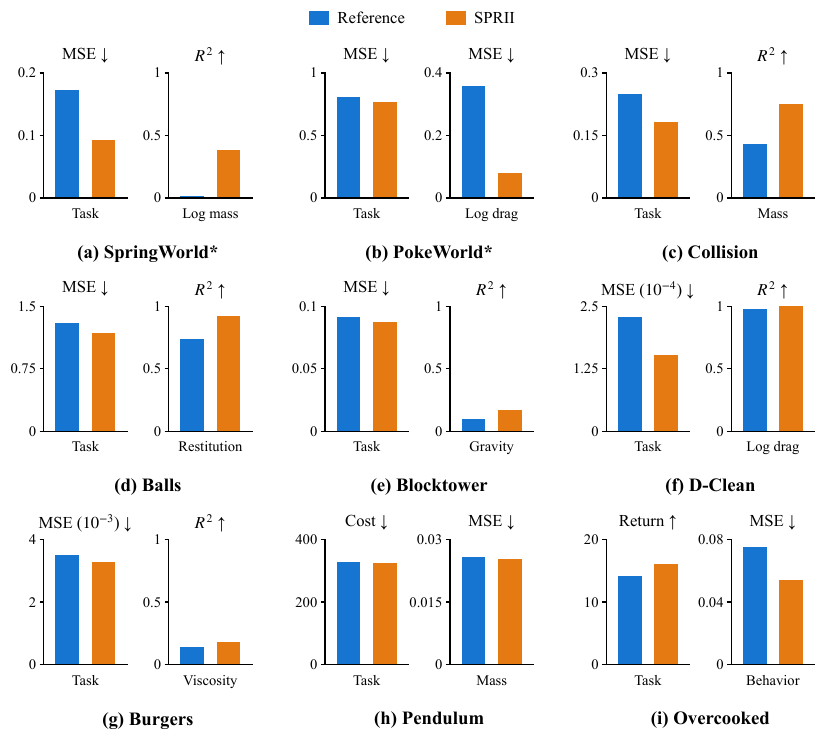}
\caption{\textbf{Original-value task and property comparisons.}
Each setting pairs task performance (left) with property readout (right).
Blue: reference; orange: SPRII. Arrows mark the preferred direction.}
\label{fig:paired-gain-sources}
\end{figure}
\begingroup
\setlength{\parskip}{0pt}
\noindent\small
\textbf{Comparisons.}
(a) Native JEPA $\to$ Align + Cross;
(b) random $\to$ correct pairing;
(c) Native CPC $\to$ Cross;
(d) Native RSSM $\to$ Cross;
(e) Native JEPA $\to$ Cross;
(f) CaDM $\to$ SPRII;
(g) Plain CoDA $\to$ Cross;
(h) CaDM $\to$ SPRII R7;
(i) H+VC $\to$ Align.
\par\smallskip
\textbf{Task metrics.}
(a) Cold-start MSE; (b) h16 state MSE; (c--e) trajectory MSE;
(f) h32 MSE; (g) ID MSE; (h) ID control cost; (i) familiar-partner return.
Collision and Blocktower probes read support $S$; Balls reads memory $M$;
Blocktower's property is vertical gravity.
\par\smallskip
\textbf{Sources and aggregation.}
(a) Table~\ref{tab:spring-complete}: source 0, reader 0;
(b) App.~\ref{app:alpha-m2}: source 0, three readers;
(c,e) Table~\ref{tab:ctp-replication-complete}: two sources;
(d) App.~\ref{app:balls-use}: three sources;
(f) App.~\ref{app:dclean-official}: three sources, three readers each;
(g) Tables~\ref{tab:coda-host-results}, \ref{tab:coda-host-probes}: three sources, code-0;
(h) App.~\ref{app:pendulum-control}: three training seeds;
(i) Table~\ref{tab:overcooked-complete}: three seeds, full history.
\par\smallskip

\noindent\small
*SpringWorld is the source-0 development example, distinct from the sealed
multi-source result. PokeWorld selects the largest task improvement among
sources 0/1/2 ($+5.6\%$, $+1.4\%$, $-15.0\%$), using the same source for
the drag probe and 5,000 reader updates. CoPhy panels use paired development
records; other panels use the stated means. The Balls memory probe includes
the fitted reader projection. Overcooked reports net cooperation return
including action costs, and partner behavior rather than a physical parameter.
\par\endgroup
\FloatBarrier
\clearpage

\endgroup
\FloatBarrier
\section{Method and Experimental Implementation}\label{app:interfaces}
This section specifies how the instances in Table~\ref{tab:suite-map} are built
and measured. We first give the common relation objectives and training update,
then the controlled JEPA implementation and learner-specific interfaces.
Appendix~\ref{app:data-splits} defines the environments and data splits;
Appendices~\ref{app:reader-interfaces}--\ref{app:measurements} define the probes,
readers, interventions and aggregation used in the results.
\FloatBarrier
\subsection{Relations, objectives and training update}
\label{app:relation-sampling}
The relation $\mathcal R_G$ specifies what histories share; the kernel
$Q_G(A\mid B)$ specifies how a legal donor is selected. Drawing a recipient
from $P$ induces the pair law
\begin{equation}
\Pi_G(A,B)=P(B)Q_G(A\mid B).
\label{eq:pairing-kernel}
\end{equation}
Relation semantics and sampling are separate choices: $\mathcal R_G$ defines
the shared conditions, whereas $Q_G$ selects actual legal pairs. Our main setting pairs
separately realized interactions. The formulation also admits broader
pairing constructions, such as overlapping clips, whose shared episode
structure offers different opportunities to rely on local cues.
Partial-factor sharing is distinct from this sampling choice: two separately
realized interactions may share selected factors while others differ. The
supports of their pairing kernels consequently need not be nested, even when
the shared-factor definitions are nested.

The factorized PokeWorld experiment uses
\begin{equation}
G_1=m,\qquad G_2=(m,\gamma),\qquad G_3=(m,\gamma,k).
\label{eq:refinement-keys}
\end{equation}
The first kernel changes drag and stiffness, the second changes stiffness,
and the third changes the rollout while preserving the complete system.
Baxter's hardness and shape relations provide a real-data counterpart.
The network receives observations and task inputs; grouping metadata is used
to construct the pairs. Parameter regression probes are fitted outside representation training.

\paragraph{Objectives and gradients.}
\label{app:training-update}
For paired code matrices $A,B\in\mathbb R^{n\times d}$, the alignment loss is
\begin{align}
I(A,B)&=\frac{1}{nd}\|A-B\|_F^2,\\
V(A,B)&=\frac{1}{2d}\sum_{j=1}^{d}\left[
\max\{0,1-\sqrt{\operatorname{Var}(A_{:j})+\epsilon}\}
+\max\{0,1-\sqrt{\operatorname{Var}(B_{:j})+\epsilon}\}\right],\\
C(A,B)&=\frac1d\sum_{i\ne j}
\left[\operatorname{Cov}(A)_{ij}^2+\operatorname{Cov}(B)_{ij}^2\right],\\
\Lalign&=25I+25V+C,\qquad \epsilon=10^{-4}.
\end{align}
Variances and covariances use the sample denominator $n-1$. Related branches
are evaluated jointly for this loss; no stop-gradient is inserted in Align.
The base learner determines target-encoder gradients for its prediction loss.
In the controlled D-Clean/PokeWorld models, both prediction and target
embeddings are produced by the online observation encoder.

The loss weights and pairing kernels are part of each instantiated training recipe.
For the controlled factorial, the predictor and representation dimensions are
shared across Structure, Align, Cross, and Align + Cross. For transferred learners, the
native comparator and the structure-matched comparator answer complementary
questions: the total effect of the adaptation and the increment from its
relation term. Configuration choices and common evaluation settings are
reported with each environment.
Each application uses a fixed configuration or validation-selected recipe.
Appendix~\ref{app:configuration-reporting} specifies the checkpoint and selection
rules; the profiles below state the enabled operations. History codes are substituted rather than added;
fixed-predictor evaluations make no gradient updates.

Algorithm~\ref{tab:ctp-algorithm} retains the application's native loss and
regularizers. Its task loss, readout, normalization and optimizer are
application-specific. Disabled relation terms are zero; a legal donor obeys
the temporal and input constraints and leaves the recipient target unchanged.

\begin{algorithm}[htbp]
\caption{\textbf{One \sprii{} training update.}}
\label{tab:ctp-algorithm}
\small
\begingroup
\renewcommand{\arraystretch}{1.05}
\begin{tabularx}{\linewidth}{@{}r@{\quad}>{\raggedright\arraybackslash}X@{}}
\multicolumn{2}{@{}l}{\textbf{Inputs:} recipient sampler $P$, donor kernel $Q_G$, encoders and task readout $q$.}\\[5pt]
1 & \textbf{Sample} recipients and related donor histories:\\
  & \hspace*{1.2em}$B\sim P,\qquad H_A\sim Q_G(A\mid B)$.\\[3pt]
2 & \textbf{Encode} the histories and current recipient input:\\
  & \hspace*{1.2em}$\zp^A=g_{\mathrm p}(H_A),\qquad \zp^B=g_{\mathrm p}(H_B),\qquad \zs^B=g_{\mathrm s}(C_B)$.\\[3pt]
3 & \textbf{Compute} the native objective $\mathcal L_{\rm base}$ on ordinary recipient inputs.\\[3pt]
4 & \textbf{Initialize} $\Lalign\leftarrow0$, $\Lcross\leftarrow0$.\\[4pt]
5 & \textbf{if Align is enabled:}\\
  & \hspace*{2.8em}$\Lalign\leftarrow\operatorname{VICReg}(Z_{\rm p}^A,Z_{\rm p}^B)$.\\[4pt]
6 & \textbf{if Cross is enabled:}\\
  & \hspace*{2.8em}$\hat y_B^{A\to B}\leftarrow q(\zs^B,\zp^A,u_B)$.\\[1pt]
  & \hspace*{2.8em}$\Lcross\leftarrow\operatorname{mean}\,\ell_{\rm task}(\hat y_B^{A\to B},y_B)$.\\[4pt]
7 & \textbf{Combine} the native and enabled relation objectives:\\
  & \hspace*{1.2em}$\mathcal L\leftarrow\mathcal L_{\rm base}+\lambda_{\rm p}\Lalign+\lambda_{\rm x}\Lcross$.\\[3pt]
8 & \textbf{Backpropagate} $\mathcal L$ and apply one optimizer update.\\
\end{tabularx}
\endgroup
\end{algorithm}

\FloatBarrier

\FloatBarrier
\subsection{Shared controlled-JEPA implementation}
\label{app:implementation}
\begin{figure}[!htbp]
\centering
\includegraphics[width=\linewidth]{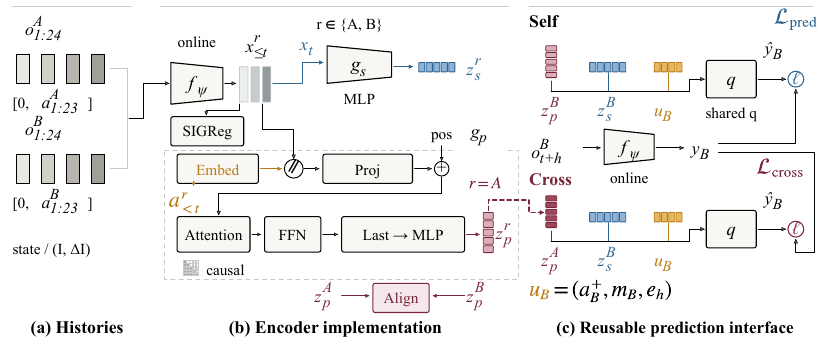}
\caption{\textbf{Controlled-JEPA implementation of SPRII.}
(a) Related observation/action histories. (b) A shared observation encoder
feeds separate current and persistent branches; Align compares the persistent
codes. (c) Self and Cross share the predictor and recipient target. Cross
replaces only the persistent input with the donor's code. Self-prediction and
SIGReg together form the native objective in Figure~\ref{fig:architecture}.
Tokens and the causal-mask crop are schematic.}
\label{fig:controlled-architecture}
\end{figure}

Figure~\ref{fig:controlled-architecture} details the controlled JEPA
instantiation of the general interface. The controlled D-Clean and PokeWorld models use histories of 24 observations and predict encoded targets at
horizons $\{1,4,16\}$. In split variants, $\zs\in\mathbb R^{64}$ is an MLP of
the last observation embedding. The persistent encoder concatenates encoded
observations with 32-dimensional action features. A leading zero action
aligns each later observation with its preceding action; learned position
embeddings are added before processing the sequence with a
four-layer causal Transformer of width 192, eight attention heads, and
four-times-width feed-forward blocks, producing $\zp\in\mathbb R^{64}$. The
predictor receives a 128-dimensional context, padded future actions and masks,
and a 32-dimensional horizon embedding; its two hidden layers have width 256.
The monolithic Native replaces the split encoders with a 128-dimensional history
context while retaining the predictor capacity.

D-Clean observations are 4D states encoded by a three-layer MLP. PokeWorld
renders a $64\times64$ current frame and temporal-difference channel. Its visual
encoder uses four stride-2 convolutional blocks with channels
$32,64,128,256$, followed by a linear projection to 128 dimensions and batch
normalization. Models train for 20,000 steps with AdamW, learning rate
$3\times10^{-4}$, weight decay $0.05$, 500 warmup steps followed by cosine
decay, bfloat16 autocasting, and gradient clipping at 1.0. A paired batch
contains 48 systems and 96 windows ordered as donor/query halves.

SIGReg uses 17 quadrature knots and 1,024 random directions, with
$\lambda_{\mathrm{reg}}=0.02$ \citep{balestriero2025lejepa}. Persistence uses
VICReg weights $25/25/1$ for invariance, variance, and covariance,
unbiased per-branch variance, and $\epsilon=10^{-4}$
\citep{bardes2022vicreg}. The controlled objective weights
$\lambda_{\mathrm p}=1.0$ and $\lambda_{\mathrm x}=0.1$ were frozen before the
three-source factorial.
Rel-InfoNCE uses the InfoNCE objective lineage of contrastive predictive coding
and the two-view group-positive construction of supervised contrastive learning
\citep{oord2018cpc,khosla2020supcon}, with the separate three-seed validation
rule in Appendix~\ref{app:revision-protocol}.

\FloatBarrier
\subsection{Cross-learner instantiation}
\label{app:method-interfaces}\label{app:instantiate-principle}
The persistent--current interface is a structural inductive bias rather than
a shared network architecture. A history encoder or an existing context state
supplies the persistent route; recipient-specific inputs remain available to the
native predictor. Table~\ref{tab:ctp-interfaces} summarizes each learner's
native predictive route, measured representation and enabled relation operations.

The controlled family tests both operations separately and jointly. The CoPhy
source-0 cross-scene development matrix uses one recipe per learner family.
The separately validation-selected Collision JEPA recipe is specified in
Appendix~\ref{app:cophy-profile}. Overcooked's
native action objective already consumes previous episodes, so its added
operation is Align; NOD likewise retains its trajectory-decoupled prediction
objective. Its alignment phase and prediction-only continuation are
specified in Appendix~\ref{app:external-nod}.
Structure isolates the added interface and Random holds the operation fixed
while changing the relation.

\begin{table}[!htbp]\centering\small
\setlength{\tabcolsep}{4pt}
\renewcommand{\arraystretch}{1.06}
\newcommand{\instancepoint}{\hangindent=0.8em\hangafter=1\noindent\makebox[0.8em][l]{\textbullet}}
\caption{\textbf{Predictive routes and relation operations across learners.}
$U$ denotes complete source state, $S$ independent support and $M$ trained
reader memory; each is probed separately.}
\label{tab:ctp-interfaces}\label{tab:v3-learner-instantiation}
\begin{tabularx}{\linewidth}{@{\hspace{3.5pt}}>{\raggedright\arraybackslash}p{0.17\linewidth}>{\raggedright\arraybackslash}p{0.47\linewidth}>{\raggedright\arraybackslash}X@{\hspace{3.5pt}}}
\toprule
\rowcolor{TableHeaderTint}
\textbf{Instance} & \textbf{Native predictive route and measured representation} & \textbf{Relation operations}\\\tableheadrule
\instancepoint \textbf{Controlled JEPA} & 24-step history $P_{64}$, current $S_{64}$; future online embedding MSE + SIGReg. D-Clean state MLP; Poke visual CNN. & Structure, Align, Cross, Align + Cross.\\[2pt]\addlinespace[1.5pt]
\instancepoint \textbf{Spring JEPA} & 96-frame visual/action history; embedding MSE + SIGReg; frozen source and separately fitted physical readers. & Align + Cross, with component controls.\\[2pt]
\midrule
\instancepoint \textbf{CoPhy JEPA/CPC} & Frozen 784D perception features; latent prediction or future contrast; native dynamics state retained. & Cross in the main matrix.\\[2pt]\addlinespace[1.5pt]
\instancepoint \textbf{CoPhy RSSM} & Reconstruction, prior prediction and KL; $U=[P_{64},h_{128},z_{32}]$, versus 160D Native $U$. & Cross; $U$, $S$, $M$ measured separately.\\[2pt]\addlinespace[1.5pt]
\instancepoint \textbf{CoPhyNet} & Coordinate MSE and stability classification; original 32D code split as $P_{16},T_{16}$; reader receives full $U$. & Align + Cross on $P$.\\[2pt]
\midrule
\instancepoint \textbf{Baxter} & 40-step tactile history, $P_{64},S_{64}$; future embedding MSE + SIGReg. & Align + Cross; hardness/shape relations.\\[2pt]\addlinespace[1.5pt]
\instancepoint \textbf{RH20T} & Fused visual/sensor history; joint latent, FT and TCP MSE + SIGReg. & Cross / Align + Cross.\\[2pt]\addlinespace[1.5pt]
\instancepoint \textbf{Overcooked AD} & Ego-action cross-entropy; disjoint prior episodes pooled in a 32D slot. & Align only; $\lambda_x=0$.\\[2pt]
\midrule
\instancepoint \textbf{NOD Burgers/FHN} & Released trajectory-decoupled/hierarchical field operators; native conditional prediction retained. & Align; prediction-only continuation recorded separately.\\[2pt]\addlinespace[1.5pt]
\instancepoint \textbf{Amortized CoDA} & History-inferred context through a frozen CoDA decoder; native field prediction. & Cross; four-arm protocol in Appendix~\ref{app:coda-sprii}.\\[2pt]
\midrule
\instancepoint \textbf{Pendulum / CaDM} & Dynamics context supplied through the existing control interface; return, success and same-checkpoint probes. & Align (R7); fixed recipe in Appendix~\ref{app:pendulum-control}.\\[2pt]\addlinespace[1.5pt]
\instancepoint \textbf{Swimmer} & Recipient-specific persistent update through frozen JEPA/GRU predictors. & Fixed-input substitution; no new Align/Cross source training.\\[2pt]
\bottomrule\end{tabularx}\end{table}

\subsection{Data, splits and admissible inputs}\label{app:data-splits}
Each profile specifies the observations, target, relation and data split.
Table~\ref{tab:suite-map} links to these profiles; Table~\ref{tab:result-identities}
separates the evaluation populations and consuming routes. CoPhy's three
scenes and NOD's two field systems count separately among the thirteen settings.
\subsubsection*{\normalfont\bfseries SpringWorld}
\label{app:spring-profile}
\label{app:spring}
\label{app:springworld}
\paragraph{Environment and information.}
SpringWorld is a controlled visual elastic-coupling environment in which mass
$m$, drag $\gamma$, and stiffness $k$ persist across independent interactions.
The main ranges are $m\in[0.5,2]$\,kg, $\gamma\in[0.25,1.5]$\,s$^{-1}$, and
$k\in[4,25]$\,N/m. Observations comprise $128\times128$ grayscale images and
causal frame differences. A 96-frame donor spans 4.75 seconds; the first
frame difference is zero. Targets are eight-dimensional position/velocity
increments standardized using training statistics. The cold-start query has
no preceding motion history, and prediction covers the next 0.8 seconds.
Physical labels are used for evaluation and downstream supervised heads.

The development readout protocol uses 144 training systems and 178 validation
systems: 64 continuous new systems, 36 unseen discrete combinations, 30 new
interpolated levels, and 48 single-factor extrapolation systems. The original
development cold-start panel uses the first two groups, 100 systems and 600 cases.
The full mixture uses 16,020 cases across the four groups and its fixed query
conditions. They are distinct evaluation profiles rather than different
training-set sizes.

\paragraph{Base learner and adaptation.}
The JEPA reference is a monolithic predictor; its \sprii{} instance introduces
separate history and current-context branches and adds Align with weight 1 and Cross
with weight 0.1. The two use the same data bank, 10,000 source updates,
48 pairs per batch, AdamW learning rate $3\times10^{-4}$, and common
regularization and schedule. The final-step checkpoint is used. Structure,
Align, and Cross isolate the interface and individual operations.

All source representations are frozen before the common readout is trained for 10,000
updates with batch 256 and the same head seed. The readout has a
128-dimensional context interface, 64-dimensional slot, and two 256-unit
hidden layers. Supervised GRU, Transformer, TCN, and DeepSets encoders provide
strong reference representations under their own training recipes; they are
reported alongside the matched JEPA comparisons.

Observation normalization uses a fixed pre-update statistics snapshot at each
source-training update; training-history features update that snapshot only
afterward. The representation is frozen for downstream readers and probes.
The 96-frame donor protocol is specific to SpringWorld, not the 24-frame
controlled D-Clean/PokeWorld input.

\subsubsection*{\normalfont\bfseries D-Clean}
\label{app:dclean-profile}
D-Clean is an analytically integrated two-dimensional inertial system,
$\dot v=F/m-\gamma v$, with $m=1$ and $\Delta t=0.05$. A system fixes drag and
varies initial state and force segments across eight independent 64-step
interactions. The train/validation/test split contains 1,000/200/200 disjoint
systems. The 4D state histories provide a clean setting for measuring factor
accessibility and prediction separately. The history, rather than a single
current state alone, supplies evidence about drag.
The original source comparison is retained in Table~\ref{tab:dclean-full}; the
state-space baseline and same-donor physical bottleneck use their own downstream
populations.

\paragraph{Independent-method comparison.}
NOD uses its released MassSpring network components with D-Clean action inputs;
CaDM uses a PyTorch implementation of its released model components, and FCRL
is implemented from the paper. Each representation is frozen before fitting
an identical downstream reader. Each method uses three source seeds and three
reader seeds per source, with 20,000 source updates and 20,000 reader updates.
Evaluation uses 100 systems from the existing validation report population,
raw-state MSE and horizons $1,4,16,32$. Readers are averaged within source before
computing the three-source mean and sample SD. All source fits are included,
including NOD source 1, whose code is constant. This comparison measures
predictive value under the common reader; the Burgers comparison uses NOD's
native field predictor. The corresponding source probe and fixed-reader donor
interventions use the same fitted models.

\subsubsection*{\normalfont\bfseries PokeWorld}
\label{app:poke-profile}
PokeWorld renders an actuated finger interacting with an object, with mass,
drag and contact stiffness fixed within each system. Observations are current
images and frame differences, both $64\times64$; actions are available to the
history encoder and predictor. Four independent 64-step rollouts vary initial
states and action modes. The original bank contains 2,000/400/400 disjoint
train/validation/test systems.

The factorized bank contains $10\times10\times28=2,800$ physical tuples. A
frozen allocation assigns 20/4/4 stiffness tuples in each mass--drag block to
train/validation/confirmation, producing 2,000/400/400 systems. Exact tuples
are disjoint across splits, while factor levels occur in each split.

Relations $G_1=m$, $G_2=(m,\gamma)$ and $G_3=(m,\gamma,k)$ define which factors are
shared. Training samples one relation-eligible donor per query and update.
For $G_3$, the eligible histories are the other three rollouts of its system.
The relation-eligible pool and the finite candidates retained by a sampler
are different objects; Appendix~\ref{app:donor-pool-control} reports both
and describes the candidate-selection rule. Random uses balanced
system derangements. The fidelity intervention fixes R8 and mixes correct and
random relations at $\alpha\in\{0,0.5,1\}$ while preserving donor/query marginals.
All model settings remain common within each intervention.
The original acquisition bank and the later relation-control banks are separate protocols; Appendix~\ref{app:formation-complete} retains their distinct measurements.

\subsubsection*{\normalfont\bfseries CoPhy}
\label{app:cophy-profile}
\paragraph{Environment and task.}
CoPhy provides visually observed systems whose physical properties govern
different initial conditions and interactions \citep{baradel2020cophy}.
We treat its three scene families as distinct environments in the suite:
\paragraph{Balls.}\label{app:cophy-balls-profile} Balls emphasizes rebound dynamics and physical-factor readout
from ball trajectories.
\paragraph{Collision.}\label{app:cophy-collision-profile} Collision emphasizes contact-dependent motion and is
the held-out task used for the main prediction comparison.
\paragraph{Blocktower.}\label{app:cophy-blocktower-profile} Blocktower emphasizes multibody stability, falling,
and breadth across a different interaction geometry.
\paragraph{Shared protocol.}
The three scenes share a visual front end and episode construction.
We construct cross-experience prediction episodes with three independent
support histories (S3) and the first three query frames (query3).
Prediction starts at query frame four, for 27 frames in Balls/Blocktower and
12 in Collision. Complete validation contains 2,000 Balls, 4,000 Collision,
and 8,088 Blocktower episodes. The original CoPhy task and this multi-support prediction task
have different input and target intervals; the reported matrix uses the latter.

\paragraph{Preserved learners and common evaluation.}
The official supervised perception frontend is frozen. JEPA, CPC, and RSSM
learn dynamics from its 784-dimensional visual features without physical
labels or coordinate targets. CPC retains conditional future discrimination;
RSSM retains a deterministic state, stochastic latent, conditional prior and
posterior, reconstruction, and KL regularization \citep{hafner2019planet}.
Its Native model already trains multistep prior prediction. Source future
prediction never reads the future posterior. Supervised CoPhyNet retains
its original source-training objective.

For JEPA/CPC/RSSM, Native preserves the corresponding base architecture;
Structure adds the persistent--current interface; SPRII adds Cross; Random
keeps the same Cross objective with randomized relations. The supervised
CoPhyNet SPRII instance instead adds both Align and Cross; its Random control
randomizes the relation under that combined objective. Collision uses all-object persistent
substitution in the reported source run. Native representation dimensions are
preserved rather than compressed to a common persistent-only code. Sources
are frozen before a common newly trained prediction head receives the
complete legal source state and S3 history. For example, structured RSSM
exports $U=[P_{64},h_{128},z_{32}]$ with 224 dimensions, whereas Native
exports its 160-dimensional recurrent state. The supervised split CoPhyNet
code is $U=[P_{16},T_{16}]$. The training-time persistent branch and the
complete state exposed to a head or probe are therefore distinct interfaces.

Every main head is trained for 100 epochs. The source budget is 100 epochs for CPC and RSSM. In the held-out
Collision comparison, all four JEPA arms use 50 source epochs, matching the
selected J2 source budget; the earlier unequal-budget control rows are not
used in this comparison. The JEPA
comparison has four arms and three joint source/reader seeds on the same
1,994 held-out Collision episodes. These fits extend evaluation of the existing
test cohort. All conditions use common queries, targets and visible-object
eligibility within each reported population.
Balls/Collision use direct query-pose perception; Blocktower uses its cached
feature and matching pose-head route. These paths are held fixed within scene.
\paragraph{Operation-specific source paths.}
CPC matches contrastive candidates by public time, object slot and visible type.
RSSM filters the first three query frames with zero persistent input, preventing
initialization from bypassing the tested history path; Cross predicts through
its prior rollout. The supervised CoPhyNet route retains the original
32-dimensional AB code, split into $P_{16}$ and $T_{16}$. Align acts on $P_{16}$;
Cross replaces that part while keeping the recipient's remaining code and
initial query state. The common frozen-source reader uses the full
32-dimensional representation. These implementations preserve native objectives
without imposing a common code.

\subsubsection*{\normalfont\bfseries Baxter tactile grasps}
\label{app:baxter-profile}
We use the public Baxter tactile hardness bank of
\citet{amin2023hardness} from the dataset release of
\citet{amin2026hardnessdata}. Each grasp peak is an $80\times16$ sequence. The
primary crossed bank retains cube and cylinder objects at the three hardness
levels shared by both shapes, yielding six known configurations and 170 grasps
per configuration. A single permutation of grasp identifiers (Table~\ref{tab:appendix-rng-settings})
is reused across configurations and split 110/30/30 into
train/validation/confirmation interactions. The 40-sample files are alternate
processed windows from the same acquisitions. The available input consists
of grasp-peak tactile histories; pre-contact frames and action sequences are
unavailable.

For $\mathcal R_H$, a donor shares the hardness level and changes
shape; for $\mathcal R_S$, it shares shape and changes hardness level. Random
uses a uniform derangement among the other five configurations. Every batch
contains eight queries from each configuration and exactly one different-grasp
donor per query, preserving balanced query and donor marginals.

The split architecture uses a per-sample MLP, a four-layer causal Transformer persistent
encoder, 64-dimensional current-context and persistent codes, and prediction at
$+1,+20,+40$ samples. All conditions use 10,000 updates, 48 query--donor
pairs per batch, and seeds 0/1/2. Six randomly initialized supervised
input-readout fits (early 8 samples versus full 80 samples) and nine world-model
fits are frozen before the single confirmation evaluation.
The source input is a 16-dimensional tactile sample and the causal history
contains 40 steps. The reported source recipe enables Align + Cross for both
partial-factor relations. Its six-configuration visualization and complete
selectivity results appear in Appendix~\ref{app:baxter}.

\subsubsection*{\normalfont\bfseries OvercookedV2}
\label{app:overcooked-profile}
\paragraph{Environment and base learner.}
OvercookedV2 is a cooperative cooking environment with coordination-dependent
reward \citep{gessler2025overcooked,rutherford2024jaxmarl}.
We use its native multi-episode AD setting \citep{jing2026icrl4aht}:
the controlled entity is a fixed neural partner whose behavior persists
while locations, actions, and outcomes change across episodes. The training
bank contains 2,560 histories and 37,376,000 transitions from 20 partners
drawn from ten partner-training sources. Nested half- and quarter-history subsets
retain 1,280 and 640 histories, or 18,688,000 and 9,344,000 transitions,
with 64 and 32 rather than 128 histories per partner.
Each history contains 146 recorded 100-step episode prefixes. 
The layout is
\texttt{grounded\_coord\_simple}. Two additional partners from one excluded
training source form the held-out development cohort; they are absent from
policy training. Relation metadata identifies the partner for sampling
without exposing its network parameters to the learner.

\paragraph{Matched adaptation.}
The AD action cross-entropy already consumes prior interactions.
The H condition adds a 32-dimensional persistent history slot; H+VC adds
variance/covariance regularization; H+SPRII additionally aligns histories of
the same partner. The matched conditions share architecture, initialization
and ordered training batches within a seed. The action objective reads the
persistent slot, so SPRII uses Align without an additional Cross loss.
The primary comparison is H+\sprii{} versus H+VC at full, half and quarter
history budgets. Each pair uses 20,000 updates, effective batch 1,024,
microbatch 128, learning rate $3\times10^{-4}$, and training seeds
4200, 4201 and 4202. The losses are
$L_{\rm CE}+0.001(25V+C)$ and $L_{\rm CE}+0.001(25I+25V+C)$, respectively.
Both retain two 100-step support episodes and a 300-step query.
All 18 fitted models use their fixed 20,000-update endpoints. The history-budget
intervention subsamples this fixed offline bank; partner training and collection
costs lie outside this budget.

\paragraph{Evaluation and replication.}
Each frozen endpoint is evaluated for 20 episodes with each of the 20 familiar
and two held-out development partners: 440 episodes per model, with no
optimizer updates. The primary return averages episodes 6--20 within partner,
then averages partners within each cohort. Return and probe summaries report
the mean and sample SD over the three training seeds. Formation and functional
probes use the same frozen endpoints and common sample plan; the Random
condition has three probe seeds and no return evaluation.
The matched/null/wrong-history donor intervention uses one frozen checkpoint.

\paragraph{History interface.}
The dynamic path retains the AD observation encoder, action/reward tokens and
causal Transformer. Each disjoint support episode is processed separately by
the shared backbone; its final valid token projects to a 32-dimensional
persistent slot. Attention pooling combines available slots, and a fusion MLP
reads that result with the recipient's causal query representation to predict
the ego action. H, H+VC and H+SPRII share this interface. Their common action
loss already consumes earlier episodes; the tested increment is alignment.

\subsubsection*{\normalfont\bfseries RH20T}
\label{app:rh-profile}
RH20T records multimodal robot manipulation \citep{fang2023rh20t}. The
operative relation groups different episodes by task identity. It supplies
repeatable task context, without assuming known physical-parameter equivalence.
Models predict normalized force/torque and TCP motion from synchronized visual,
robot-state and action inputs. The representation and prediction heads are
trained jointly; this application uses supervised prediction targets.

\paragraph{Matched conditions.}
The complete matrix contains 14 conditions and three seeds each. All share
prediction windows, normalization, targets and checkpoint step. ``Indep''
denotes a matched different-episode donor, ``SameEp'' a causal non-overlapping
donor from the query episode, and ``Random'' balanced random training groups.
LowDim is a distinct input/representation
route whose controls are compared within that route. Evaluation keeps the
query, target, horizon and actions fixed while changing self, matched and
shuffled-task context. Task-macro results aggregate 25 held-out tasks.

Visual and low-dimensional sensor embeddings are fused before the causal
history encoder. A common prediction trunk feeds latent, six-dimensional
force/torque and three-dimensional TCP heads, all trained jointly.
Cross changes the persistent input while retaining current context and task
conditioning. The 25-task forecasting results are in
Appendix~\ref{app:rh20t}; the 24-task hard-negative bank is specified separately
in Appendix~\ref{app:rh-specificity}.

\subsubsection*{\normalfont\bfseries Articulated Swimmer}
\label{app:swimmer-profile}
Articulated Swimmer is a three-link MuJoCo system with three link masses and
two joint damping parameters. The task predicts state responses to actions.
Each frozen architecture retains its own training-only normalization.
Appendix~\ref{app:fixed-weight} compares own- and wrong-system persistent updates
within each architecture, keeping the source and predictor weights fixed.

\subsubsection*{\normalfont\bfseries Pendulum control}
\label{app:pendulum-profile}
Pendulum tests whether the learned dynamics context supports an online
controller when mass and length vary between systems. The development-selected
R7 relation recipe uses weight 0.003 and variance target 0.1 with the existing
CaDM control interface. It is evaluated intact on ID and four OOD groups,
with three training seeds and ten 200-step episodes per group. Return,
unchanged upright-success criteria and same-checkpoint physical probes are
reported together in Appendix~\ref{app:pendulum-control}; logged-action donor
interventions are not presented as online controller interventions.

\subsubsection*{\normalfont\bfseries NOD-Burgers: one-dimensional field evolution}
\label{app:nod-burgers-profile}
\begin{figure}[htbp]\centering
\includegraphics[width=\linewidth]{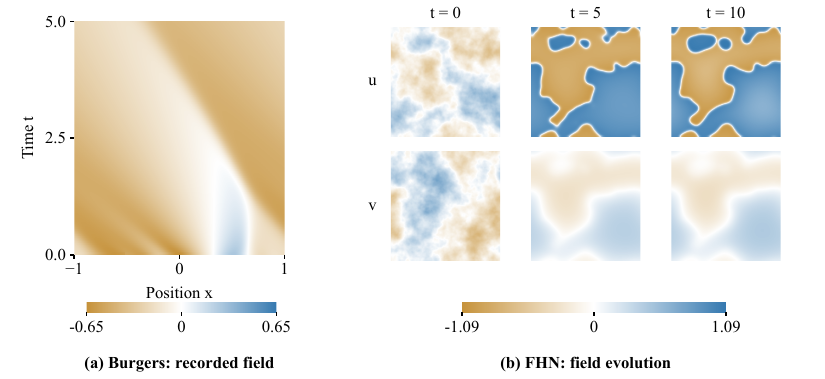}
\caption{\textbf{External field environments.}
(a) A released Burgers training trajectory at $\nu=0.01$.
(b) A separately simulated FHN trajectory at $k=0.03,\beta=0.20$.
FHN panels share a signed field scale; Burgers uses a separate scale.
These are environment illustrations, not prediction results.}
\label{fig:nod-environments}
\end{figure}

Burgers uses a spatial field rather than a rigid-body scene. Independent
trajectories share viscosity $\nu$ while differing in initial conditions;
the recipient field provides the query state. The released trajectory shown
in Figure~\ref{fig:nod-environments}(a) has 401 spatial samples and 101 saved
times. Native prediction, viscosity accessibility and donor-use evidence,
including three-seed alignment and continuation comparisons, are specified in
Appendix~\ref{app:external-nod}.

\subsubsection*{\normalfont\bfseries NOD-FHN: two-dimensional reaction--diffusion}
\label{app:nod-fhn-profile}
The FitzHugh--Nagumo (FHN/DR2D) environment evolves two coupled fields $u,v$
on a periodic $128\times128$ grid. Related trajectories share reaction
parameters $(k,\beta)$ with independent Gaussian-random-field initial
conditions. The Python port follows the released fourth-order spatial
stencil and RK4 update ($\Delta t=0.001$), saving 101 frames through $t=10$.
Conditioning uses frames $0,25,50,75,100$ from an independent trajectory;
the hierarchical operator predicts the recipient's field evolution.
The missing original initial-condition bank was regenerated, so these are
reproducible Python simulations, not the authors' exact data.
\paragraph{Training schedules and selection.}
Both methods use batch size 8 and RMSprop, with initial learning rate
$0.5/\sqrt{P}$, where $P$ is the number of trainable parameters, and a factor
of 0.95 decay every 500 updates. NOD uses prediction-only stages ending at
20,000, 30,000, 40,000 and 45,000 updates. SPRII uses alignment weight 0.01
with a 2,000-update linear warmup through update 10,000, followed by
prediction-only stages ending at 30,000, 40,000 and 50,000. Both recipes
reset the learning rate to $10^{-5}$ after update 40,000 while retaining
RMSprop moments and scheduler state. Each continuation restarts its seeded
sampling stream.

Seed 42 selects the recipes using mean ID relative-$L_2$ error across
H1/H5/H50 on initial condition 5. Seeds 43/44 repeat the selected stages and
endpoints without further checkpoint selection; reporting uses initial
conditions 15/24/45. The NOD and SPRII recipes therefore have different
update budgets. Appendix~\ref{app:external-fhn} reports the three-source
comparison and conditioning-history experiments.

\FloatBarrier
\subsection{Probes, fresh readers and fixed consuming routes}
\label{app:reader-interfaces}
\label{app:comparison-protocols}
A \emph{complete-method comparison} changes the trained source recipe; a
\emph{fresh-reader comparison} fits alternative task readers on a frozen source;
a \emph{fixed-route intervention} changes context through the same fitted weights.
They answer different questions even when the same source model appears in all
three. The measurement definitions below complement the task-specific controls
in Section~\ref{sec:experiments}.

\begin{table}[!htbp]\centering\small
\caption{Terminology used in the main text and appendix.}
\label{tab:reader-terminology}
\begin{tabularx}{\linewidth}{@{\hspace{3.5pt}}lX@{\hspace{3.5pt}}}\toprule
\rowcolor{TableHeaderTint}
\textbf{Term} & \textbf{Meaning}\\\tableheadrule
Persistent context & History-derived representation supplied to a predictor; source weights are frozen for the stated diagnostic evaluations.\\\addlinespace[1pt]
Probe & Supervised diagnostic fitted to a frozen code; measures accessibility.\\\addlinespace[1pt]
Task reader & Predictor fitted on fixed source features; measures realized task value.\\\addlinespace[1pt]
Fixed-predictor intervention & Donor replacement with all source and consuming weights fixed.\\\addlinespace[1pt]
Native & History-conditioned learner with its original context interface and objective.\\\addlinespace[1pt]
Structure & Separate persistent/current interface, without relation objectives.\\\addlinespace[1pt]
Random & The instantiated relation objective with a wrong-donor pairing control.\\\bottomrule
\end{tabularx}\end{table}

\paragraph{Input evidence and representation accessibility.}
An estimator trained on the model-visible history checks whether factor evidence
is recoverable from the input; a privileged-input estimator is a separate
reference. A representation probe fits a supervised factor readout of a frozen
representation, using the training systems
for fitting. Each result identifies its input: code centroids, actual donor
windows, complete source state $U$, independent-support code $S$, or learned
reader memory $M$. Probes measure accessibility, while cross-history geometry
measures organization; both inform Formation. A memory probe includes the
trained reader projection, while a source probe reads the frozen encoder alone.

\paragraph{Fresh task readers and fixed-route use.}
Null, Persistent, Decode and Oracle fit separate task readers with their stated
inputs; Oracle supplies true parameters to a learned reader, not an optimal-risk
bound. A fixed Wrong intervention substitutes context through an unchanged
reader, keeping source, predictor and any output decoder fixed. The resulting
output or risk change measures Use under that substitution. Fresh-reader task
errors measure realized Value under the compared inputs and fitting recipe.
A separately trained Shuffled/Wrong arm differs from fixed-input substitution.

\paragraph{Frozen-source realization routes.}
Let $h$ denote the frozen recipient feature, $p$ the frozen persistent code,
and $R_0(h)$ the separately fitted, then frozen recipient-only base reader.
The 128-unit residual reader computes $v=\mathrm{GELU}(W_vh)$ and
$q=W_qh+b_q+W_{\rm proxy}h$. With elementwise product $\odot$,
\begin{align}
M_1(h)&=R_0(h)+W_o[v\odot\tanh(q)],\\
M_2(h,p)&=R_0(h)+W_o[v\odot\{\tanh(q+W_pp)-\tanh(q)\}].
\end{align}
The persistent projection and output projection are bias-free, so
$M_2(h,0)=R_0(h)$ by construction (up to floating-point evaluation order).
A zero-context penalty therefore cannot alone identify donor specificity.
The query-only arm registers the same persistent
projection but does not use it or update it by gradients: the
architecture is shared, active capacity is not exactly matched. The Oracle
route replaces the learned code with a projection of true parameters and
uses the same residual family.

The \texttt{phys} variants multiply the residual output coordinatewise by a
fixed, nonnegative weight vector derived from the prescribed physical
sensitivity procedure. They do not add a physical loss or supply true
parameters to $M_2$. Source encoders and the 256-unit base are frozen during
residual fitting. Reader fitting uses MSE, AdamW, gradient clipping at 1,
and nominal 1,000-step base and route budgets with batch size 128 and
learning rate $10^{-3}$; development error selects the checkpoint.
The physical target and evaluation population are held fixed within each pair.

The results of these routes appear in Appendix~\ref{app:realization-slot}.
We evaluate $M_2(h,sp)$, using $s$ for the post-fitting context amplitude
and reserving $\alpha$ for training-pair reliability. The zero-amplitude identity is $M_2(h,0)=R_0(h)$,
not $M_1(h)$.
\FloatBarrier
\subsection{Measurement, aggregation and selection}
\label{app:measurements}
\paragraph{Geometry and factor accessibility.}
Controlled codes are standardized using training statistics. Within-system
distances measure stability across interactions; between-system distances
measure separation. The reported between/within ratio is averaged at the
source level, not computed from pooled distances. The Balls memory assay uses
the inverse within/between ratio and labels it explicitly. Partial factor
geometry correlates representation distance with factor distance while
controlling the stated other factors; it is distinct from supervised-probe
$R^2$. Probe fitting and selection use training systems. Transient position
and velocity targets use regression, while contact uses AUROC.

\paragraph{Prediction targets and scales.}
The controlled source objective predicts each learner's own observation
embedding, so self losses characterize native training. Physical-space
comparisons use downstream targets: SpringWorld readers predict
train-standardized eight-state increments; D-Clean readers in the state-space
and same-donor studies report raw-state MSE on different populations. RH20T
uses normalized force/torque MSE and TCP error in cm.

\paragraph{Evaluation populations.}
We use \emph{sealed test} for the separately held-back evaluations identified as
such in the experiment records, \emph{held-out official episodes} for the CoPhy
test cohort, and \emph{confirmation} for the frozen relation-design evaluations.
Development results retain their own populations and selection rules.
SpringWorld's 100-system common-adapter cold-start comparison and its 178-system
cold-query horizon curve are different evaluations: their h16 history gains are
36.10\% and 35.68\%, respectively. The sealed method comparison uses a separate
256-system population.

Table~\ref{tab:result-identities} lists the distinct evaluation populations
and consuming routes used by the functional analyses.
\begin{table}[tbp]\centering\small
\setlength{\tabcolsep}{4pt}
\renewcommand{\arraystretch}{1.08}
\caption{\textbf{Evaluation populations, fitted-model coverage and consuming routes.}}
\label{tab:result-identities}
\begin{tabular}{@{\hspace{3.5pt}}>{\raggedright\arraybackslash}p{0.23\linewidth}>{\raggedright\arraybackslash}p{0.31\linewidth}>{\raggedright\arraybackslash}p{\dimexpr.46\linewidth-4\tabcolsep-7pt\relax}@{\hspace{3.5pt}}}
\toprule
\rowcolor{TableHeaderTint}
\textbf{Analysis} & \textbf{Population} & \textbf{Source, reader and measurement}\\
\tableheadrule
\addlinespace[2pt]\multicolumn{3}{@{}l}{\normalsize\bfseries SpringWorld}\\*[3pt]\cmidrule{1-3}\addlinespace[2pt]
Sealed comparison & 256 continuous primary;\newline 36 factorial secondary & $3$ sources $\times$ $3$ readers; matched inputs.\newline The 292-system pool is auxiliary.\\\tableentryrule
Original reader & 100-system cold start;\newline 178-system mixture & Separate grids:\newline source 0/reader 0 reference;\newline three-source component mean with reader 0;\newline original $3\times3$ comparison.\\\tableentryrule
Common adapter & 100-system bottleneck;\newline 178-system horizon/probes & FCRL uses each source's own query encoder;\newline the four bottleneck arms share one source's query encoder.\\\tableentryrule
Landscapes & 64 development systems & Three sources, reader 0; fixed donor--target interventions.\\[1.5pt]
\midrule
\addlinespace[2pt]\multicolumn{3}{@{}l}{\normalsize\bfseries Controlled D-Clean and PokeWorld}\\*[3pt]\cmidrule{1-3}\addlinespace[2pt]
D-Clean TDS & 100-system report & $3\times3$ grid; selected state-space source and common raw-state reader.\\\tableentryrule
D-Clean same-donor & 200 validation systems & Common Align + Cross query encoder;\newline $3\times3$ grid, with 27 Null/Persistent/Oracle and 9 Decode cells.\\\tableentryrule
Poke geometry/probes & 400 confirmation;\newline 400 validation systems & Centroid probes and actual donor-window probes are separate measurements.\\\tableentryrule
Poke physical-risk use & 96 development systems & Wrong-factor physical MSE;\newline native predictor and state decoder fixed.\\\tableentryrule
Poke native direction & 144 validation systems & G1/G2, three sources each;\newline dimensionless projection in each source's native embedding.\\\tableentryrule
Poke fresh-reader value & 400 development systems & Separately fitted task readers;\newline Shuffled is a trained arm, not a fixed-input substitution.\\[1.5pt]
\midrule
\addlinespace[2pt]\multicolumn{3}{@{}l}{\normalsize\bfseries Transferred learners, robot data and partners}\\*[3pt]\cmidrule{1-3}\addlinespace[2pt]
Collision test & 1,994 held-out episodes & JEPA, CPC and RSSM each have 3 joint source/reader seeds per arm;\newline variation across trained fits is reported as source SD.\\\tableentryrule
Balls extension & 2,000 development recipients & The reporting cohort includes source/readout selection examples.\\\tableentryrule
RH20T & 25-task forecast;\newline 24-task hard negatives;\newline separate 25-task input validation & Specificity and absolute risk share the hard-negative bank.\newline Input controls use the separate validation bank.\\\tableentryrule
Overcooked & 20 familiar and 2 held-out development partners & One layout; 3 training seeds for return/probes at each history budget.\newline Return mean and sample SD are over seeds;\newline donor interventions retain one checkpoint.\\[1.5pt]
\bottomrule
\end{tabular}
\end{table}

\paragraph{Source and reader dispersion.}
For cell error $E_{sr}$, the mean first averages $R_s$ fitted readers within
source, $\bar E_s=R_s^{-1}\sum_r E_{sr}$, then averages $S$ sources equally.
The displayed source SD is
\[
 s_{\rm source}=\sqrt{\frac{1}{S-1}\sum_{s=1}^{S}
 (\bar E_s-\bar E)^2},\qquad
 \bar E=\frac1S\sum_{s=1}^{S}\bar E_s.
\]
A $3\times3$ grid therefore represents three independently trained sources,
each evaluated with three fitted readers.
When source and reader seeds vary jointly, SD describes the complete fitting
pipeline. In incomplete grids, the fitted reader counts are reported rather
than treating the grid as balanced.

\paragraph{Paired effects.}
Complete-method improvement is
$100(\bar E_{\rm control,matched}-\bar E_{\rm SPRII,matched})/
\bar E_{\rm control,matched}$.
Within-source fresh-reader history gain instead compares separately trained
Null and Matched readers. Fixed-weight wrong-minus-matched error measures
reliance on the supplied donor. Relative effects use ratios of mean errors,
not averages of case-wise percentages. In the same-donor bottleneck discussion,
relative differences use Decode as the denominator: Persistent is 22.8\% lower
in D-Clean and 2.12\% higher in SpringWorld. Raw Decode-minus-Persistent
differences retain their original units and intervals. Intervals pair conditions within the
appropriate unit before averaging fitted seeds and resampling that unit.
Analyses specify their system/task/recipient unit, stratification,
multiplicity and bootstrap seed.
\begin{table}[!htbp]\centering\small
\caption{Fixed randomization settings used in the appendix.}
\label{tab:appendix-rng-settings}
\begin{tabular}{@{\hspace{3.5pt}}llr@{\hspace{3.5pt}}}\toprule
\textbf{Experiment} & \textbf{Operation} & \textbf{Seed}\\\tableheadrule
Baxter & Grasp-identifier split permutation & 20260822\\
RH20T additional 25-task bank & Paired task bootstrap & 20260923\\\bottomrule
\end{tabular}\end{table}

\paragraph{Checkpoint and configuration selection.}
CoPhy's cross-scene development and held-out Collision comparisons use the
distinct recipes specified in Appendix~\ref{app:cophy-profile}.
Each analysis specifies whether it uses a final-step checkpoint, a
validation-selected reader or a released-default reference.

\paragraph{Coverage and selection.}
\label{app:replication-status}\label{app:records-identity}
\label{app:baseline-provenance}\label{app:history-provenance}\label{app:configuration-reporting}
The source-0 CoPhy matrix has second-source JEPA fits in all three scenes
and CPC Collision, plus three-source RSSM Collision. The held-out Collision
comparison includes four arms with three joint source/reader seeds each for
JEPA, CPC and RSSM.
Balls separately has three-source RSSM (four arms) and CoPhyNet
(three arms, reader seed 0) extensions. These are distinct fitted grids;
Table~\ref{tab:result-identities} and the experiment-specific captions
retain their evaluation populations. Overcooked has three training seeds for
returns and probes at each history budget; its donor intervention remains a
single-checkpoint study (Appendix~\ref{app:overcooked-profile}). Swimmer's three
frozen bases span separate architectures.

CoPhy development reporting includes reader-selection examples; CoPhyNet
Balls also uses its source-selection cohort. The held-out Collision episode
comparison uses a separate population. Recipient-bootstrap intervals describe
variation across recipients conditional on fitted source/readers; source SD
describes fitted-model variation.

\label{app:provenance-boundaries}\label{app:new-result-integration}

Public datasets and released implementations retain
their original access and license terms.

\FloatBarrier

\section{Predictive Performance and Closest Comparisons}
\label{app:task-complete}\label{app:records-task}\label{app:external}\label{app:external-environments}

This section expands Tables~\ref{tab:main-task}--\ref{tab:extended-validation}: the primary task comparisons come
first, followed by closest adapted constructions and the released NOD operator.
Each comparison retains its own inputs, readout and evaluation population;
these task effects motivate the representation analyses in
Appendix~\ref{app:formation-complete}.

\paragraph{Primary-comparison average.}
The abstract's 14.6\% is the equal-weight mean of five relative error
reductions in Table~\ref{tab:main-task}: SpringWorld versus Native JEPA
(22.3\%), Collision CPC/RSSM/JEPA versus Structure (15.4\%, 11.4\%, 14.2\%),
and Balls RSSM versus Structure (9.9\%). Calculations use values
before rounding; each setting--learner comparison receives one
weight. This summarizes these five primary comparisons, not all thirteen settings.

\FloatBarrier
\subsection{Cold-start prediction on sealed systems: SpringWorld}
\label{app:spring-sealed}\label{app:records-spring-sealed}
Align + Cross reduces cold-start error relative to Native and TDS on the 256 continuous systems in the primary sealed SpringWorld test. All methods receive an independent
96-frame same-system donor and six cold-start queries per system at $h=16$.
The final 10,000-update sources and readers are frozen: three source seeds
and three readers per source for each recipe. Queries average within system,
then readers within source, then sources equally. Paired 95\% intervals use
10,000 stratified physical-system draws, conditional on these fitted models.

All three source means and all nine paired cells favor SPRII against each
comparator. The 292-system pool weights systems rather than strata equally;
it does not replace the continuous primary endpoint.

\begin{table}[!htbp]\centering\small
\caption{\textbf{Sealed SpringWorld prediction and paired contrasts.}
Results average three sources and three readers per source. Brackets show
95\% physical-system intervals conditional on these fitted models.
Positive gaps favor Align + Cross; the system-weighted pooled result is
auxiliary.}
\label{tab:v3-spring-sealed}\label{tab:spring-sealed-primary}\label{tab:spring-sealed-populations}
\begin{tabular}{@{\hspace{3.5pt}}lrrr@{\hspace{3.5pt}}}\toprule
\textbf{A. Absolute prediction error} &  &  & \\\addlinespace[2pt]
\textbf{Source} & \textbf{Primary: 256 systems} & \textbf{Secondary: 36} & \textbf{Pooled: 292}\\\tableheadrule
Native & 0.13820 $[0.11850, 0.15920]$ & 0.21370 & 0.14750\\
TDS & 0.13030 $[0.11180, 0.15020]$ & 0.19670 & 0.13840\\
\rowcolor{KeyRowTint}
\textbf{Align + Cross} & \textbf{0.10740} $[0.09036, 0.12670]$ & \textbf{0.12730} & \textbf{0.10990}\\\bottomrule
\end{tabular}\par\smallskip
\begin{tabular}{@{\hspace{3.5pt}}llrr@{\hspace{3.5pt}}}\toprule
\multicolumn{4}{@{}l}{\textbf{B. Paired contrasts within each population}}\\\addlinespace[2pt]
\textbf{Population} & \textbf{Comparator} & \textbf{MSE gap [95\% interval]} & \textbf{Reduction [95\% interval]}\\\tableheadrule
Primary & Native & 0.03084 $[0.01735, 0.04429]$ & 22.31\% $[13.15, 30.69]$\\
Primary & TDS & 0.02284 $[0.01039, 0.03487]$ & 17.54\% $[8.40, 25.87]$\\
Secondary & Native & 0.08645 $[0.02195, 0.16640]$ & 40.45\% $[16.34, 54.55]$\\
Secondary & TDS & 0.06944 $[0.01587, 0.13300]$ & 35.30\% $[12.10, 49.13]$\\
Pooled & Native & 0.03769 $[0.02285, 0.05302]$ & 25.55\% $[16.90, 33.32]$\\
Pooled & TDS & 0.02859 $[0.01560, 0.04176]$ & 20.65\% $[12.00, 28.45]$\\\bottomrule
\end{tabular}\end{table}

\FloatBarrier
\subsection{Held-out prediction: Collision and Balls}
\label{app:cophy-collision-test}\label{app:records-collision}
SPRII lowers error across CPC, RSSM and JEPA, improving on Native,
Structure and Random in every fitted seed. Reductions relative to Structure
are 15.4\%, 11.4\% and 14.2\%, respectively
(Table~\ref{tab:v3-collision}). Three independent donor histories and the
first three query frames predict twelve future frames under the S3/query3
task. A frozen donor plan admits 1,994 of 2,000 official normal-test episodes;
six lack three distinct legal donors in one object slot.

All four arms cover three joint source/reader seeds. CPC and RSSM use
source-100 checkpoints; every JEPA arm uses source-50 checkpoints. Readers use a 100-epoch budget. This Collision normal-test comparison uses the above donor eligibility
and checkpoint selection in Appendix~\ref{app:interfaces}.

\begin{figure}[!htbp]\centering
\includegraphics[width=\linewidth]{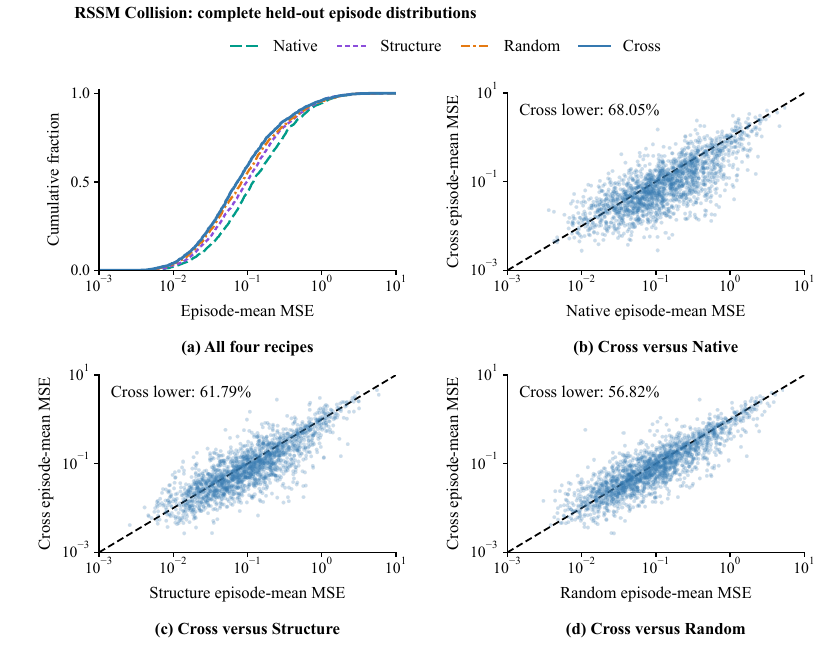}
\caption{\textbf{Cross lowers RSSM error on most held-out episodes.}
(a) Error distributions on 1,994 episodes. (b)--(d) Paired comparisons with
each control. Episode MSEs are averaged over three source/reader runs.
Points below the diagonal favor Cross; fractions report their share.
Axes are logarithmic.}
\label{fig:visual-collision-pairs}
\end{figure}
\begin{table}[!htbp]\centering\small
\caption{\textbf{Held-out Collision prediction.}
Physical trajectory MSE on 1,994 eligible episodes, mean $\pm$ sample SD
across three joint source/reader seeds. Training budgets match within each
learner family.}
\label{tab:v3-collision}\label{tab:cophy-collision-test-grid}
\label{tab:cophy-collision-test-contrasts}\label{tab:jepa-collision-followup}
\setlength{\tabcolsep}{4pt}
\begin{tabular}{@{\hspace{3.5pt}}lccccrr@{\hspace{3.5pt}}}\toprule
 & \multicolumn{4}{c}{\textbf{Physical trajectory MSE}} & \multicolumn{2}{c}{\textbf{Reduction}}\\
\cmidrule(lr){2-5}\cmidrule(l){6-7}
\textbf{Family} & \textbf{Native} & \textbf{Structure} & \textbf{Random} & \textbf{SPRII} & \textbf{\shortstack{vs.\\Structure}} & \textbf{\shortstack{vs.\\Random}}\\\tableheadrule
CPC & $0.248\sdev{0.008}$ & $0.223\sdev{0.006}$ & $0.221\sdev{0.006}$ & $\mathbf{0.189}\sdev{0.007}$ & 15.4\% & 14.7\%\\
RSSM & $0.268\sdev{0.007}$ & $0.228\sdev{0.005}$ & $0.222\sdev{0.007}$ & $\mathbf{0.202}\sdev{0.007}$ & 11.4\% & 9.0\%\\
JEPA & $0.250\sdev{0.004}$ & $0.215\sdev{0.007}$ & $0.265\sdev{0.003}$ & $\mathbf{0.184}\sdev{0.005}$ & 14.2\% & 30.4\%\\
\bottomrule\end{tabular}
\end{table}

Table~\ref{tab:followup-complete-summary} reports
additional CoPhy configuration comparisons.
\paragraph{Held-out Balls4 prediction with RSSM.}
\label{app:balls-heldout}
On the same 1,000 four-ball recipients and eligibility mask, all four RSSM
arms use 100 source epochs and 100 reader epochs, with three independent
source/reader seeds. The existing checkpoints were evaluated after a
12-cell development-consistency check; this is an extension of the existing
test evaluation, not a newly sealed population. Cross lowers mean error by
10.7\% against Native, 9.9\% against Structure and 13.1\% against Random,
with the same direction in all three seeds. This result concerns four-ball
trajectory prediction, not extrapolation to different ball counts.
\begin{table}[!htbp]\centering\small
\caption{\textbf{RSSM on 1,000 held-out Balls4 recipients.} MSE: mean $\pm$ sample SD over three seeds.}
\label{tab:balls-heldout}
\begin{tabular}{@{\hspace{3.5pt}}lcc@{\hspace{3.5pt}}}\toprule
\textbf{Method} & \textbf{MSE} & \textbf{Source / reader epochs}\\\tableheadrule
Native & $1.292\sdev{0.037}$ & 100 / 100\\
Structure & $1.280\sdev{0.031}$ & 100 / 100\\
Random & $1.327\sdev{0.011}$ & 100 / 100\\
\sprii{} (Cross) & $1.154\sdev{0.025}$ & 100 / 100\\
\bottomrule\end{tabular}\end{table}

\FloatBarrier
\subsection{Matched comparisons with adapted constructions}
\label{app:closest-baselines}\label{app:records-closest}
SPRII gives lower matched-reader error than the closest constructions under
the task interfaces specified in Appendix~\ref{app:interfaces}. Tables below
group results by reader and evaluation population.

\paragraph{SpringWorld trajectory-decoupled prediction.}
\label{app:tds-spring}
SPRII lowers matched cold-start error relative to both full and compact visual
TDS under the original SpringWorld reader.
The NOD-inspired TDS adaptation predicts recipient visual embeddings from a
same-system donor, current input and future actions, using
$\mathcal L_{\rm cross}+0.02\mathcal L_{\rm SIGReg}$ and no self or alignment
term. It receives the same 10,000 source updates and independent 96-frame
histories as SPRII. Compact TDS selects a zero-dimensional history slot ($d=0$) from
$0,\ldots,6$ on a separate selection-development bank. A common original physical reader evaluates three
sources by three readers on 100 development systems.

Pooled Matched reductions are 29.73\% and 34.40\%, respectively, with the
same direction in all three source means. This visual TDS construction uses
the trajectory-decoupled principle of \citet{chen2026nod}.

\paragraph{Fixed-recipe temporal FCRL-style comparison.}
\label{app:fcrl-spring}
The fixed temporal FCRL-style adaptation realizes little history gain, whereas
SPRII lowers matched error under the common-adapter comparison.
This temporal adaptation of same-function contrastive learning
\citep{gondal2021fcrl} uses two independent histories per system, a shared
visual/history encoder and a discarded contrastive projection. Temperature
$0.07$ is fixed without a sweep. Each of three sources uses 10,000 updates;
three fresh readers follow the common-adapter recipe. Each method retains
its own frozen query encoder, so the absolute comparison is between complete
source recipes.

SPRII lowers Matched error by 58.44\% [52.90,63.74]\% under this adaptation
and budget. Training-only physical probes remain near zero for the contrastive
codes on the separate 178-system mixture.

\paragraph{D-Clean state-space TDS.}
\label{app:tds-dclean}
SPRII has lower h32 common-reader error than state-space TDS even though both
frozen sources retain strong drag accessibility.
State-space TDS predicts true state increments from independent donor history,
current state and actions at horizons $1,4,16$. A separate source selects
$d=4$ from $0,\ldots,4$ using native h16 error. Three sources by three readers
then evaluate 100 report systems. At h32, common-reader raw-state Matched MSE
is 0.00015072 for SPRII and 0.00134349 for TDS (88.78\% lower); the paired gap
is 0.00119278 [0.00046947,0.00223990], with five-comparison adjusted interval
[0.00038066,0.00267765]. All source means agree in direction.
TDS retains drag $R^2=0.9712$--$0.9767$ (SPRII: 0.9965--0.9967), so larger task
error does not mean absent physical information. Native direct h16 errors
(0.00026174--0.00028337) and iterative h32 errors (0.00182717--0.00198696) use
different input constructions. This 100-system
comparison does not share the population of the 200-system same-donor
experiment in Appendix~\ref{app:horizon}. The separate released-component comparison below uses the same report population.

\paragraph{D-Clean: common-reader dynamics baselines.}
\label{app:dclean-official}
SPRII lowers long-horizon error relative to NOD, FCRL, CaDM and the DALI context adaptation
(Table~\ref{tab:dclean-all-budgets}). CaDM learns a context encoder through
forward and backward dynamics prediction \citep{lee2020cadm}; SPRII reduces its h16 and
h32 errors by 32.16\% and 33.70\%, respectively. Both reductions hold across all three sources. The h32
reduction against FCRL is 88.70\%, also with the same direction in every source.
The implementation and common-reader protocol are specified in
Appendix~\ref{app:interfaces}.

The same source checkpoints support factor-accessibility and fixed-reader
use measurements. Training-fitted damping probes attain
$R^2_{\log\gamma}=0.9739$ for CaDM and $0.9962$ for SPRII. Their h32
wrong-minus-matched MSEs are $0.05329$ and $0.05305$, respectively:
both fitted readers use the supplied history.

The DALI comparison adapts its context-learning component to the same frozen-code
common-reader interface; it is not a native control-return ranking. The
selected 160,000-update source uses a separate 100-system selection bank;
all three source seeds and their three 20,000-update readers are included on
the 100-system report bank. Other sources use 20,000 updates. SPRII lowers
h32 mean error by 96.7\% relative to this longer-budget DALI source, in the
same direction for all three sources. The 20k/80k/160k DALI rows in
Table~\ref{tab:dclean-all-budgets} retain the budget dependence; unequal source
budgets are not described as equal total computation. At 160k, DALI's
damping probe has $R^2=0.785$ and its h32 wrong-minus-matched MSE is 0.0398,
so its larger prediction error does not mean absent context information or use.
\begin{table}[!htbp]\centering\small
\caption{\textbf{D-Clean prediction with a common reader.}
Raw-state MSE in units of $10^{-4}$, mean $\pm$ source SD after averaging
three readers per source ($3$ sources). Source budgets vary as shown.
The 100-system report population was previously evaluated; it is not a new
sealed test.}
\label{tab:dclean-all-budgets}
\setlength{\tabcolsep}{4pt}
\begin{tabular}{@{\hspace{3.5pt}}lrcccc@{\hspace{3.5pt}}}\toprule
\textbf{Source} & \textbf{Updates} & \textbf{h1} & \textbf{h4} & \textbf{h16} & \textbf{h32}\\\tableheadrule
SPRII & 20,000 & $0.0314\sdev{0.002915}$ & $0.1166\sdev{0.005082}$ & $0.5232\sdev{0.01173}$ & $1.507\sdev{0.1355}$\\
CaDM & 20,000 & $0.03324\sdev{0.001838}$ & $0.1474\sdev{0.008051}$ & $0.7711\sdev{0.008115}$ & $2.273\sdev{0.06284}$\\
FCRL & 20,000 & $0.1228\sdev{0.01122}$ & $0.762\sdev{0.08086}$ & $4.833\sdev{0.492}$ & $13.34\sdev{1.014}$\\
NOD & 20,000 & $0.3894\sdev{0.6261}$ & $3.814\sdev{6.412}$ & $23.79\sdev{40.18}$ & $61.88\sdev{104.1}$\\
DALI & 20,000 & $0.5579\sdev{0.02027}$ & $5.307\sdev{0.2763}$ & $35.49\sdev{1.735}$ & $89.3\sdev{3.886}$\\
DALI & 80,000 & $0.3314\sdev{0.035}$ & $2.962\sdev{0.2441}$ & $20.54\sdev{1.948}$ & $53.9\sdev{5.067}$\\
DALI & 160,000 & $0.3201\sdev{0.03572}$ & $2.732\sdev{0.3115}$ & $17.26\sdev{1.991}$ & $45.92\sdev{5.362}$\\
\bottomrule\end{tabular}\end{table}

\begingroup\raggedright
\paragraph{Source dispersion and native controls.}
For SPRII/NOD/FCRL/CaDM, h32 source SD is
0.135/104.076/1.014/0.063 in $10^{-4}$ raw-state MSE units.
Damping-probe $R^2$ is 0.996/0.643/0.852/0.974 and fixed-reader
wrong-minus-matched MSE is 0.05305/0.03556/0.04891/0.05329.
\par\endgroup

\begin{table}[!htbp]\centering\small
\caption{\textbf{Prediction with adapted baselines and context controls.}
A/B report Spring standardized MSE; C reports D-Clean raw-state MSE, each
averaged equally over three sources and three readers. Fixed Wrong uses
the Matched reader. D uses native TDS decoders rather than C's common reader;
its direct and rollout errors are separate diagnostics.}
\label{tab:v3-tds-spring}\label{tab:v3-fcrl}\label{tab:external-dclean}\label{tab:closest-spring}\label{tab:spring-fixed-baseline}\label{tab:closest-dclean}\label{tab:dclean-native-rollout}\label{tab:external-dclean-records}
\textbf{A. Spring: original reader}\par\smallskip
\begin{tabular}{@{\hspace{3.5pt}}llrrr@{\hspace{3.5pt}}}\toprule
\textbf{Reader} & \textbf{Source} & \textbf{Fresh Null} & \textbf{Matched} & \textbf{Fixed Wrong}\\\tableheadrule
Original & Align + Cross & 0.1775 & 0.1121 & 0.2439\\
Original & TDS & 0.1716 & 0.1596 & 0.1787\\
Original & Compact TDS ($d=0$) & 0.1709 & 0.1709 & 0.1709\\
\bottomrule
\end{tabular}
\par\smallskip\textbf{B. Spring: common adapter}\par\smallskip
\begin{tabular}{@{\hspace{3.5pt}}llrrr@{\hspace{3.5pt}}}\toprule
\textbf{Reader} & \textbf{Source} & \textbf{Fresh Null} & \textbf{Matched} & \textbf{Fixed Wrong}\\\tableheadrule
Common adapter & Align + Cross & 0.1778 & 0.1136 & ---\\
Common adapter & FCRL-style & 0.2767 & 0.2735 & ---\\\bottomrule
\end{tabular}
\par\medskip\textbf{C. D-Clean: common-reader h32 context controls}\par\smallskip
\begin{tabular}{@{\hspace{3.5pt}}lrrrr@{\hspace{3.5pt}}}\toprule
\textbf{Source} & \textbf{Matched} & \textbf{Fresh Null} & \textbf{Fixed Wrong} & \textbf{Fixed Zero}\\\tableheadrule
SPRII & 0.0001507 & 0.01821 & 0.05320 & 0.03105\\
State-space TDS & 0.0013430 & 0.01821 & 0.05257 & 0.03143\\
\bottomrule
\end{tabular}
\par\smallskip Other source comparisons and budgets are in Table~\ref{tab:dclean-all-budgets}.
\par\medskip\textbf{D. D-Clean: TDS native-decoder diagnostics}\par\smallskip
\begin{tabular}{@{\hspace{3.5pt}}lrrr@{\hspace{3.5pt}}}\toprule
\textbf{Measurement} & \textbf{Source 0} & \textbf{Source 1} & \textbf{Source 2}\\\tableheadrule
Direct h16 & 0.0002706 & 0.0002834 & 0.0002617\\
Free rollout h32 & 0.001948 & 0.001827 & 0.001987\\\bottomrule
\end{tabular}
\end{table}

\label{app:records-spring-original}

\FloatBarrier
\subsection{Prediction and continuation on a released operator: Burgers}
\label{app:external-nod}\label{app:records-nod}
The fixed SPRII endpoint lowers native prediction error relative to NOD
across ID, viscous OOD and inviscid OOD. We use the released NOD
implementation and training configuration \citep{chen2026nod}, with source
seeds 1234, 5678 and 9012. The field environment and available viscosity
grid are specified in Appendix~\ref{app:nod-burgers-profile} and
Figure~\ref{fig:nod-environments}(a).

\paragraph{Training and selection.}
SPRII retains the native prediction objective during relation training,
then continues prediction with $\lambda_{\rm align}=0$. Zero-alignment
training was checked against the unchanged NOD path using matched seeds,
batches and initialization. Both source searches span 200 epochs and
select checkpoints by ID development loss. The SPRII pre-continuation checkpoints receive
another 25, 113 and 127 epochs, respectively, at 45 updates per epoch;
the final 25 epochs use learning rate $5\times10^{-5}$. All reported SPRII
values use these fixed final endpoints. Continued NOD starts from its own
selected checkpoints and receives the same additional updates. This control
matches continuation updates and does not equate total ancestral training
and search budgets.

\paragraph{Native prediction.}
SPRII reduces error by 1.075\%, 6.554\% and 2.921\% across the three
distributions (Table~\ref{tab:v3-nod-native}). Native MSE follows the
released evaluator's unweighted mean of batch MSE over the full trajectory,
with 45 ID, 500 viscous-OOD and 100 inviscid-OOD examples. These populations
were used in earlier comparisons; the recipe was selected on development
data. Source SD and paired-source intervals accompany the table.

\begin{table}[!htbp]
\centering\small
\setlength{\tabcolsep}{4pt}
\caption{\textbf{Burgers prediction and continuation controls.}
Native trajectory MSE in units of $10^{-3}$, mean $\pm$ sample SD across
three sources. SPRII includes prediction-only continuation; controls match
its added updates. Released-default NOD has no added updates.
Dashes denote unreported source SD.}
\label{tab:v3-nod-native}\label{tab:nod-protocol-record}\label{tab:external-burgers-control}
\begin{tabular}{@{\hspace{3.5pt}}lccc@{\hspace{3.5pt}}}
\toprule
\textbf{Track} & \textbf{ID} & \textbf{Viscous OOD} & \textbf{Inviscid OOD}\\
\tableheadrule
\multicolumn{4}{@{}l}{\emph{Primary comparison}}\\
Released-default NOD & \msd{9.096}{0.544} & \msd{8.012}{0.815} & \msd{11.86}{0.628}\\
\rowcolor{KeyRowTint}
\textbf{SPRII fixed endpoint} & \msd{\mathbf{8.999}}{0.798} & \msd{\mathbf{7.487}}{0.183} & \msd{\mathbf{11.51}}{0.360}\\
\addlinespace[3pt]
\multicolumn{4}{@{}l}{\emph{Matched additional-update controls}}\\
Continued NOD & \msd{9.379}{0.340} & \msd{7.408}{0.130} & \msd{11.49}{0.215}\\
Random continuation & $9.538\ (\text{--})$ & $8.900\ (\text{--})$ & $11.50\ (\text{--})$\\
\bottomrule
\end{tabular}
\begin{papertablenotes}
Paired-source 95\% $t$ intervals for SPRII minus released-default NOD
($n=3$, df=2; MSE $\times10^{-3}$): ID $[-2.472, 2.277]$,
viscous OOD $[-2.982, 1.932]$, and inviscid OOD $[-2.783, 2.091]$.
\end{papertablenotes}
\end{table}

Relative to continued NOD, SPRII lowers ID error by 4.06\%, with
1.06\%/0.20\% higher viscous/inviscid OOD errors. Relative to Random
continuation, it lowers ID and viscous-OOD errors, while inviscid error
is nearly tied. The complete three-source controls appear in the same table.

\paragraph{Organization and fixed-predictor use.}
The primary-checkpoint measurements in Table~\ref{tab:external-burgers-probes}A
use training-standardized latent geometry, training-fitted viscosity probes
and a common history bank. SPRII has a higher between/within ratio and
viscosity probe $R^2$ than released-default NOD. At direct horizon $H=50$,
the mean wrong-donor MSE changes are close to zero for both models.
The additional-update analysis in panel B uses a distinct 11-system,
relative-$L_2$ donor assay: both models incur positive wrong-donor penalties,
and continued NOD has stronger geometry and viscosity accessibility.
These measurements separate accessible viscosity from the response of a
fixed predictor to donor substitution.

\begin{table}[!htbp]
\centering\small\setlength{\tabcolsep}{4pt}
\caption{\textbf{Burgers organization, viscosity accessibility and donor use.}
Entries are mean $\pm$ SD across three sources. B/W is the latent
between/within squared-distance ratio. A uses 45 ID recipients and a direct
MSE donor assay; B uses training geometry and a separate 11-system
relative-$L_2$ assay. Positive penalties indicate harm from wrong donors;
95\% intervals resample paired sources and systems.}
\label{tab:nod-mechanism}\label{tab:nod-donor-use}\label{tab:nod-donor-controls}\label{tab:external-burgers-probes}
\textbf{A. Primary comparison: released NOD and SPRII}\par\smallskip
\begin{tabular}{@{\hspace{3.5pt}}lcccc@{\hspace{3.5pt}}}\toprule
 & \textbf{ID B/W} & \textbf{ID $R^2$} & \multicolumn{2}{c}{\textbf{$H=50$ MSE ($\times10^{-3}$)}}\\
\cmidrule(l){4-5}
\textbf{Method} &  &  & \textbf{Matched} & \textbf{Wrong$-$matched}\\\tableheadrule
NOD & \msd{22.3}{19.09} & \msd{0.651}{0.458} & \msd{12.89}{0.992} & \msd{0.014}{0.391}\\
\rowcolor{KeyRowTint}
\textbf{SPRII} & \msd{\mathbf{31.64}}{4.506} & \msd{\mathbf{0.762}}{0.141} & \msd{\mathbf{12.19}}{1.944} & \msd{\mathbf{-0.054}}{0.597}\\\bottomrule
\end{tabular}
\par\medskip\textbf{B. Additional-update comparison}\par\smallskip
\begin{tabular}{@{\hspace{3.5pt}}lrrrc@{\hspace{3.5pt}}}\toprule
\textbf{Method} & \textbf{Train B/W} & \textbf{ID $R^2$} & \textbf{\shortstack{Viscous-OOD\\$R^2$}} & \textbf{\shortstack{$H=50$ relative-$L_2$ penalty\\{[95\% interval]}}}\\\tableheadrule
Continued NOD & \msd{42.07}{14.74} & \msd{0.924}{0.006} & \msd{0.975}{0.004} & 0.1331 $[0.06729, 0.2127]$\\
SPRII & \msd{24.58}{8.62} & \msd{0.762}{0.141} & \msd{0.895}{0.116} & 0.1275 $[0.06538, 0.2041]$\\\bottomrule
\end{tabular}
\end{table}

\label{app:records-nod-checkpoints}

\paragraph{Test-time-adaptive operator references.}
GEPS and CoDA use 50 support-code optimization steps for each query, while
NOD and the SPRII endpoint use none. Table~\ref{tab:burgers-adaptation-cost}
retains those completed references under the released native metric.
CoDA has lower mean error in all three distributions; GEPS has lower ID
error and higher OOD means, with substantial source dispersion. These
comparisons do not establish equal inference cost or uniform dominance.
\begin{table}[!htbp]\centering\small
\caption{\textbf{Burgers prediction with test-time adaptation.}
Native full-trajectory MSE in units of $10^{-3}$, mean $\pm$ sample SD across
three sources. Both methods use 50 support-code updates per query;
zero-step results are in Table~\ref{tab:v3-nod-native}.}
\label{tab:burgers-adaptation-cost}
\begin{tabular}{@{\hspace{3.5pt}}lcccc@{\hspace{3.5pt}}}\toprule
\textbf{Method} & \textbf{ID} & \textbf{Viscous OOD} & \textbf{Inviscid OOD} & \textbf{Code steps}\\\tableheadrule
GEPS & $5.60\sdev{7.05}$ & $12.37\sdev{14.21}$ & $13.84\sdev{17.41}$ & 50\\
CoDA & $3.38\sdev{0.57}$ & $6.48\sdev{1.42}$ & $5.89\sdev{1.55}$ & 50\\\bottomrule
\end{tabular}\end{table}

\FloatBarrier
\subsection{Relation supervision on a frozen CoDA host}
\label{app:coda-sprii}
An amortized CoDA instance adds a history encoder to each of three fixed
CoDA source decoders. The four arms in Table~\ref{tab:coda-host-design}
share a teacher-fit initialization, a fixed encoder budget and final-checkpoint
selection. Each source seed retains its own decoder and train-only
teacher-code cache. Seeds 1234, 5678 and 9012 retain their completed
5,000-update decoders. All arms use the same 101-frame support histories and
fixed evaluation manifest: 45 ID, 500 viscous-OOD and 100 inviscid-OOD pairs.
The relation-trained arm uses Cross.

\begin{table}[!htbp]\centering\small
\caption{\textbf{Objectives on the frozen CoDA decoder.}}
\label{tab:coda-host-design}
\begin{tabularx}{\linewidth}{@{\hspace{3.5pt}}l>{\raggedright\arraybackslash}X>{\raggedright\arraybackslash}X@{\hspace{3.5pt}}}\toprule
\textbf{Arm} & \textbf{Objective addition} & \textbf{Role}\\\tableheadrule
Plain amortized CoDA & Self prediction and common regularization & History-encoder baseline.\\
Matched Cross & A second prediction using the related donor & Relation-supervised arm.\\
Random Cross & The same second prediction using a wrong-system donor & Pairing control with matched donor marginals.\\
Duplicate Self & A second self-prediction forward/backward pass & Matched computation and reconstruction-loss scale.\\\bottomrule
\end{tabularx}\end{table}

A shared 500-update teacher fit precedes 1,000 encoder updates per arm.
Within each seed, initialization and sampled histories are identical across
arms. Plain uses one prediction branch; the other arms use two, so equal
update counts do not imply equal training FLOPs.
The primary endpoint uses the inferred context with no support-code
optimization (code-0). The auxiliary code-50 endpoint initializes the code
from the same encoder and applies fifty support-code optimization steps;
the encoder and decoder remain frozen.
The prespecified code-0 criterion compares Matched Cross with both Plain
and Duplicate Self in each distribution: lower mean error and the same
direction in a majority of source seeds. Random Cross measures sensitivity
to the relation assignment. Paired contrasts use the same decoder seed,
evaluation inputs and adaptation budget.
Native MSE averages batch-8 errors equally and includes the initial frame.
Viscosity ridge probes fit 360 training histories, with regularization selected
by five-fold cross-validation over training cases.

Matched Cross meets this code-0 criterion in all three distributions:
relative to Plain, mean error falls by 6.70\%, 5.17\% and 6.50\%; relative
to Duplicate Self, it falls by 4.98\%, 6.74\% and 7.39\%, respectively.
Each comparison favors Matched Cross in all three seeds. Its smaller
advantages over Random Cross (1.74\%, 0.38\% and 1.46\%) have directions
2/3, 1/3 and 2/3 across seeds. With only three source runs, all nine paired
$t$ intervals include zero (Table~\ref{tab:coda-host-contrasts-0}); these
results do not establish a stable advantage of correct over random pairing.

After fifty support-code updates, Matched Cross has lower mean error than
Duplicate Self in all three distributions, but lower mean error than Plain
only on viscous OOD; Random Cross has slightly lower means than Matched
Cross in all three distributions.
All nine paired intervals include zero
(Table~\ref{tab:coda-host-contrasts-50}). Each arm's mean query error is
higher at code-50 than at code-0 in every distribution. Matched Cross retains the highest mean
viscosity-probe $R^2$ (Table~\ref{tab:coda-host-probes}), separating
parameter accessibility from forecasting value under this fixed adaptation
budget. The predictive gains reported here concern amortized inference.

\begin{table}[!htbp]\centering\small
\caption{\textbf{Amortized prediction with a frozen CoDA decoder.}
Native full-trajectory MSE in units of $10^{-3}$, mean $\pm$ sample SD across
three sources. Code-0 is the primary endpoint.}
\label{tab:coda-host-results}
\begin{tabular}{@{\hspace{3.5pt}}llcc@{\hspace{3.5pt}}}\toprule
\textbf{Distribution} & \textbf{Arm} & \textbf{Code-0} & \textbf{Code-50}\\\tableheadrule
ID & Plain & $3.4834\sdev{0.5669}$ & $\mathbf{3.5123}\sdev{0.5353}$\\
ID & Matched Cross & $\mathbf{3.2500}\sdev{0.7286}$ & $3.5566\sdev{0.4909}$\\
ID & Random Cross & $3.3076\sdev{0.7055}$ & $3.5164\sdev{0.4933}$\\
ID & Duplicate Self & $3.4203\sdev{0.7573}$ & $3.5899\sdev{0.4809}$\\
\addlinespace[2.5pt]
Viscous OOD & Plain & $6.3900\sdev{1.4357}$ & $7.1890\sdev{1.6605}$\\
Viscous OOD & Matched Cross & $\mathbf{6.0598}\sdev{1.1847}$ & $7.0845\sdev{1.4797}$\\
Viscous OOD & Random Cross & $6.0832\sdev{1.2241}$ & $\mathbf{7.0310}\sdev{1.5672}$\\
Viscous OOD & Duplicate Self & $6.4976\sdev{1.5413}$ & $7.4010\sdev{1.6684}$\\
\addlinespace[2.5pt]
Inviscid OOD & Plain & $5.8833\sdev{1.6998}$ & $\mathbf{6.7663}\sdev{2.2725}$\\
Inviscid OOD & Matched Cross & $\mathbf{5.5006}\sdev{1.4956}$ & $6.8105\sdev{2.1187}$\\
Inviscid OOD & Random Cross & $5.5822\sdev{1.5558}$ & $6.8046\sdev{2.2922}$\\
Inviscid OOD & Duplicate Self & $5.9394\sdev{1.8774}$ & $7.0893\sdev{2.5458}$\\
\bottomrule\end{tabular}\end{table}
\FloatBarrier

\begin{table}[!htbp]\centering\small
\caption{\textbf{Paired CoDA contrasts at code-0.}
Differences are comparator minus Matched Cross in units of $10^{-3}$;
positive favors Matched Cross. Intervals are paired-source 95\% $t$
intervals ($n=3$); $n_+$ counts positive source differences.}
\label{tab:coda-host-contrasts-0}
\setlength{\tabcolsep}{4pt}
\begin{tabular}{@{\hspace{3.5pt}}llccccc@{\hspace{3.5pt}}}\toprule
\textbf{Distribution} & \textbf{Comparator} & \textbf{Seed 1} & \textbf{Seed 2} & \textbf{Seed 3} & \textbf{Mean [95\% CI]} & \textbf{$n_+$}\\\tableheadrule
ID & Plain & 0.1270 & 0.4228 & 0.1502 & $0.2333\,[-0.1752, 0.6418]$ & 3\\
ID & Random Cross & $-0.0113$ & 0.0710 & 0.1129 & $0.0576\,[-0.0994, 0.2145]$ & 2\\
ID & Duplicate Self & 0.3036 & 0.1917 & 0.0155 & $0.1703\,[-0.1904, 0.5310]$ & 3\\
\addlinespace[2.5pt]
Viscous OOD & Plain & 0.1989 & 0.7125 & 0.0793 & $0.3302\,[-0.5055, 1.1659]$ & 3\\
Viscous OOD & Random Cross & $-0.0904$ & 0.1676 & $-0.0071$ & $0.0234\,[-0.3038, 0.3506]$ & 1\\
Viscous OOD & Duplicate Self & 0.4304 & 0.8332 & 0.0499 & $0.4378\,[-0.5351, 1.4108]$ & 3\\
\addlinespace[2.5pt]
Inviscid OOD & Plain & 0.2188 & 0.7455 & 0.1836 & $0.3826\,[-0.3993, 1.1646]$ & 3\\
Inviscid OOD & Random Cross & $-0.0237$ & 0.2427 & 0.0257 & $0.0816\,[-0.2704, 0.4335]$ & 2\\
Inviscid OOD & Duplicate Self & 0.2938 & 0.9736 & 0.0488 & $0.4387\,[-0.7515, 1.6290]$ & 3\\
\bottomrule\end{tabular}\end{table}
\begin{table}[!htbp]\centering\small
\caption{\textbf{Paired CoDA contrasts at code-50.}
Differences are comparator minus Matched Cross in units of $10^{-3}$;
positive favors Matched Cross. Intervals are paired-source 95\% $t$
intervals ($n=3$); $n_+$ counts positive source differences.}
\label{tab:coda-host-contrasts-50}
\setlength{\tabcolsep}{4pt}
\begin{tabular}{@{\hspace{3.5pt}}llccccc@{\hspace{3.5pt}}}\toprule
\textbf{Distribution} & \textbf{Comparator} & \textbf{Seed 1} & \textbf{Seed 2} & \textbf{Seed 3} & \textbf{Mean [95\% CI]} & \textbf{$n_+$}\\\tableheadrule
ID & Plain & 0.0046 & $-0.0930$ & $-0.0446$ & $-0.0443\,[-0.1655, 0.0768]$ & 1\\
ID & Random Cross & $-0.0361$ & $-0.0165$ & $-0.0681$ & $-0.0402\,[-0.1049, 0.0244]$ & 0\\
ID & Duplicate Self & 0.0232 & 0.0760 & 0.0006 & $0.0333\,[-0.0629, 0.1294]$ & 3\\
\addlinespace[2.5pt]
Viscous OOD & Plain & 0.1935 & 0.2277 & $-0.1076$ & $0.1045\,[-0.3537, 0.5628]$ & 2\\
Viscous OOD & Random Cross & $-0.0976$ & 0.1018 & $-0.1646$ & $-0.0534\,[-0.3977, 0.2908]$ & 1\\
Viscous OOD & Duplicate Self & 0.3286 & 0.5325 & 0.0882 & $0.3165\,[-0.2360, 0.8689]$ & 3\\
\addlinespace[2.5pt]
Inviscid OOD & Plain & 0.0396 & 0.0626 & $-0.2347$ & $-0.0442\,[-0.4551, 0.3668]$ & 2\\
Inviscid OOD & Random Cross & $-0.0369$ & 0.1850 & $-0.1656$ & $-0.0059\,[-0.4464, 0.4347]$ & 1\\
Inviscid OOD & Duplicate Self & 0.0376 & 0.8296 & $-0.0308$ & $0.2788\,[-0.9092, 1.4669]$ & 2\\
\bottomrule\end{tabular}\end{table}
\FloatBarrier

\begin{table}[!htbp]\centering\small
\caption{\textbf{CoDA viscosity accessibility and decoder invariance.}
Probe $R^2$ is mean $\pm$ sample SD across three sources on 45 ID histories.
Decoder change is the maximum absolute change during encoder fitting, over
all three decoders; all remain unchanged.}
\label{tab:coda-host-probes}
\begin{tabular}{@{\hspace{3.5pt}}lccc@{\hspace{3.5pt}}}\toprule
\textbf{Arm} & \textbf{Code-0 viscosity $R^2$} & \textbf{Code-50 viscosity $R^2$} & \textbf{Max. decoder change}\\\tableheadrule
Plain & $0.1325\sdev{0.1388}$ & $0.2765\sdev{0.2184}$ & 0\\
Matched Cross & $\mathbf{0.1752}\sdev{0.1545}$ & $\mathbf{0.3010}\sdev{0.1946}$ & 0\\
Random Cross & $0.1291\sdev{0.1324}$ & $0.2845\sdev{0.1945}$ & 0\\
Duplicate Self & $0.1527\sdev{0.1604}$ & $0.2878\sdev{0.2334}$ & 0\\
\bottomrule\end{tabular}\end{table}

\begin{table}[!htbp]\centering\small
\caption{\textbf{CoDA fitting and evaluation costs.}
Times are seconds, averaged over three sources: per fit for training and
per query for evaluation. Code-0 includes encoder inference and forecasting;
Code-50 rows exclude encoder inference. Evaluation times are amortized
execution costs, excluding data transfer, decoder construction and scoring.
Shared decoder and teacher costs are reported separately.}
\label{tab:coda-host-cost}
\setlength{\tabcolsep}{4pt}
\begin{tabular}{@{\hspace{3.5pt}}lcccc@{\hspace{3.5pt}}}\toprule
\textbf{Quantity} & \textbf{Plain} & \textbf{Matched Cross} & \textbf{Random Cross} & \textbf{Duplicate Self}\\\tableheadrule
Trainable encoder parameters & 89,282 & 89,282 & 89,282 & 89,282\\
Encoder fitting time & 1267.92 & 2489.37 & 2517.80 & 2519.17\\
Code-0 encoder + forecast & 0.03416 & 0.03500 & 0.03510 & 0.03353\\
Code-50 adaptation time & 6.84052 & 6.74525 & 6.57095 & 6.47388\\
Code-50 forecasting time & 0.03490 & 0.03604 & 0.03407 & 0.03370\\
\midrule
Shared decoder fitting time & \multicolumn{4}{c}{5372.08}\\
Shared teacher-code adaptation & \multicolumn{4}{c}{2286.44}\\
Shared teacher-fit time & \multicolumn{4}{c}{1.44}\\\bottomrule
\end{tabular}\end{table}
\FloatBarrier

A separate single-source development comparison used 45 support/query pairs
from nine systems. At zero adaptation steps, Cross reduced mean error by
3.84\% relative to Plain, with all nine systems in the same direction;
it fell below that study's prespecified 5\% improvement threshold.
The four-arm analysis retains random-pairing and duplicate-self controls.
Original CoDA with zero initialization and fifty support optimization steps
is the external reference in Table~\ref{tab:burgers-adaptation-cost}.
The report populations follow the released evaluation protocol; development
pairs serve only the preceding single-source comparison. The completed
four-arm evaluation extends the historical report populations and
is not a new sealed test; its results did not change the fixed recipe or
the earlier development-gate conclusion.

\FloatBarrier
\section{SpringWorld Case Study}
\label{app:spring-case}
This section collects the questions needed to interpret the SpringWorld
instance: what history alone provides, how Align and Cross contribute, and
what training resources and objective weights accompany those effects.
The existing comparisons use the visual input and training protocol in
Appendix~\ref{app:spring-profile}. Development component measurements and
sealed prediction results use separate evaluation populations.

\subsection{History-conditioned baselines}
\label{app:spring-history-baselines}
Native and Structure both receive history. Introducing the separate channel
alone does not reproduce the joint method's benefit: the development component
means are 0.170 for Native, 0.172 for Structure and 0.110 for Align + Cross
(Table~\ref{tab:v3-spring-components}). Appendix~\ref{app:spring-sealed} gives the primary comparison against Native
and trajectory-decoupled prediction on the separate sealed systems.

\subsection{Contributions of Align and Cross}
\label{app:spring-objective-components}
Align accounts for most of the observed organization and task improvement in
the component study. The joint recipe lowers development error to 0.110,
compared with 0.123 for Align and 0.168 for Cross; its source-wise
between/within ratio is 2.360, compared with 2.045 and 0.252, respectively.
Table~\ref{tab:v3-spring-components} gives the full component comparison
and separate task and geometry populations.

\subsection{Computational cost}
\label{app:spring-cost}
Both sources use 10,000 optimizer updates and their frozen-context readers
use 10,000 updates. Align and Cross are added within source updates; this
SpringWorld instance has no continuation stage. Table~\ref{tab:spring-cost}
separates optimizer budgets, parameter counts and measured running time.
\begin{table}[!htbp]\centering\small
\caption{\textbf{SpringWorld training cost per fit.} Update times use a
common RTX 6000D benchmark. Full-fit times average three original runs
under different hardware and runtime conditions and are not directly
comparable to the update benchmark.}
\label{tab:spring-cost}
\begin{tabular}{@{\hspace{3.5pt}}lcc@{\hspace{3.5pt}}}\toprule
\textbf{Quantity} & \textbf{Native} & \textbf{Align + Cross}\\\tableheadrule
Source optimizer updates & 10,000 & 10,000\\
Additional continuation updates & 0 & 0\\
Reader updates per fitted reader & 10,000 & 10,000\\
Trainable source parameters & 4,509,120 & 4,525,632\\
Compute-only time per source update (s) & 0.1756 & 0.1948\\
Original full-fit wall time (s) & 34,270.26 & 34,324.20\\\bottomrule
\end{tabular}
\end{table}

\subsection{Budgets and configuration selection}
\label{app:spring-budget}
Table~\ref{tab:spring-budgets} separates the source budget, reader budget and
selection rule. Three fitted readers on one source are not three independent
source runs. The initial comparisons fix the relation weights at
$\lambda_{\mathrm p}=1$ and $\lambda_{\mathrm x}=0.1$; the sensitivity study
below is a separate development analysis.
\begin{table}[!htbp]\centering\small
\caption{\textbf{SpringWorld budgets and configuration selection.} Updates
are per fit.}
\label{tab:spring-budgets}
\begin{tabularx}{\linewidth}{@{\hspace{3.5pt}}lrrX@{\hspace{3.5pt}}}\toprule
\textbf{Comparison} & \textbf{Source} & \textbf{Reader} & \textbf{Configuration and selection}\\\tableheadrule
Native / Structure / Align / Cross / joint & 10,000 & 10,000 & Fixed component weights; final source step. Component table uses reader 0.\\
Sealed Native / TDS / joint & 10,000 & 10,000 & Three sources $\times$ three readers; sources and readers frozen before the primary test.\\
Full / compact TDS & 10,000 & 10,000 & $d\in\{0,\ldots,6\}$ for compact TDS; separate selection-development bank.\\
Temporal FCRL-style adaptation & 10,000 & 10,000 & Fixed temperature 0.07; three sources $\times$ three common-adapter readers.\\
\bottomrule\end{tabularx}
\end{table}
\FloatBarrier

The TDS and FCRL-style inputs and matched-reader restrictions are specified in
Appendix~\ref{app:tds-spring}. Configuration counts describe distinct recipes
in the reported comparison and its declared selection stage. They exclude
seed replicas, the separate relation-weight sensitivity study and earlier
prototypes. In particular, the current fixed-temperature FCRL adaptation
is distinct from the earlier temperature-selected FCRL experiment.
Native, Structure, Align, Cross, the joint recipe, full TDS and temporal
FCRL each use one fixed source configuration. Compact TDS declares seven
candidate dimensions, $d\in\{0,\ldots,6\}$, for its selection stage.

\subsection{Sensitivity to relation weights}
\label{app:spring-lambda}
The one-at-a-time development analysis varies
$\lambda_{\mathrm p}\in\{0.25,1,4\}$ at $\lambda_{\mathrm x}=0.1$, and
$\lambda_{\mathrm x}\in\{0.025,0.1,0.4\}$ at $\lambda_{\mathrm p}=1$.
The shared default gives five distinct settings, with three source seeds per
setting. Architecture, data, update count and reader protocol are fixed.
Prediction and training-fitted factor probes use the development population;
the sealed primary test is excluded from configuration selection.
All fifteen sources and forty-five readers are included. The prediction
measure is cold-start $q=0$, $H=16$ error on 100 development systems
(600 cases); the probes are evaluated on 178 development systems.

\begin{table}[!htbp]\centering\small
\caption{\textbf{Sensitivity to Align and Cross weights.} Mean $\pm$ SD over
three sources; MSE first averages three readers per source. Probes predict
$\log m$, $\log\gamma$ and $\log k$.}
\label{tab:spring-lambda}
\setlength{\tabcolsep}{5pt}
\begin{tabular}{@{\hspace{3.5pt}}cccccc@{\hspace{3.5pt}}}\toprule
\textbf{$\lambda_{\mathrm p}$} & \textbf{$\lambda_{\mathrm x}$} & \textbf{Dev. MSE} & \textbf{Mass $R^2$} & \textbf{Drag $R^2$} & \textbf{Stiffness $R^2$}\\\tableheadrule
0.25 & 0.1 & $0.1203\sdev{0.0025}$ & $0.2837\sdev{0.0494}$ & $0.1441\sdev{0.1210}$ & $-0.0204\sdev{0.0218}$\\
1 & 0.1 & $0.1100\sdev{0.0081}$ & $0.3630\sdev{0.0408}$ & $0.1598\sdev{0.0220}$ & $0.1430\sdev{0.0284}$\\
4 & 0.1 & $0.1609\sdev{0.0360}$ & $0.3498\sdev{0.0989}$ & $0.2694\sdev{0.1131}$ & $0.3166\sdev{0.2892}$\\
1 & 0.025 & $0.1049\sdev{0.0112}$ & $0.4162\sdev{0.0846}$ & $0.1645\sdev{0.0447}$ & $0.3578\sdev{0.2626}$\\
1 & 0.4 & $0.1274\sdev{0.0027}$ & $0.3751\sdev{0.0322}$ & $0.1331\sdev{0.0277}$ & $0.2403\sdev{0.1694}$\\
\bottomrule\end{tabular}
\end{table}

Relative to the default $(1,0.1)$, changing Align to $0.25$ or $4$, or
Cross to $0.4$, increases mean cold-start error by 9.38\%, 46.23\% and
15.83\%, respectively, with the same direction in all three sources.
Reducing Cross to $0.025$ lowers the mean by 4.69\%, with lower error in two
of three sources.
Probe scores and prediction error do not vary in lockstep: for example,
$\lambda_{\rm p}=4$ raises mean drag and stiffness $R^2$ while increasing
prediction error.

\subsection{Gradient relationships}
\label{app:spring-gradient}
Every 100 optimizer steps, the diagnostic computes the cosine between
$\nabla\mathcal L_{\rm align}$ and $\nabla\mathcal L_{\rm cross}$ on the shared
persistent encoder $g_p$, using the same diagnostic batch and parameter subset.
It records both gradient norms and zero-norm cases, restores RNG state and
training mode, and makes no parameter update. Cosine summaries exclude
zero-norm observations and retain negative values. Early, middle and late
phases use update intervals $(0,10000/3]$, $(10000/3,20000/3]$ and
$(20000/3,10000]$, respectively. These local gradient
measurements describe objective interaction; task performance is measured
separately in Table~\ref{tab:spring-lambda}.

\begin{table}[!htbp]\centering\small
\caption{\textbf{Distribution of Align--Cross gradient cosines.} Quantiles
and negative frequencies use valid observations pooled across three sources;
zero-norm frequencies use all observations.}
\label{tab:spring-gradient-distribution}
\setlength{\tabcolsep}{5pt}
\begin{tabular}{@{\hspace{3.5pt}}cccccccc@{\hspace{3.5pt}}}\toprule
\textbf{$(\lambda_{\rm p},\lambda_{\rm x})$} & \textbf{Valid $n$} & \textbf{Mean} & \textbf{Q1} & \textbf{Median} & \textbf{Q3} & \textbf{Negative (\%)} & \textbf{Zero norm (\%)}\\\tableheadrule
$(0.25,0.1)$ & 300 & 0.0263 & $-0.0152$ & 0.0177 & 0.0720 & 35.0 & 0.0\\
$(1,0.1)$ & 300 & 0.0214 & $-0.0387$ & 0.0181 & 0.0770 & 40.7 & 0.0\\
$(4,0.1)$ & 300 & 0.0246 & $-0.0355$ & 0.0173 & 0.0842 & 40.3 & 0.0\\
$(1,0.025)$ & 300 & 0.0308 & $-0.0194$ & 0.0214 & 0.0765 & 34.3 & 0.0\\
$(1,0.4)$ & 300 & 0.0379 & $-0.0383$ & 0.0317 & 0.1028 & 37.3 & 0.0\\
\bottomrule\end{tabular}
\end{table}

\begin{table}[!htbp]\centering\small
\caption{\textbf{Gradient norms and cosines across sources and training phases.}
Norms are unweighted medians; other columns show mean valid cosines.
Repeated diagnostic observations are not independent source runs.}
\label{tab:spring-gradient-source-phase}
\setlength{\tabcolsep}{4pt}
\begin{tabular}{@{\hspace{3.5pt}}ccccccccc@{\hspace{3.5pt}}}\toprule
\textbf{Weights} & \textbf{$\|\nabla\mathcal L_{\rm align}\|$} & \textbf{$\|\nabla\mathcal L_{\rm cross}\|$} & \textbf{Source 0} & \textbf{Source 1} & \textbf{Source 2} & \textbf{Early} & \textbf{Middle} & \textbf{Late}\\\tableheadrule
$(0.25,0.1)$ & 5.9381 & 0.0086 & 0.0327 & 0.0328 & 0.0135 & 0.0378 & 0.0326 & 0.0092\\
$(1,0.1)$ & 6.3315 & 0.0428 & 0.0208 & 0.0343 & 0.0092 & 0.0270 & 0.0286 & 0.0089\\
$(4,0.1)$ & 5.4435 & 0.0800 & 0.0299 & 0.0182 & 0.0256 & 0.0273 & 0.0260 & 0.0205\\
$(1,0.025)$ & 6.9317 & 0.0501 & 0.0393 & 0.0169 & 0.0363 & 0.0469 & 0.0307 & 0.0153\\
$(1,0.4)$ & 6.6684 & 0.0334 & 0.0297 & 0.0455 & 0.0385 & 0.0152 & 0.0577 & 0.0408\\
\bottomrule\end{tabular}
\end{table}
\FloatBarrier

Mean gradient cosines range from 0.0214 to 0.0379, while 34.3--40.7\% of
observations are negative. No diagnostic observation has a zero gradient
norm. Thus the small positive mean does not imply uniformly aligned
objectives, and these local measurements do not establish task-level synergy.

\subsection{Relation reliability: PokeWorld comparison}
\label{app:spring-reliability}
Appendix~\ref{app:poke-formation} and Figure~\ref{fig:relation-mechanism}
vary the correct-pair mixture weight $\alpha$ in PokeWorld. Proposition 2 in
Appendix~\ref{app:pairing-theory} isolates the corresponding alignment pressure.

\FloatBarrier
\Needspace{12\baselineskip}
\section{Formation: Relation Design and Representation Organization}
\label{app:formation-complete}\label{app:records-formation-group}\label{app:controlled}\label{app:real-data}

PokeWorld, SpringWorld and Baxter test relation design, objective components
and tactile selectivity. Fixed-route Use is measured separately in
Appendix~\ref{app:information-use}.

\FloatBarrier
\subsection{Relation reliability, shared factors, and objectives: PokeWorld}
\label{app:poke-formation}\label{app:pokeworld}\label{app:pairing-results}
\label{app:selectivity}\label{app:components}\label{app:poke-components}
\label{app:revision-protocol}\label{app:factorial}\label{app:acquisition}\label{app:poke-acquisition}
\begin{figure}[H]
\centering
\includegraphics[width=\linewidth]{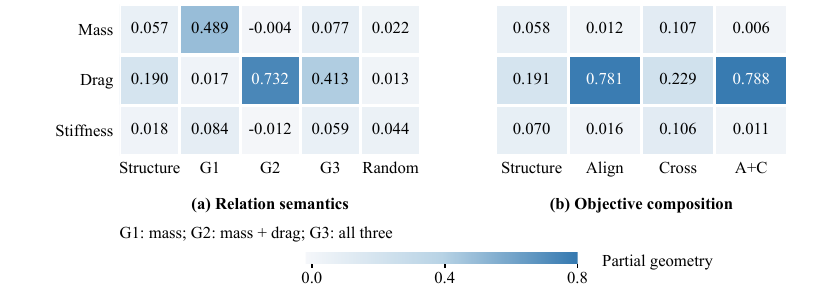}
\caption{\textbf{Relation semantics and objective composition organize different factors.}
Mean partial geometry across three sources on 400 confirmation systems.
(a) Shared factors $G_1=m$, $G_2=(m,\gamma)$ and $G_3=(m,\gamma,k)$;
(b) objective components. Blocks use separate Structure references and a
common color scale. Source SDs are in Tables~\ref{tab:refinement-full}
and~\ref{tab:factorial-confirmation}.}
\label{fig:v3-poke-formation-use}\label{fig:poke-common-use-full}\label{fig:relation-confirmation}\label{fig:factorial-confirmation}
\end{figure}

\takeawaybox{More reliable relations strengthen persistent organization; changing the pairing rule redirects frozen-code accessibility.}
The experiments vary pair reliability at fixed R8, shared factors and objective
composition. Within each study, sources 0/1/2 share architecture and budget;
checkpoints and endpoints are fixed after validation and evaluated on
400 confirmation systems. Figure~\ref{fig:v3-poke-formation-use} shows the two formation maps.

\paragraph{Input evidence and interaction diversity.}\label{app:matched-control}
Rendered histories support factor readout, with higher accuracy for drag than
for mass and stiffness. A separate privileged-dynamics estimator quantifies
recoverability when numerical state is available. Replacing eligible
same-rollout donors with independent rollouts raises drag readout from 0.2647
to 0.3469 while native h16 self loss changes from 0.3641 to 0.3869. Diversity
and reliability answer separate questions: the R2/R4/R8 comparison changes
available realizations, whereas $\alpha$ changes pair quality at fixed R8.
\begin{table}[H]\centering\small
\caption{\textbf{Factor readout from rendered histories and privileged dynamics.}
Rendered-history entries show means and source SD.}
\label{tab:v3-poke-input-diversity}
\begin{tabular}{@{\hspace{3.5pt}}lrrr@{\hspace{3.5pt}}}\toprule
\textbf{Input readout ($R^2$)} & \textbf{Mass} & \textbf{Drag} & \textbf{Stiffness}\\\tableheadrule
Rendered history & \msd{0.270}{0.026} & \msd{0.574}{0.009} & \msd{0.149}{0.034}\\
Privileged dynamics & 0.762 & 0.937 & 0.760\\\bottomrule

\end{tabular}\end{table}

\begin{figure}[!htbp]
\centering
\includegraphics[width=\linewidth]{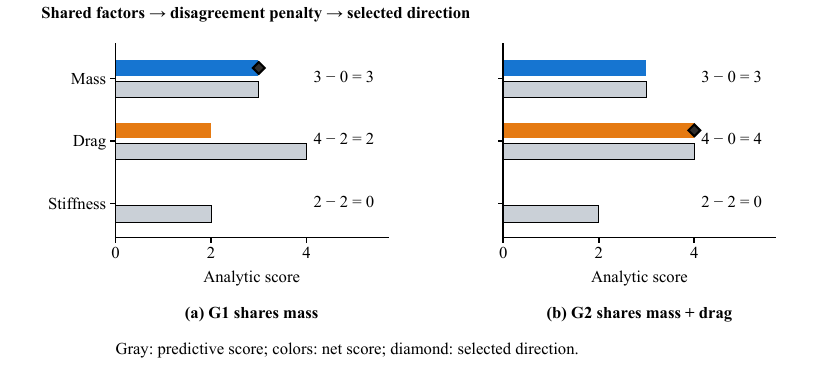}
\caption{\textbf{A broader sharing relation can select a different predictive direction.}
In the Proposition 3 example, broadening (a) $G_1$ to (b) $G_2$ changes the
selected direction (diamond) from mass to drag. Gray bars show predictive
scores; colored bars subtract relation disagreement.}
\label{fig:v3-spectral}
\end{figure}

\paragraph{Shared factors and frozen accessibility.}
Changing from $G_1$ to $G_2$ shifts accessibility from mass toward drag.
Training-only ridge and MLP probes of frozen code centroids both favor $G_1$
for mass and $G_2$ for drag in every source replica on 400 validation systems.
The spectral construction in Figure~\ref{fig:v3-spectral} shows how a wider
sharing relation can change the selected predictive direction without lowering
the original factor's score.
\begin{table}[H]\centering\small
\caption{\textbf{Both linear and nonlinear readouts show the $G_1$/$G_2$ accessibility shift.}
Frozen centroid probes on validation systems. Error contrasts are $G_2$ minus
$G_1$ with paired-system 95\% intervals; positive favors $G_1$.}
\label{tab:v3-poke-accessibility}
\begin{tabular}{@{\hspace{3.5pt}}llrrrr@{\hspace{3.5pt}}}\toprule
\textbf{Probe} & \textbf{Target} & \textbf{$G_1$ $R^2$} & \textbf{$G_2$ $R^2$} & \textbf{MSE contrast} & \textbf{95\% interval}\\\tableheadrule
Ridge & $\log m$ & 0.5271 & 0.0709 & 0.1492 & $[0.1231, 0.1758]$\\
Ridge & $\gamma$ & 0.3075 & 0.9249 & $-0.7704$ & $[-0.8572, -0.6869]$\\
MLP & $\log m$ & 0.5272 & 0.0273 & 0.1635 & $[0.1303, 0.1974]$\\
MLP & $\gamma$ & 0.1291 & 0.9644 & $-1.042$ & $[-1.143, -0.9438]$\\\bottomrule
\end{tabular}\end{table}

\begin{figure}[!htbp]
\centering
\includegraphics[width=\linewidth]{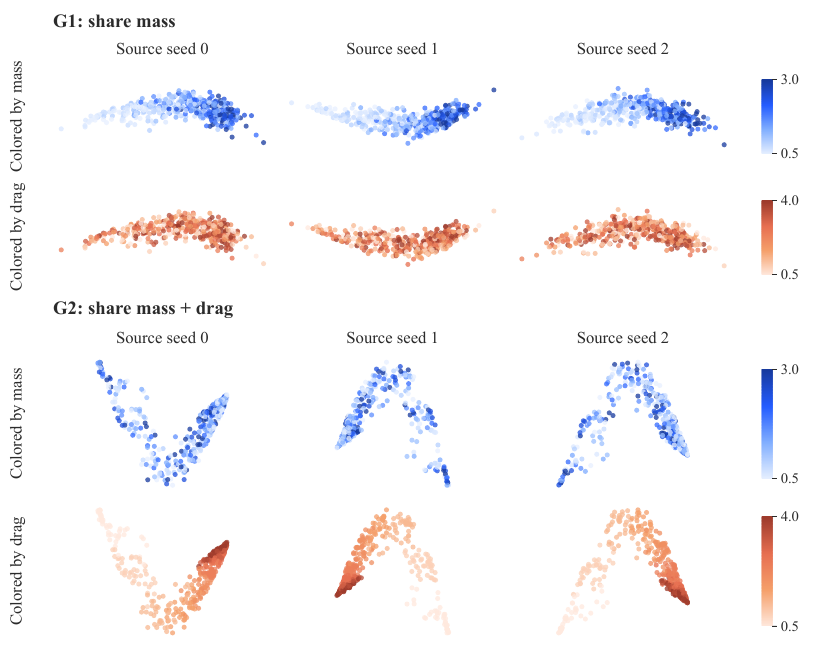}
\caption{\textbf{Persistent-context atlas under $G_1$ and $G_2$.}
Each panel shows the same 400 validation systems. Each frozen source has an
independent PCA projection; mass and drag rows recolor identical coordinates
using common physical-value ranges. Axes are not aligned across sources.
The atlas illustrates the representations; quantitative comparisons use the
geometry and probes in Appendix~\ref{app:selectivity}.}
\label{fig:poke-persistent-atlas}
\end{figure}

\paragraph{Visualizing the frozen representation.}
Figure~\ref{fig:poke-persistent-atlas} displays all three replicas under both
relations. Its factor coloring complements the probes; apparent spacing or
orientation across independently projected sources is not a quantitative comparison.

\paragraph{Persistent-context atlas.}
The atlas uses the same six step-20,000 G1/G2 checkpoints and existing
400-system validation bank as the frozen factor probes. The registered
training systems are the sorted first 400 indices from the training-bank
permutation under seed 20260903, followed by the validation-bank permutation;
this selects all 400 validation systems. Each code averages rollouts 0--3 and
anchors $24,32,40,47$. As in the geometry analysis, we first average anchors
within each rollout, fit coordinatewise moments on the 400 training systems'
four rollout means (standard-deviation floor $10^{-6}$), normalize validation
rollout means, and then form system centroids. No source update or
confirmation/test access occurs. We center each source's validation centroids
and compute a full-SVD PCA without whitening; the largest-magnitude loading
fixes each component's sign, independently of physical labels. No axes are
aligned across sources, and no systems, sources, components or checkpoints
are selected for visual appearance.

\FloatBarrier

\paragraph{Objective controls and native references.}
The factorial keeps the two-view batch, predictor, dimensions and budget
fixed. The relational InfoNCE alternative uses normalized persistent codes,
one positive and 94 negatives per anchor, with all three sources in the
$\tau\in\{0.07,0.10,0.20\}\times\lambda\in\{0.1,1\}$ validation search.
Confirmation geometry is $\msd{0.730}{0.019}$ and ratio $\msd{1.736}{0.104}$.
Native embedding losses retain each learner's own target space.
See Table~\ref{tab:factorial-confirmation} for the complete objective factorial.

\label{app:dclean}\label{app:dclean-acquisition}
The acquisition studies use separate D-Clean and PokeWorld banks:
Poke's raw 24-step drag estimator scores 0.921, versus $\msd{0.320}{0.025}$ for
monolithic JEPA and $\msd{0.481}{0.092}$ for Align + Cross, with 7.1\% lower h16
object-state error on the original acquisition bank.

\label{app:records-formation}
Actual-input versus privileged input estimates remain in
Table~\ref{tab:v3-poke-input-diversity}; D-Clean acquisition uses its own
one-shot sealed population. Native self prediction has no donor route.
\begin{table}[H]
\centering
\small
\setlength{\tabcolsep}{4pt}
\caption{\textbf{D-Clean results on 200 one-shot sealed-test systems.}
Means and sample SD over three sources. Errors use each source's own embedding
target. Correct uses a same-system donor from another rollout; Gap is
shuffled-minus-correct error. Native has no donor route.}
\label{tab:dclean-full}
\begin{tabular}{lcccc}
\toprule
\textbf{Method} & \textbf{Drag $R^2$ $\uparrow$} & \textbf{Self h16 $\downarrow$} & \textbf{Correct $\downarrow$} & \textbf{Gap $\uparrow$}\\
\tableheadrule
Native & \msd{0.9932}{0.0007} & \msd{0.0789}{0.0356} & n.a. & n.a. \\
Structure & \msd{0.9935}{0.0007} & \msd{0.0736}{0.0404} & \msd{0.5512}{0.0681} & \msd{0.0797}{0.0079} \\
Same-rollout alignment & \msd{0.4434}{0.0367} & \msd{0.0520}{0.0078} & \msd{0.0542}{0.0090} & \msd{0.00635}{0.00104} \\
Align & \msd{0.9984}{0.0001} & \msd{0.0140}{0.0040} & \msd{0.0141}{0.0041} & \msd{0.0948}{0.0037} \\
Align + Cross & \msd{0.9986}{0.0002} & \msd{0.0155}{0.0073} & \msd{0.0155}{0.0072} & \msd{0.0941}{0.0031} \\
Align--Random & \msd{0.5299}{0.0400} & \msd{0.0563}{0.0100} & \msd{0.0591}{0.0100} & \msd{0.00034}{0.00009} \\
Supervised $\gamma$ & \msd{0.9985}{0.0002} & \msd{0.0119}{0.0024} & \msd{0.0124}{0.0024} & \msd{0.0969}{0.0044} \\
\bottomrule
\end{tabular}
\end{table}

\begin{table}[!htbp]
\centering\small
\caption{\textbf{PokeWorld geometry under history diversity and relation fidelity.}
Means and source SD over three seeds. A: validation geometry with
train-standardized codes; partial drag correlation controls mass and
stiffness. B: validation and frozen-confirmation results.}
\label{tab:pokeworld-geometry}\label{tab:fidelity-full}
\textbf{A. Independent-history budget}\par\smallskip
\begin{tabular}{lccccc}
        \toprule
\textbf{Relation} & \textbf{Budget} & \textbf{Within $\downarrow$} & \textbf{Between $\uparrow$} & \textbf{Ratio $\uparrow$}
 & \textbf{Partial $\rho_\gamma$ $\uparrow$}\\
        \tableheadrule
        Correct & R2 & \msd{78.35}{4.67} & \msd{29.38}{1.78} & \msd{0.375}{0.017} & \msd{0.095}{0.022} \\
        Correct & R4 & \msd{55.66}{5.17} & \msd{43.32}{2.82} & \msd{0.781}{0.062} & \msd{0.361}{0.036} \\
        Correct & R8 & \msd{28.09}{1.93} & \msd{104.30}{2.01} & \msd{3.729}{0.313} & \msd{0.773}{0.009} \\
        Random & R2 & \msd{70.80}{8.74} & \msd{17.28}{2.03} & \msd{0.244}{0.002} & \msd{-0.001}{0.017} \\
        Random & R4 & \msd{83.68}{16.91} & \msd{20.39}{3.64} & \msd{0.245}{0.008} & \msd{-0.010}{0.016} \\
        Random & R8 & \msd{126.90}{0.89} & \msd{34.18}{0.39} & \msd{0.269}{0.004} & \msd{-0.006}{0.004} \\
        \bottomrule
    \end{tabular}
\par\medskip\textbf{B. Relation fidelity}\par\smallskip
\begin{tabular}{ccccc}
\toprule
 & \multicolumn{2}{c}{\textbf{Validation}} & \multicolumn{2}{c}{\textbf{Confirmation}}\\
\cmidrule(lr){2-3}\cmidrule(lr){4-5}
\textbf{$\alpha$} & \textbf{Ratio $\uparrow$} & \textbf{Drag $\uparrow$} & \textbf{Ratio $\uparrow$} & \textbf{Drag $\uparrow$}\\
\tableheadrule
$0$ & \msd{0.272}{0.006} & \msd{0.061}{0.031} & \msd{0.260}{0.006} & \msd{0.031}{0.026} \\
$0.5$ & \msd{0.976}{0.051} & \msd{0.243}{0.056} & \msd{0.880}{0.047} & \msd{0.264}{0.063} \\
$1$ & \msd{3.749}{0.058} & \msd{0.781}{0.027} & \msd{4.391}{0.156} & \msd{0.796}{0.023} \\
\bottomrule
\end{tabular}
\end{table}

\begin{table}[!htbp]
\centering
\small\setlength{\tabcolsep}{2.8pt}
\caption{\textbf{Factor-selective geometry under relation refinement.}
Means and sample SD over three sources. Exact factor tuples are split-disjoint;
individual factor levels are shared.}
\label{tab:refinement-full}

\begin{tabular}{lcccccc}
\toprule
 & \multicolumn{3}{c}{\textbf{Validation}} & \multicolumn{3}{c}{\textbf{Confirmation}}\\
\cmidrule(lr){2-4}\cmidrule(lr){5-7}
\textbf{Condition} & \textbf{Mass} & \textbf{Drag} & \textbf{Stiffness} & \textbf{Mass} & \textbf{Drag} & \textbf{Stiffness}\\
\tableheadrule
Structure  & \msd{0.069}{0.008} & \msd{0.194}{0.007} & \msd{0.037}{0.012} & \msd{0.057}{0.010} & \msd{0.190}{0.002} & \msd{0.018}{0.008} \\
$G_1$ & \msd{0.419}{0.006} & \msd{0.013}{0.009} & \msd{0.103}{0.004} & \msd{0.489}{0.009} & \msd{0.017}{0.006} & \msd{0.084}{0.003} \\
$G_2$ & \msd{0.015}{0.000} & \msd{0.720}{0.009} & \msd{-0.013}{0.001} & \msd{-0.004}{0.001} & \msd{0.732}{0.012} & \msd{-0.012}{0.001} \\
$G_3$ & \msd{0.064}{0.013} & \msd{0.427}{0.087} & \msd{0.067}{0.035} & \msd{0.077}{0.035} & \msd{0.413}{0.082} & \msd{0.059}{0.023} \\
Random & \msd{0.030}{0.007} & \msd{0.036}{0.014} & \msd{0.054}{0.008} & \msd{0.022}{0.009} & \msd{0.013}{0.007} & \msd{0.044}{0.006} \\
\bottomrule
\end{tabular}
\end{table}

\begin{table}[!htbp]
\centering\small
\caption{\textbf{Objective components on 400 confirmation systems.}
Means and sample SD over three sources. Factor columns give partial geometry;
Ratio is between/within separation. Native h16 errors use each source's own
embedding target.}
\label{tab:factorial-confirmation}
\begin{tabular}{lccccc}\toprule
\textbf{Condition} & \textbf{Mass $\uparrow$} & \textbf{Drag $\uparrow$} & \textbf{Stiff. $\uparrow$} & \textbf{Ratio $\uparrow$} & \textbf{h16 $\downarrow$}\\ \tableheadrule
Structure & \msd{0.058}{0.005} & \msd{0.191}{0.005} & \msd{0.070}{0.011} & \msd{0.299}{0.006} & \msd{0.404}{0.033} \\
Align & \msd{0.012}{0.009} & \msd{0.781}{0.004} & \msd{0.016}{0.005} & \msd{4.203}{0.289} & \msd{0.386}{0.019} \\
Cross & \msd{0.107}{0.009} & \msd{0.229}{0.009} & \msd{0.106}{0.002} & \msd{0.362}{0.009} & \msd{0.373}{0.005} \\
Align + Cross & \msd{0.006}{0.004} & \msd{0.788}{0.001} & \msd{0.011}{0.002} & \msd{4.411}{0.142} & \msd{0.381}{0.010} \\
\bottomrule\end{tabular}
\end{table}

\paragraph{Donor support and candidate composition.}
\label{app:donor-pool-control}
$G_1$ and $G_2$ use the same frozen donor-bank file. Each recipient has three
candidate donor systems and four rollouts per candidate, giving twelve
supported donor histories under either relation. At each update, the sampler
selects one of the three systems uniformly, then selects one of its four
rollouts uniformly. This fixed training support is distinct from the full
set of semantically eligible systems (Table~\ref{tab:donor-pool-audit}).

$G_1$ candidate construction uses fixed drag-index offsets $(1,3,7)$ and a
different-stiffness system at each offset. $G_2$ selects three candidates from
the same-mass, same-drag set with different stiffness. The observed $G_1$/$G_2$
accessibility contrast therefore compares equal-sized training supports with
different candidate composition and preselection. A literal reduction of $G_1$
to $G_2$'s training support size leaves three candidates unchanged. Candidate composition and donor-exposure frequencies follow the pairing rule;
equal support does not separate them from shared-factor semantics.

\begin{table}[!htbp]\centering\small
\caption{\textbf{Semantic eligibility and the fixed training support.} Counts
are per recipient. Eligible systems satisfy the relation; training uses
donors from only three preselected systems.}
\label{tab:donor-pool-audit}
\setlength{\tabcolsep}{5pt}
\begin{tabular}{@{\hspace{3.5pt}}llcccc@{\hspace{3.5pt}}}\toprule
\textbf{Split} & \textbf{Relation} & \textbf{Eligible systems} & \textbf{Selected systems} & \textbf{Rollouts/system} & \textbf{Supported histories}\\\tableheadrule
Training & $G_1$ & 173--174 & 3 & 4 & 12\\
Training & $G_2$ & 19 & 3 & 4 & 12\\
Validation & $G_1$ & 35--36 & 3 & 4 & 12\\
Validation & $G_2$ & 3 & 3 & 4 & 12\\\bottomrule
\end{tabular}\end{table}
\FloatBarrier

\subsection{Objective components and reference learners: SpringWorld}
\label{app:spring-components}\label{app:spring-source-geometry}
Alignment-containing recipes improve cold-start prediction and cross-history
organization. Task error uses one separately fitted reader per source recipe
(reader seed 0) on 100 systems and 600 cases. Geometry uses frozen $P_{64}$
before reader fitting on 178 systems with four histories each.

Align accounts for most of the original cold-start gain. Adding Cross to
Align lowers error in all three sources, by 23.04\%, 8.27\% and 0.10\%. Geometry uses train-standardized codes and
partial Spearman factor-distance correlations, controlling the other two
factors ($k/m$ controls drag). Larger ratios indicate greater between-system separation relative to
within-system variability.

Supervised reference encoders use factor-related supervision and serve as
references rather than competitors. The question for the predictive learners
is whether they organize this information with relation assignments but without
regressing the numerical factor labels.
\paragraph{Supervised reference encoders.}\label{app:spring-development}
Table~\ref{tab:spring-complete} compares GRU, Transformer, TCN and DeepSets
using one source and reader per recipe. The supervised TCN has the lowest
development error in this reference grid. The three-source objective comparison
is in Table~\ref{tab:v3-spring-components}.

\label{app:records-spring-dev}
Table~\ref{tab:v3-spring-components} reports task error and source geometry
for the two cohorts.

\begin{table}[!htbp]\centering\small\setlength{\tabcolsep}{4pt}
\caption{\textbf{Align accounts for most of the SpringWorld gain; Cross lowers error further.}
Three-source means from separate cohorts: A, cold-start error on 100
development systems with one reader per source; B, frozen-source geometry on
178 systems. Ratio averages the source-wise between/within ratios.}
\label{tab:v3-spring-components}\label{tab:spring-cold-replication}\label{tab:spring-source-geometry}
\textbf{A. Three-source component mean (reader 0)}\par\smallskip
\begin{tabular}{@{\hspace{3.5pt}}lrrrrr@{\hspace{3.5pt}}}\toprule
 & \textbf{Native} & \textbf{Structure} & \textbf{Align} & \textbf{Cross} & \textbf{Align + Cross}\\\tableheadrule
Mean & 0.1703 & 0.1716 & 0.1232 & 0.1675 & 0.1104\\\bottomrule
\end{tabular}
\par\medskip\textbf{B. Frozen-source geometry}\par\smallskip
\begin{tabular}{@{\hspace{3.5pt}}lrrrrrrr@{\hspace{3.5pt}}}\toprule
\textbf{Recipe} & \textbf{Within} & \textbf{Between} & \textbf{Ratio} & \textbf{$\rho_m$} & \textbf{$\rho_\gamma$} & \textbf{$\rho_k$} & \textbf{$\rho_{k/m}$}\\\tableheadrule
Structure & 131.6 & 31.64 & 0.240 & 0.001 & 0.005 & $-0.014$ & $-0.025$\\
Align & 47.25 & 90.9 & 2.045 & 0.361 & 0.224 & 0.132 & 0.414\\
Cross & 115.7 & 29.21 & 0.252 & 0.200 & 0.184 & 0.056 & 0.066\\
Align + Cross & 40.58 & 93.63 & 2.360 & 0.354 & 0.193 & 0.141 & 0.421\\\bottomrule
\end{tabular}
\end{table}

The pooled Cross addition to Align is 10.37\%; adding Align to Cross gives
34.10\%. Align + Cross has greater source separation than Structure in all
three geometry replicas.

\begin{table}[!htbp]\centering\small
\caption{\textbf{SpringWorld development references and physical readout.}
A: single-source standardized MSE on 100 cold-start and 178 query-mixture
systems; supervised references use their own training recipes.
B: training-fitted ridge $R^2$ for log factors on 100 validation systems.
These are separate from the sealed and multi-reader evaluations.}
\label{tab:spring-complete}\label{tab:spring-probe}
\textbf{A. Single-source reference (source 0, reader 0)}\par\smallskip
\begin{tabular}{@{\hspace{3.5pt}}lrr@{\hspace{3.5pt}}}\toprule
\textbf{Source} & \textbf{Cold start} & \textbf{Query mixture}\\\tableheadrule
\multicolumn{3}{@{}l}{\emph{Controlled JEPA objectives}}\\
Native & 0.1716 & 0.8511\\
Structure & 0.1711 & 0.8468\\
Align & 0.1195 & 0.8033\\
Cross & 0.1646 & 0.8443\\
\rowcolor{KeyRowTint}
\textbf{Align + Cross} & \textbf{0.0920} & \textbf{0.7892}\\
\addlinespace[3pt]\multicolumn{3}{@{}l}{\emph{Other reference learners}}\\
CPC & 0.2218 & 0.9858\\
RSSM & 0.1653 & 0.7882\\
Supervised split & 0.1841 & 0.8864\\
\addlinespace[3pt]\multicolumn{3}{@{}l}{\emph{Supervised reference encoders}}\\
GRU & 0.0934 & 0.6328\\
Transformer & 0.0887 & 0.6631\\
TCN & 0.0593 & 0.5948\\
DeepSets & 0.0892 & 0.6597\\\bottomrule
\end{tabular}
\par\medskip\textbf{B. Frozen-source physical probes}\par\smallskip
\begin{tabular}{@{\hspace{3.5pt}}lrrrr@{\hspace{3.5pt}}}\toprule
\textbf{Source} & \textbf{$m$} & \textbf{$\gamma$} & \textbf{$k$} & \textbf{$k/m$}\\\tableheadrule
Native & 0.0076 & $-0.0234$ & $-0.0265$ & $-0.0085$\\
Align + Cross & 0.3806 & 0.1529 & 0.2023 & 0.4945\\\bottomrule
\end{tabular}
\end{table}

\FloatBarrier
\Needspace{8\baselineskip}
\subsection{Relation-selective tactile organization: Baxter}
\label{app:baxter}
Hardness and shape relations selectively organize real tactile histories. The crossed bank contains cube/cylinder objects at three common
hardness levels; confirmation holds out grasps from these known configurations.
\begin{figure}[!htbp]
\centering
\includegraphics[width=\linewidth]{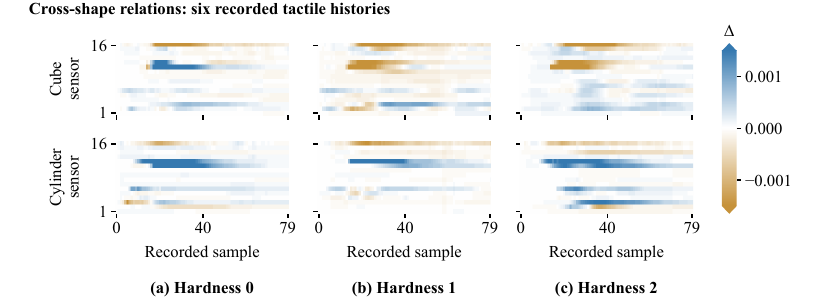}
\caption{\textbf{Shared hardness connects tactile histories across object shapes.}
Columns share hardness levels 0--2; rows differ in shape. Sensor changes are
relative to the initial sample, with a common symmetric scale saturated at
the pooled 97th percentile of absolute change.}
\label{fig:v3-baxter}\label{fig:baxter-interface}
\end{figure}
\begin{figure}[!htbp]\centering
\includegraphics[width=\linewidth]{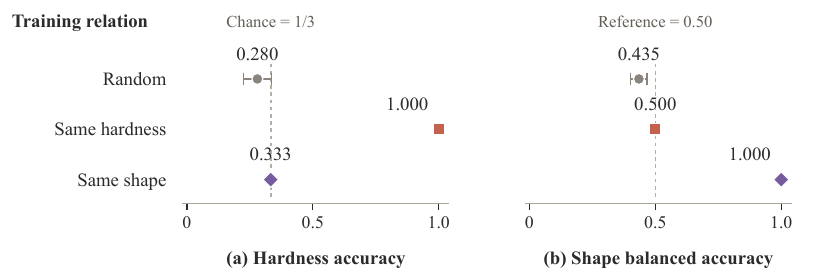}
\caption{\textbf{Relation choice selects which tactile property becomes accessible.}
Means and sample SD over three sources on held-out grasps from six known
configurations. (a) Cross-shape hardness accuracy; (b) shape balanced accuracy
with one hardness level held out. Dashed lines mark the respective chance
references, $1/3$ and $0.50$.}
\label{fig:baxter-selectivity}
\end{figure}

The input recoverability test reaches hardness accuracy $\msd{0.954}{0.003}$ using eight
samples and $\msd{1.000}{0.000}$ using the full peak; shape is already perfect
at both lengths. Thus, the input already supports both classifications.

\Needspace{7\baselineskip}
The key comparison is between the two relation-trained models: hardness
accuracy is $1.000$ under $\mathcal R_H$ versus $0.333$ under $\mathcal R_S$;
shape balanced accuracy is $1.000$ under $\mathcal R_S$ versus $0.500$ under
$\mathcal R_H$. The corresponding contrasts $C_H=0.667$ and $C_S=0.500$
are positive in every source on validation and confirmation. This reversal
shows that the pairing rule selects which property is accessible within the
six known configurations; Random provides an additional unstructured-pairing
reference.

Forty-sample files are alternate windows from the same acquisitions.

\FloatBarrier
\subsection{Structured relation errors}
\label{app:structured-mismatch}
This three-source PokeWorld comparison separates correct-pair frequency from
the structure of pairing errors. The conditions are correct pairing;
random mismatch with $\alpha=0.5$; drag-only mismatch with $\alpha=0.5$;
and mass-only mismatch with $\alpha=0.5$. Structured errors preserve the
other two physical factors. All four conditions use a common training bank
with legal exact single-factor mismatches, and share initialization, recipient
draws, correct/error masks, source budgets and evaluation protocol within seed.
All twelve Align-only sources ($\lambda_p=1$, $\lambda_x=0$) were trained
for 20,000 updates on this new 2,000-system bank, with source seeds 0--2;
no source from the earlier fidelity bank is reused. Probe fitting uses
200 training systems, and evaluation uses 200 common validation systems
with four rollouts and four history anchors each. All results below are
development-only. The finite-bank mismatch kernel selects a distinct legal donor; it differs
from independent factor resampling in Appendix~\ref{app:formation-theory}.

Partial geometry and training-fitted probe $R^2$ quantify each factor's
organization and accessibility. The between/within ratio measures system
separation. Differences from correct pairing are computed within source.
The corollary characterizes alignment pressure for a fixed representation;
the factor-specific response of a trained encoder is measured by these
comparisons.

\begin{table}[!htbp]\centering\small
\caption{\textbf{Factor measurements under structured relation errors.}
Means and sample SD across three sources. Probe targets and partial
correlations follow Appendix~\ref{app:poke-formation}'s PokeWorld protocol.}
\label{tab:structured-mismatch-factors}
\begin{tabular}{@{\hspace{3.5pt}}llcc@{\hspace{3.5pt}}}\toprule
\textbf{Pairing} & \textbf{Factor} & \textbf{Probe $R^2$} & \textbf{Partial geometry $\rho$}\\\tableheadrule
Correct & Mass & $0.1630\sdev{0.0138}$ & $0.0796\sdev{0.0105}$\\
Correct & Drag & $0.6087\sdev{0.0881}$ & $0.4665\sdev{0.0461}$\\
Correct & Stiffness & $0.0589\sdev{0.0432}$ & $0.0168\sdev{0.0186}$\\
Random mismatch & Mass & $0.2229\sdev{0.0274}$ & $0.1954\sdev{0.0013}$\\
Random mismatch & Drag & $0.3987\sdev{0.0305}$ & $0.3092\sdev{0.0043}$\\
Random mismatch & Stiffness & $-0.0938\sdev{0.1124}$ & $0.0806\sdev{0.0024}$\\
Drag-only mismatch & Mass & $0.3377\sdev{0.0457}$ & $0.2978\sdev{0.0152}$\\
Drag-only mismatch & Drag & $0.3608\sdev{0.0682}$ & $0.1162\sdev{0.0224}$\\
Drag-only mismatch & Stiffness & $0.0970\sdev{0.0451}$ & $0.1312\sdev{0.0320}$\\
Mass-only mismatch & Mass & $0.1766\sdev{0.1025}$ & $0.0180\sdev{0.0064}$\\
Mass-only mismatch & Drag & $0.9105\sdev{0.0435}$ & $0.7013\sdev{0.0186}$\\
Mass-only mismatch & Stiffness & $0.1166\sdev{0.0127}$ & $-0.0129\sdev{0.0026}$\\
\bottomrule\end{tabular}\end{table}

\begin{figure}[!htbp]\centering
\includegraphics[width=\linewidth]{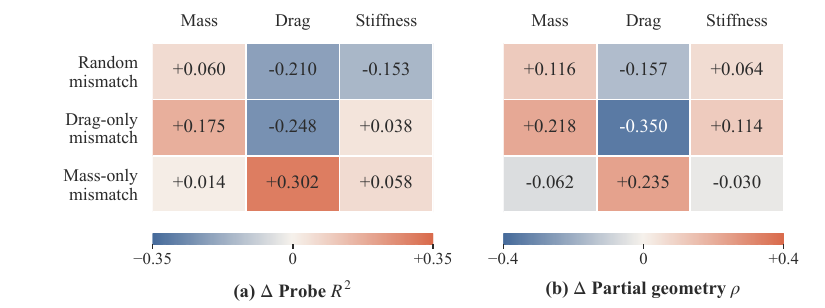}
\caption{\textbf{Structured mismatches redistribute factor organization.}
Mean paired-source changes from correct pairing in (a) probe $R^2$ and
(b) partial geometry $\rho$; positive indicates an increase. All mismatch
rules use $\alpha=0.5$ and three Align-only sources on 200 development
validation systems. Mass-only mismatch lowers mass geometry without lowering
its mean probe score. Source SDs are in Table~\ref{tab:structured-mismatch-factors}.}
\label{fig:structured-mismatch-differences}
\end{figure}

\begin{table}[!htbp]\centering\small
\caption{\textbf{Pairing quality, system separation and native prediction.}
Coverage is the fraction of recipients with a legal donor; factor columns
give match rates, and separation is the between/within ratio. Native latent
MSE uses each encoder's own target space, with target variance in brackets.
Errors and ratios show means and sample SD across three sources.}
\label{tab:structured-mismatch-diagnostics}
\setlength{\tabcolsep}{4pt}
\begin{tabular}{@{\hspace{3.5pt}}lccccc@{\hspace{3.5pt}}}\toprule
\textbf{Pairing} & \textbf{Coverage (\%)} & \textbf{Mass (\%)} & \textbf{Drag (\%)} & \textbf{Stiffness (\%)} & \textbf{Separation}\\\tableheadrule
Correct & 100 & 100 & 100 & 100 & $1.0601\sdev{0.1093}$\\
Random mismatch & 100 & 54.97 & 54.99 & 51.76 & $0.9694\sdev{0.0236}$\\
Drag-only mismatch & 100 & 100 & 50 & 100 & $1.1618\sdev{0.1712}$\\
Mass-only mismatch & 100 & 50 & 100 & 100 & $3.8995\sdev{1.3415}$\\
\bottomrule\end{tabular}
\par\smallskip
\begin{tabular}{@{\hspace{3.5pt}}lccc@{\hspace{3.5pt}}}\toprule
\textbf{Pairing} & \textbf{$H=1$} & \textbf{$H=4$} & \textbf{$H=16$}\\\tableheadrule
Correct & $0.0719\sdev{0.0071}$ [0.6226] & $0.1103\sdev{0.0054}$ [0.6267] & $0.2312\sdev{0.0040}$ [0.6285]\\
Random mismatch & $0.0108\sdev{0.0008}$ [0.7416] & $0.0506\sdev{0.0010}$ [0.7439] & $0.2126\sdev{0.0015}$ [0.7457]\\
Drag-only mismatch & $0.0265\sdev{0.0034}$ [0.7061] & $0.0688\sdev{0.0045}$ [0.7075] & $0.2216\sdev{0.0038}$ [0.7083]\\
Mass-only mismatch & $0.0223\sdev{0.0004}$ [0.6841] & $0.0675\sdev{0.0043}$ [0.6885] & $0.2157\sdev{0.0041}$ [0.6966]\\
\bottomrule\end{tabular}\end{table}

Drag-only mismatch changes drag geometry from $0.4665$ to $0.1162$ and mass
geometry from $0.0796$ to $0.2978$. Mass-only mismatch changes mass geometry
to $0.0180$ and drag geometry to $0.7013$. All four directions hold in each
of the three source seeds. Accessibility is less uniform: mass-only mismatch
does not lower mass probe $R^2$ ($0.1766$ versus $0.1630$ for correct pairing).
These results support factor-directed organization under structured errors.
Correct pairing does not minimize native latent MSE at any evaluated horizon;
the prediction panel reports each frozen encoder's own target space and
scale, and does not measure common physical-state error or control return.
\FloatBarrier

\subsection{Relations inferred from learned histories}
\label{app:inferred-relations}
A Structure teacher trained without relation supervision encodes one
24-frame history per training rollout. Each source seed defines a frozen
graph over 8,000 rollouts from the 2,000-system bank. After per-coordinate
standardization on training codes, each rollout receives three Euclidean
nearest neighbors, excluding itself. Physical labels and system grouping
are not used to construct or select the graph. RandomHistory draws three
neighbors from the same candidate universe. The graphs are fixed before
source training; physical labels are used only for subsequent diagnostics.

Inferred and RandomHistory each use three newly trained Align-only sources
($\lambda_p=1$, $\lambda_x=0$), with 20,000 updates per source. True reuses
three $G_3$ sources from the same bank with the same source hyperparameters.\footnote{The reused $G_3$ sources were trained on L20 GPUs with PyTorch 2.6;
the new graph-based sources used RTX PRO 6000 GPUs with PyTorch 2.8.}
All nine models use the common evaluation protocol: probes are fitted on
200 training systems, and evaluation uses 200 validation systems with
3,200 windows per model. No sealed data enter this analysis.
The earlier $\alpha$ curve uses a different bank and is a descriptive
reference, not an interpolation rule for these graphs.

\begin{table}[!htbp]\centering\small
\caption{\textbf{Agreement and coverage of history-inferred relations.}
Match rates and coverage are percentages, averaged over three source-specific
graphs. Donor usage is donor-rollout entropy normalized by $\log 8000$.
True uses exact same-system relations.}
\label{tab:inferred-relation-quality}
\setlength{\tabcolsep}{4pt}
\begin{tabular}{@{\hspace{3.5pt}}lcccccc@{\hspace{3.5pt}}}\toprule
\textbf{Relation} & \textbf{Tuple $p$} & \textbf{Mass} & \textbf{Drag} & \textbf{Stiffness} & \textbf{Coverage} & \textbf{Donor usage}\\\tableheadrule
True & 100 & 100 & 100 & 100 & 100 & 1.000\\
Inferred & 0.0639 & 11.15 & 13.30 & 3.64 & 100 & 0.974\\
RandomHistory & 0.0431 & 10.04 & 10.01 & 3.50 & 100 & 0.980\\
\bottomrule\end{tabular}\end{table}

\begin{table}[!htbp]\centering\small
\caption{\textbf{Factor accessibility and geometry with inferred relations.}
Entries are means and sample SD across three sources.}
\label{tab:inferred-relation-factors}
\begin{tabular}{@{\hspace{3.5pt}}llcc@{\hspace{3.5pt}}}\toprule
\textbf{Relation} & \textbf{Factor} & \textbf{Probe $R^2$} & \textbf{Partial geometry $\rho$}\\\tableheadrule
True & Mass & $0.1700\sdev{0.0154}$ & $0.0737\sdev{0.0092}$\\
True & Drag & $0.6556\sdev{0.0791}$ & $0.4807\sdev{0.0877}$\\
True & Stiffness & $0.0670\sdev{0.0020}$ & $0.0314\sdev{0.0453}$\\
Inferred & Mass & $0.0538\sdev{0.0423}$ & $0.0360\sdev{0.0069}$\\
Inferred & Drag & $0.1563\sdev{0.0316}$ & $0.0061\sdev{0.0111}$\\
Inferred & Stiffness & $0.0314\sdev{0.0111}$ & $0.0666\sdev{0.0028}$\\
RandomHistory & Mass & $0.0911\sdev{0.0318}$ & $0.0256\sdev{0.0117}$\\
RandomHistory & Drag & $0.1065\sdev{0.0870}$ & $0.0342\sdev{0.0043}$\\
RandomHistory & Stiffness & $-0.0235\sdev{0.0202}$ & $0.0323\sdev{0.0255}$\\
\bottomrule\end{tabular}\end{table}

\begin{table}[!htbp]\centering\small
\caption{\textbf{System separation and native prediction with inferred relations.}
Means and sample SD across three sources. The bracketed second line gives
mean target variance. Each source defines its own latent prediction target.}
\label{tab:inferred-relation-task}
\setlength{\tabcolsep}{4pt}
\begin{tabular}{@{\hspace{3.5pt}}lcccc@{\hspace{3.5pt}}}\toprule
\textbf{Relation} & \textbf{Between/within} & \textbf{Native $H=1$} & \textbf{Native $H=4$} & \textbf{Native $H=16$}\\\tableheadrule
True & $1.0683\sdev{0.2664}$ & $0.0721\sdev{0.0103}$ & $0.1124\sdev{0.0042}$ & $0.2327\sdev{0.0099}$\\
 & & [0.6195] & [0.6225] & [0.6249]\\
Inferred & $0.2577\sdev{0.0013}$ & $0.0326\sdev{0.0031}$ & $0.0743\sdev{0.0043}$ & $0.2357\sdev{0.0051}$\\
 & & [0.7309] & [0.7366] & [0.7482]\\
RandomHistory & $0.2499\sdev{0.0117}$ & $0.0121\sdev{0.0076}$ & $0.0542\sdev{0.0144}$ & $0.2268\sdev{0.0051}$\\
 & & [0.7405] & [0.7419] & [0.7436]\\
\bottomrule\end{tabular}\end{table}

For a shared-space score $T$, recovery denotes the fraction of the
True--RandomHistory gap attained by Inferred:
$(T_{\rm Inferred}-T_{\rm RandomHistory})/(T_{\rm True}-T_{\rm RandomHistory})$.
Ratios use source means and are not clipped. We report them only where the
True reference exceeds RandomHistory in every paired-source resample.
Native latent errors are retained as diagnostics, not common physical-task
errors or recovery scores.

\begin{table}[!htbp]\centering\small
\caption{\textbf{Inferred-relation changes and recovery relative to True.}
$\Delta$ is Inferred minus RandomHistory. Recovery intervals are 95\%
paired-source bootstrap intervals over three seeds; they do not measure
cross-task stability. Dashes denote inapplicable ratios.}
\label{tab:inferred-relation-recovery}
\setlength{\tabcolsep}{5pt}
\begin{tabular}{@{\hspace{3.5pt}}llrc@{\hspace{3.5pt}}}\toprule
\textbf{Measurement} & \textbf{Factor} & \textbf{$\Delta$} & \textbf{Recovery (\%)}\\\tableheadrule
Probe $R^2$ & Mass & $-0.0373$ & $-47.3\,[-135.5, 1.0]$\\
Probe $R^2$ & Drag & $+0.0498$ & $9.1\,[-2.7, 30.5]$\\
Probe $R^2$ & Stiffness & $+0.0549$ & $60.7\,[55.3, 70.0]$\\
Geometry $\rho$ & Mass & $+0.0104$ & $21.6\,[-27.6, 53.4]$\\
Geometry $\rho$ & Drag & $-0.0281$ & $-6.3\,[-8.6, -4.2]$\\
Geometry $\rho$ & Stiffness & $+0.0343$ & \textemdash{}$^{a}$\\
Between/within ratio & \textemdash{} & $+0.0077$ & $0.9\,[-0.5, 2.4]$\\
Native h16 error & \textemdash{} & $+0.0090$ & \textemdash{}$^{b}$\\
\bottomrule\end{tabular}
\begin{papertablenotes}
\tablenote{a}{The True--RandomHistory stiffness-geometry gap crosses zero under
source resampling; no consistent True advantage defines a recovery scale.}
\tablenote{b}{The native models use different latent target spaces.}
\end{papertablenotes}\end{table}

The inferred graph modestly enriches drag matches (13.30\% versus 10.01\%)
and mass matches (11.15\% versus 10.04\%). Probe effects remain
factor-dependent: mean stiffness $R^2$ rises from $-0.0235$ to $0.0314$,
with increases in all three sources; drag rises by $0.0498$ on average but improves in only one
source, while mass falls by $0.0373$ on average. Drag geometry falls in all
three sources. True has higher probe $R^2$ for every factor and source.
Thus, this fixed history-based rule changes selected accessible factors
but does not recover the effect of the supplied physical relations.

\FloatBarrier
\Needspace{12\baselineskip}
\section{Use: Fixed-Route Context Interventions}\label{app:information-use}\label{app:records-use-group}
This section expands
Figure~\ref{fig:functional-landscape}: SpringWorld provides physical donor landscapes,
PokeWorld tests factor-specific direction, and RH20T and Balls examine recorded
task context and actual reader memory. Swimmer tests recipient-specific updates
through frozen predictor weights. Probes measure accessibility; fixed-route interventions measure
dependence on supplied context.
\FloatBarrier
\subsection{Physical donor landscapes: SpringWorld}
\label{app:landscape}\label{app:spring-functional-landscape}
Changing donor physics redirects the fixed reader toward the corresponding target dynamics. The assay uses 64 continuous development systems,
three frozen sources and reader seed 0. For each factor, five donor levels
and five simulator target levels share initial states and actions; the two
other physical parameters remain fixed. A fixed rule selects the query without
using model error. All $5\times5$ grids are complete; h16 is primary.

Valley depth is mean off-diagonal error minus mean diagonal error, averaging
20 and five cells respectively. Models use the original training-only target
normalization. Systems then sources receive equal weight; intervals use
10,000 paired-system draws conditional on these fitted routes.
\begin{figure}[!htbp]\centering
\includegraphics[width=\linewidth]{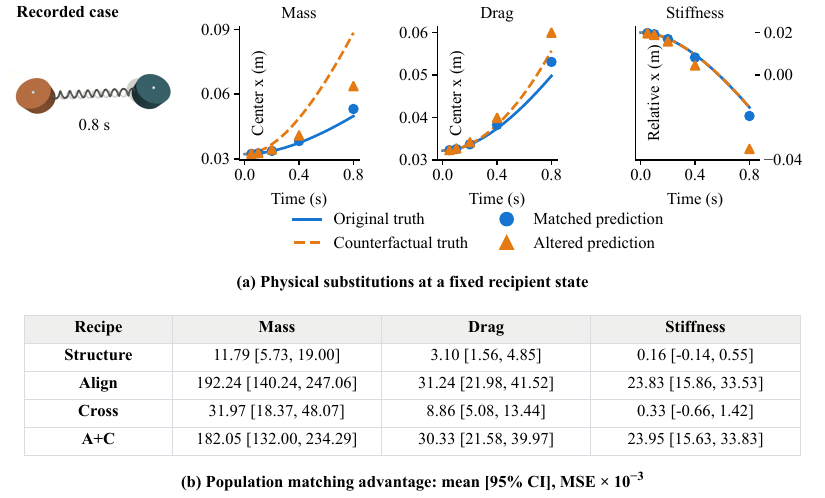}
\caption{\textbf{Physical donor substitutions redirect predictions and produce error valleys.}
(a) One preselected case: lines show simulator truth and markers show
predictions at horizons 1/2/4/8/16. (b) Valley depths for four separately
fitted recipes on 64 systems, with paired-system 95\% intervals. Align + Cross
uses the pooled intervals in Table~\ref{tab:spring-functional-landscape}.}
\label{fig:v3-spring-use}\label{fig:spring-functional-cases}
\end{figure}

Align + Cross has positive valley depth for all three factors in every source.
The balanced factorial main effect of Align,
$(\mathrm{Align}-\mathrm{Structure}+\mathrm{Align+Cross}-\mathrm{Cross})/2$,
is 0.16526, 0.02480 and 0.02365 for
mass, drag and stiffness, with respective reported intervals
[0.12092,0.21175], [0.01720,0.03286] and [0.01573,0.03306]; the corresponding Cross main effects
0.00499, 0.00243 and 0.00015 have intervals spanning zero. Each contrast
compares complete source/reader recipes; donor substitutions within each recipe
keep its fitted reader fixed.
\paragraph{Output-coordinate decomposition.}\label{app:physical-response}
The equal-mass center/relative transform decomposes the frozen reader's
physical output response. Simulator relative shares are 0.7143/0.7149/1.0000
for mass/drag/stiffness; reader shares span 0.610--0.700/0.650--0.689/0.637--0.696
across sources. The stiffness relative-minus-center statistic is
0.00155107238, quantifying the distribution of the response across physical
output coordinates.

\label{app:records-use}
The center/relative transform acts on the position and velocity coordinates
$(x_1,y_1,x_2,y_2,v_{x1},v_{y1},v_{x2},v_{y2})$.
\par\medskip
\begingroup\footnotesize
\setlength{\tabcolsep}{3.5pt}
\noindent\begin{minipage}{\linewidth}
\centering
\makeatletter\def\@captype{table}\makeatother
\caption{\textbf{Align + Cross yields positive donor-matching valleys in every source.}
Depth is off-diagonal minus diagonal h16 error on 64 development systems.
Source results and recipe comparisons use three frozen sources and one reader
per source. Paired-system 95\% intervals are conditional on fitted models.}
\label{tab:spring-functional-landscape}\label{tab:spring-factorial-pooled}
\begin{tabular}{@{\hspace{3.5pt}}llrrr@{\hspace{3.5pt}}}
\toprule
\textbf{Factor} & \textbf{Source} & \textbf{Valley depth} & \textbf{95\% interval} & \textbf{$\rho$}\\\tableheadrule
Mass & Pooled & 0.1820 & $[0.1320, 0.2343]$ & 0.591\\
 & Seed 0 & 0.1950 & $[0.1379, 0.2558]$ & 0.606\\
 & Seed 1 & 0.1612 & $[0.1131, 0.2134]$ & 0.592\\
 & Seed 2 & 0.1899 & $[0.1364, 0.2462]$ & 0.601\\
Drag & Pooled & 0.0303 & $[0.0216, 0.0400]$ & 0.585\\
 & Seed 0 & 0.0334 & $[0.0238, 0.0439]$ & 0.600\\
 & Seed 1 & 0.0335 & $[0.0229, 0.0458]$ & 0.578\\
 & Seed 2 & 0.0240 & $[0.0159, 0.0327]$ & 0.542\\
Stiffness & Pooled & 0.0240 & $[0.0156, 0.0338]$ & 0.672\\
 & Seed 0 & 0.0285 & $[0.0184, 0.0407]$ & 0.776\\
 & Seed 1 & 0.0229 & $[0.0143, 0.0328]$ & 0.401\\
 & Seed 2 & 0.0205 & $[0.0124, 0.0299]$ & 0.628\\
\bottomrule
\end{tabular}
\end{minipage}
\par\medskip
\noindent\begin{minipage}{\linewidth}
\centering
\textbf{Table~\ref*{tab:spring-functional-landscape} (continued). Recipe comparison.}\par\smallskip
\begin{tabular}{@{\hspace{3.5pt}}llrrr@{\hspace{3.5pt}}}
\toprule
\multicolumn{2}{@{\hspace{3.5pt}}l}{\textbf{Recipe}} & \textbf{Mass: depth [95\% CI]} & \textbf{Drag: depth [95\% CI]} & \textbf{Stiffness: depth [95\% CI]}\\\tableheadrule
\multicolumn{2}{@{\hspace{3.5pt}}l}{Structure} & 0.01179 $[0.00573, 0.01900]$ & 0.00310 $[0.00156, 0.00485]$ & 0.00016 $[-0.00014, 0.00055]$\\
\multicolumn{2}{@{\hspace{3.5pt}}l}{Align} & 0.1922 $[0.1402, 0.2471]$ & 0.03124 $[0.02198, 0.04152]$ & 0.02383 $[0.01586, 0.03353]$\\
\multicolumn{2}{@{\hspace{3.5pt}}l}{Cross} & 0.03197 $[0.01837, 0.04807]$ & 0.00886 $[0.00508, 0.01344]$ & 0.00033 $[-0.00066, 0.00142]$\\
\multicolumn{2}{@{\hspace{3.5pt}}l}{Align + Cross} & 0.182 $[0.132, 0.2343]$ & 0.03033 $[0.02148, 0.03983]$ & 0.02395 $[0.01539, 0.03398]$\\
\bottomrule
\end{tabular}
\begin{papertablenotes}
The landscape figure uses the upper pooled intervals; this panel uses the four-recipe intervals.
\end{papertablenotes}
\end{minipage}
\par\endgroup

\FloatBarrier
\subsection{Factor-specific responses and direction: PokeWorld}
\label{app:poke-fixed}\label{app:poke-common-use}\label{app:poke-direction}
\begin{figure}[!htbp]\centering
\includegraphics[width=\linewidth]{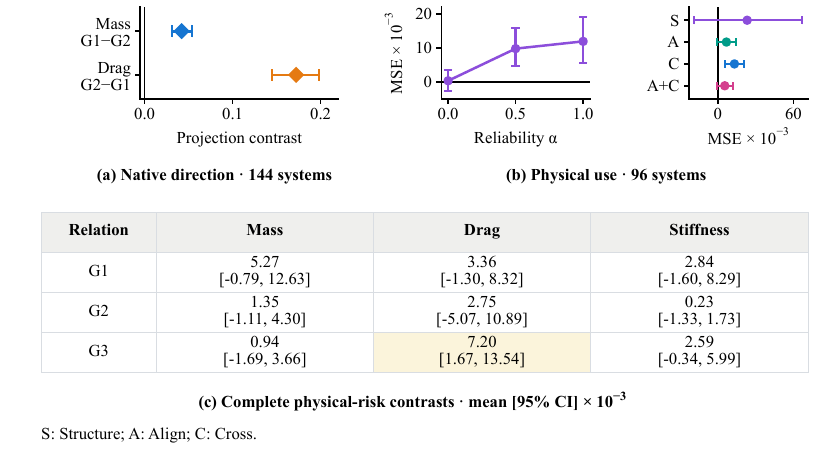}
\caption{\textbf{Shared-factor relations redirect predictive responses.}
(a) Mean $G_1$-minus-$G_2$ mass and $G_2$-minus-$G_1$ drag projections on
144 systems. (b)--(c) Wrong-factor minus matched h16 object-state MSE on a
separate 96-system bank. Pointwise 95\% intervals resample systems with three sources fixed
(and state decoders fixed in (b)--(c)). Yellow marks an interval above zero.
Native projections and physical MSE have distinct units.}
\label{fig:v3-poke-semantics}
\end{figure}

\takeaway{Shared-factor relations redirect frozen predictive responses toward the corresponding mass or drag counterfactual direction.} Physical-risk and
native-embedding assays measure this response in different spaces.
The 96-system physical-risk assay in Figure~\ref{fig:v3-poke-semantics}
holds the native predictor and training-only ridge state decoder fixed.
Matched histories come from an independent recipient interaction; a wrong-factor
donor changes exactly one physical parameter. Three hash-fixed windows per
system retain current observations, actions, target and h16. All three source
seeds use common training-target normalization; $\Delta_f$ is wrong-factor
minus matched four-coordinate object-state MSE.
The principal direction assay evaluates step-20,000 Align $G_1$/$G_2$ checkpoints
(three sources per relation) on 144 eligible validation systems with three
queries each. It uses each source's native 128-dimensional observation
embedding without fitting a new encoder, reader or decoder. For factor $f$ and relation $G$, let
$\Delta\hat z_{f,s}$ be the wrong-donor minus matched prediction displacement
and $\Delta z^{\rm cf}_{f,s}$ the paired counterfactual target displacement
in the native embedding space. We define
\[
P(f,G)=\frac{\sum_s\langle\Delta\hat z_{f,s},\Delta z^{\rm cf}_{f,s}\rangle}
{\sum_s\|\Delta z^{\rm cf}_{f,s}\|_2^2},
\]
pooling within a physical system and then giving source seeds equal weight.
The reported contrasts are
\begin{align*}
\Delta_m &=P(m,G_1)-P(m,G_2)=0.041969,
 &95\%\ \mathrm{CI}&=[0.031176,0.053899],\\
\Delta_\gamma &=P(\gamma,G_2)-P(\gamma,G_1)=0.172192,
 &95\%\ \mathrm{CI}&=[0.144846,0.198290].
\end{align*}
Reported intervals use paired physical-system bootstrap.
Their sum is $0.214161$ $[0.184608,0.243162]$. The already fitted state-decoder
version reports the same directions at a smaller scale:
$\Delta_m=0.004986$ and $\Delta_\gamma=0.020592$.
The primary statistic is a dimensionless projection in each model's native
128-D embedding. Both contrasts show relation-specific functional redirection
on this development bank; the state-decoder assay expresses that response
in state coordinates.

\label{app:records-poke-use}

\subsection{Task-specific donor use in recorded interactions: RH20T}
\label{app:rh20t}\label{app:rh20t-results}\label{app:rh-specificity}
RH20T relations join different episodes sharing task metadata. Three
evaluations test different aspects of the resulting context: a 25-task
forecasting cohort compares matched-donor prediction; a 24-task hard-negative
bank compares donor specificity and absolute risk; and a separate 25-task
validation bank tests action use and donor-count sensitivity.

\paragraph{Matched-donor forecasting.}
Cross lowers all five endpoint means relative to Structure under matched-donor
evaluation. The models jointly predict force/torque and TCP from synchronized
visual, state and action histories. Table~\ref{tab:rh20t-complete-self}
reports all 14 conditions and three fitted seeds.
\begin{figure}[!htbp]\centering
\includegraphics[width=\linewidth]{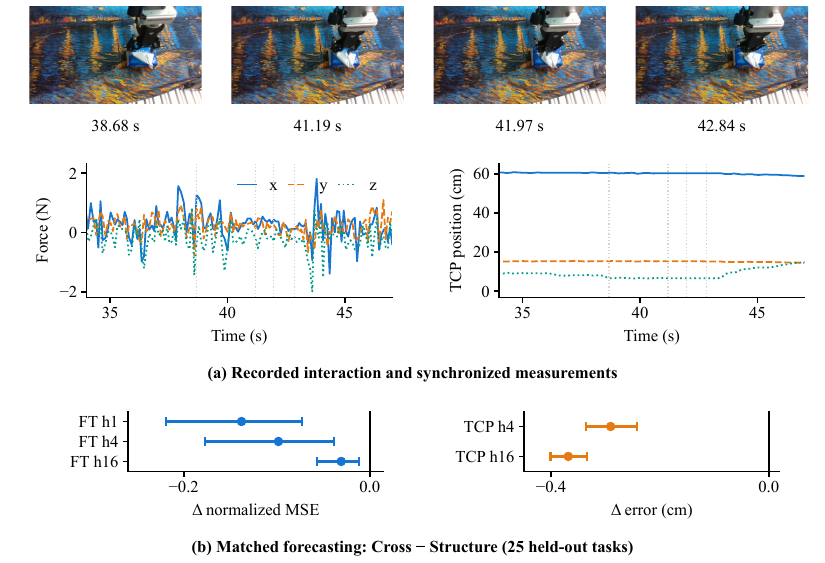}
\caption{\textbf{Cross reduces all five matched-donor forecasting endpoint means.}
(a) RH20T frames and synchronized force/TCP observations.
(b) Cross-Indep minus Structure on 25 held-out tasks; negative favors Cross.
Force/torque uses normalized MSE and TCP uses cm. Task-paired 95\% intervals
follow averaging over three sources. Absolute risks on the separate 24-task
bank are in Table~\ref{tab:v3-rh-risk}.}
\label{fig:v3-rh20t}\label{fig:rh20t-interface}\label{fig:app-rh20t-observation}\label{fig:rh20t-utility-specificity}
\end{figure}
\label{app:records-rh}
\begin{table}[!htbp]
\centering\small\setlength{\tabcolsep}{5pt}\renewcommand{\arraystretch}{1.0}
\caption{\textbf{Self-context and matched-donor forecasting on RH20T.}
Equal-weight means over 25 held-out tasks and three sources. Force/torque uses
normalized MSE; TCP uses cm. LD denotes LowDim; A+C denotes Align + Cross.
Dashes indicate unavailable matched inputs. Shading highlights the
Structure/Cross comparison.}
\label{tab:rh20t-complete-self}\label{tab:rh20t-complete-correct}
\textbf{A. Self-context}\par\smallskip
\begin{tabular}{@{\hspace{3.5pt}}lrrrrr@{\hspace{3.5pt}}}\toprule
 & \multicolumn{3}{c}{\textbf{Force/torque MSE}} & \multicolumn{2}{c}{\textbf{TCP (cm)}}\\
\cmidrule(lr){2-4}\cmidrule(lr){5-6}
\textbf{Condition} & \textbf{h1} & \textbf{h4} & \textbf{h16} & \textbf{h4} & \textbf{h16}\\\tableheadrule
Native & 0.6346 & 1.117 & 1.497 & 1.962 & 4.063\\
\rowcolor{KeyRowTint}
\textbf{Structure} & \textbf{0.6199} & \textbf{1.099} & \textbf{1.498} & \textbf{1.944} & \textbf{3.965}\\
\addlinespace[2pt]
Align--Indep & 0.6633 & 1.132 & 1.503 & 2.262 & 4.429\\
Align--Random & 0.6232 & 1.106 & 1.510 & 2.045 & 4.257\\
\addlinespace[2pt]
Cross--SameEp & 0.6145 & 1.090 & 1.495 & 1.963 & 3.965\\
\rowcolor{KeyRowTint}
\textbf{Cross--Indep} & \textbf{0.6230} & \textbf{1.093} & \textbf{1.497} & \textbf{1.901} & \textbf{3.902}\\
Cross--Random & 0.6307 & 1.096 & 1.495 & 1.948 & 3.950\\
\addlinespace[2pt]
A+C--SameEp & 0.6407 & 1.120 & 1.500 & 2.098 & 4.371\\
A+C--Indep & 0.6602 & 1.130 & 1.505 & 2.268 & 4.435\\
A+C--Random & 0.6225 & 1.101 & 1.507 & 2.036 & 4.258\\
\addlinespace[2pt]
LD--Native & 0.5909 & 1.089 & 1.489 & 1.730 & 3.887\\
LD--Cross--Indep & 0.5760 & 1.069 & 1.488 & 1.598 & 3.748\\
LD--A+C--SameEp & 0.6162 & 1.118 & 1.504 & 2.032 & 4.302\\
LD--A+C--Random & 0.6090 & 1.122 & 1.505 & 1.866 & 4.191\\
\bottomrule\end{tabular}
\par\medskip
\textbf{B. Matched donor}\par\smallskip
\begin{tabular}{@{\hspace{3.5pt}}lrrrrr@{\hspace{3.5pt}}}\toprule
 & \multicolumn{3}{c}{\textbf{Force/torque MSE}} & \multicolumn{2}{c}{\textbf{TCP (cm)}}\\
\cmidrule(lr){2-4}\cmidrule(lr){5-6}
\textbf{Condition} & \textbf{h1} & \textbf{h4} & \textbf{h16} & \textbf{h4} & \textbf{h16}\\\tableheadrule
Native & -- & -- & -- & -- & --\\
\rowcolor{KeyRowTint}
\textbf{Structure} & \textbf{0.7991} & \textbf{1.226} & \textbf{1.551} & \textbf{2.429} & \textbf{4.952}\\
\addlinespace[2pt]
Align--Indep & 0.6667 & 1.135 & 1.504 & 2.536 & 4.644\\
Align--Random & 0.6255 & 1.108 & 1.510 & 2.062 & 4.274\\
\addlinespace[2pt]
Cross--SameEp & 0.6709 & 1.140 & 1.527 & 2.254 & 4.745\\
\rowcolor{KeyRowTint}
\textbf{Cross--Indep} & \textbf{0.6612} & \textbf{1.128} & \textbf{1.521} & \textbf{2.139} & \textbf{4.584}\\
Cross--Random & 0.6665 & 1.127 & 1.516 & 2.158 & 4.604\\
\addlinespace[2pt]
A+C--SameEp & 0.6498 & 1.127 & 1.506 & 2.407 & 4.578\\
A+C--Indep & 0.6633 & 1.133 & 1.506 & 2.573 & 4.670\\
A+C--Random & 0.6250 & 1.103 & 1.507 & 2.045 & 4.269\\
\addlinespace[2pt]
LD--Native & -- & -- & -- & -- & --\\
LD--Cross--Indep & 0.6039 & 1.131 & 1.531 & 1.893 & 4.471\\
LD--A+C--SameEp & 0.6187 & 1.121 & 1.506 & 2.078 & 4.339\\
LD--A+C--Random & 0.6108 & 1.124 & 1.506 & 1.905 & 4.230\\
\bottomrule\end{tabular}
\end{table}

\FloatBarrier

\paragraph{Donor specificity and absolute risk.}
The 24-task bank contains 10,681 windows with matched,
action-metadata-matched/different-task, calibration-matched/different-task and
random donor eligibility. Align + Cross and Monolithic Q+D share the manifest,
48-pair batch, 20,000-step checkpoint and three seeds; parameter counts differ
by 0.309\%. Define
\begin{equation}
S_f=E_f(\mathrm{hard})-E_f(\mathrm{matched}),\qquad
D_{\rm spec}=S_{\rm A+C}-S_{\rm Mono}.
\label{eq:rh20t-a2-did}
\end{equation}
Align + Cross has greater FT donor specificity, while Monolithic has lower
matched error. Table~\ref{tab:v3-rh-risk} reports both comparisons and their
paired intervals; the TCP matched-error interval favors Monolithic.
\begin{table}[H]
\centering\small
\setlength{\tabcolsep}{3pt}
\caption{\textbf{Donor specificity and absolute risk on the 24-task bank.}
Means weight tasks equally after averaging three sources, with task-paired
95\% intervals. Specificity is hard-minus-matched error; contrasts are A+C minus
Monolithic. Positive risk differences favor Monolithic. FT h4 is primary;
TCP h16 is secondary.}
\label{tab:v3-rh-risk}\label{tab:rh20t-hard-negative}\label{tab:rh20t-absolute-hard}\label{tab:rh20t-absolute-contrast}
\textbf{A. Donor specificity}\par\smallskip
\begin{tabular}{@{\hspace{3.5pt}}lccc@{\hspace{3.5pt}}}
\toprule
\textbf{Endpoint} & \textbf{$S_{\mathrm{A+C}}$} & \textbf{$S_{\rm Mono}$} & \textbf{$D_{\rm spec}$}\\
\tableheadrule
\rowcolor{KeyRowTint}
\textbf{FT h4 MSE}
& $0.00260\;[0.00107, 0.00429]$
& $0.00030\;[-0.00050, 0.00134]$
& $\mathbf{0.00230}\;[0.00073, 0.00404]$ \\
TCP h16 (cm)
& $0.0891\;[0.0209, 0.1614]$
& $0.0176\;[-0.0025, 0.0414]$
& $0.0715\;[-0.0042, 0.1480]$ \\
\bottomrule
\end{tabular}
\par\medskip
\textbf{B. Absolute matched and hard-donor risk}\par\smallskip
\begin{tabular}{@{\hspace{3.5pt}}llrrrr@{\hspace{3.5pt}}}
\toprule
\textbf{Endpoint} & \textbf{Model} & \textbf{Matched} & \textbf{95\% interval} & \textbf{Hard} & \textbf{95\% interval}\\
\tableheadrule
FT h4 & Align+Cross & 0.3343 & $[0.2501, 0.4262]$ & 0.3369 & $[0.2518, 0.4300]$ \\
FT h4 & Monolithic & 0.3205 & $[0.2390, 0.4104]$ & 0.3208 & $[0.2391, 0.4105]$ \\
TCP h16 (cm) & Align+Cross & 4.899 & $[4.355, 5.482]$ & 4.989 & $[4.446, 5.561]$ \\
TCP h16 (cm) & Monolithic & 4.349 & $[3.925, 4.786]$ & 4.367 & $[3.948, 4.800]$ \\
\bottomrule
\end{tabular}
\par\medskip
\textbf{C. A+C minus Monolithic risk}\par\smallskip
\begin{tabular}{@{\hspace{3.5pt}}llrr@{\hspace{3.5pt}}}
\toprule
\textbf{Endpoint} & \textbf{Donor} & \textbf{Difference} & \textbf{95\% paired interval}\\
\tableheadrule
FT h4 & matched & 0.0138 & $[-0.0002, 0.0345]$ \\
FT h4 & hard & 0.0161 & $[0.0028, 0.0363]$ \\
TCP h16 (cm) & matched & 0.5503 & $[0.3699, 0.7485]$ \\
TCP h16 (cm) & hard & 0.6218 & $[0.4302, 0.8282]$ \\
\bottomrule
\end{tabular}
\end{table}

\paragraph{Action use and donor-count controls.}\label{app:rh-closure-25}
The additional 25-task validation bank holds checkpoints fixed while zeroing actions, replacing donors,
or averaging $K=1,2,4$ histories. Task-paired differences first average three
seeds, then use 10,000 task draws. Align + Cross has positive donor
penalties for FT and TCP; Cross has positive FT but unresolved TCP penalties.
The Align + Cross TCP donor effect decreases with donor count, while Random retains a
small positive FT effect. Ablation establishes action use. Table~\ref{tab:v3-rh-validation}
lists action, donor and history-count effects.

\begin{table}[H]
\centering\small
\setlength{\tabcolsep}{3pt}
\caption{\textbf{Action use and donor-count sensitivity on a separate 25-task bank.}
A: action-zero minus observed, and shuffled-minus-correct donor effects.
B: random-minus-correct effects at fixed donor count $K$ and fixed weights.
Pointwise 95\% task-bootstrap intervals follow averaging three sources.
FT uses MSE; TCP uses cm.}
\label{tab:v3-rh-validation}\label{tab:rh-closure-25}\label{tab:rh-multidonor-closure}
\textbf{A. Action and donor interventions}\par\smallskip
\begin{tabular}{@{\hspace{3.5pt}}llrl@{\hspace{3.5pt}}}\toprule
\textbf{Intervention/model} & \textbf{Metric} & \textbf{Difference} & \textbf{95\% task interval}\\\tableheadrule
A+C: action zero & FT h4 & 0.002074 & $[0.000740, 0.003650]$\\
A+C: action zero & TCP h4 & 0.06439 & $[0.04507, 0.08274]$\\
A+C: action zero & TCP h16 & 0.1399 & $[0.07687, 0.2025]$\\
Align + Cross & FT h4 & 0.009692 & $[0.004828, 0.01624]$\\
Align + Cross & TCP h4 & 0.4705 & $[0.2801, 0.6948]$\\
Align + Cross & TCP h16 & 0.3385 & $[0.1768, 0.5304]$\\
Cross & FT h4 & 0.01711 & $[0.009042, 0.02876]$\\
Cross & TCP h4 & 0.004189 & $[-0.02059, 0.02922]$\\
Cross & TCP h16 & 0.01571 & $[-0.0472, 0.07335]$\\
\bottomrule\end{tabular}
\end{table}
\begin{table}[H]
\centering\small
\setlength{\tabcolsep}{3pt}
\textbf{Table~\ref{tab:v3-rh-validation} (continued). B. Donor-count sensitivity at fixed weights}\par\smallskip
\begin{tabular}{@{\hspace{3.5pt}}llccc@{\hspace{3.5pt}}}\toprule
\textbf{Model} & \textbf{$K$} & \textbf{FT h4 MSE} & \textbf{TCP h4 (cm)} & \textbf{TCP h16 (cm)}\\\tableheadrule
Align + Cross & 1 & \shortstack{$0.01019$\\{\footnotesize$[0.00476, 0.01786]$}} & \shortstack{$0.4686$\\{\footnotesize$[0.2741, 0.6968]$}} & \shortstack{$0.3368$\\{\footnotesize$[0.1719, 0.5327]$}}\\\addlinespace[1pt]
Align + Cross & 2 & \shortstack{$0.00874$\\{\footnotesize$[0.00325, 0.01648]$}} & \shortstack{$0.3652$\\{\footnotesize$[0.1675, 0.6004]$}} & \shortstack{$0.2614$\\{\footnotesize$[0.09066, 0.4618]$}}\\\addlinespace[1pt]
Align + Cross & 4 & \shortstack{$0.00794$\\{\footnotesize$[0.00240, 0.01566]$}} & \shortstack{$0.2978$\\{\footnotesize$[0.08930, 0.5383]$}} & \shortstack{$0.2191$\\{\footnotesize$[0.04070, 0.4249]$}}\\\addlinespace[2pt]
Cross & 1 & \shortstack{$0.01606$\\{\footnotesize$[0.00836, 0.02703]$}} & \shortstack{$0.00488$\\{\footnotesize$[-0.01979, 0.03002]$}} & \shortstack{$0.01573$\\{\footnotesize$[-0.04830, 0.07455]$}}\\\addlinespace[1pt]
Cross & 2 & \shortstack{$0.01593$\\{\footnotesize$[0.00634, 0.02960]$}} & \shortstack{$0.00691$\\{\footnotesize$[-0.01343, 0.02892]$}} & \shortstack{$0.01599$\\{\footnotesize$[-0.02912, 0.05778]$}}\\\addlinespace[1pt]
Cross & 4 & \shortstack{$0.01602$\\{\footnotesize$[0.00531, 0.03114]$}} & \shortstack{$0.00869$\\{\footnotesize$[-0.01117, 0.02953]$}} & \shortstack{$0.01166$\\{\footnotesize$[-0.02378, 0.04752]$}}\\\addlinespace[2pt]
Random & 1 & \shortstack{$0.00071$\\{\footnotesize$[0.00017, 0.00133]$}} & \shortstack{$0.00162$\\{\footnotesize$[-0.00353, 0.00672]$}} & \shortstack{$0.00113$\\{\footnotesize$[-0.01151, 0.01315]$}}\\\addlinespace[1pt]
Random & 2 & \shortstack{$0.00067$\\{\footnotesize$[0.00017, 0.00126]$}} & \shortstack{$0.00110$\\{\footnotesize$[-0.00355, 0.00570]$}} & \shortstack{$-0.00018$\\{\footnotesize$[-0.01234, 0.01100]$}}\\\addlinespace[1pt]
Random & 4 & \shortstack{$0.00066$\\{\footnotesize$[0.00016, 0.00122]$}} & \shortstack{$0.00078$\\{\footnotesize$[-0.00368, 0.00521]$}} & \shortstack{$-0.00091$\\{\footnotesize$[-0.01283, 0.00794]$}}\\
\bottomrule\end{tabular}
\end{table}

\FloatBarrier
\subsection{Physical information in reader memory: Balls}
\label{app:balls-use}
RSSM reader memory selectively exposes restitution, and fixed-predictor substitutions show corresponding donor use.
The measured memory includes the trained support projection and is distinct
from complete source state $U$ and independent-support code $S$. Twelve RSSM fits (four recipes by three joint source/reader seeds)
reproduce stored matched error before one-factor donor substitution.
\begin{figure}[H]\centering
\includegraphics[width=\linewidth]{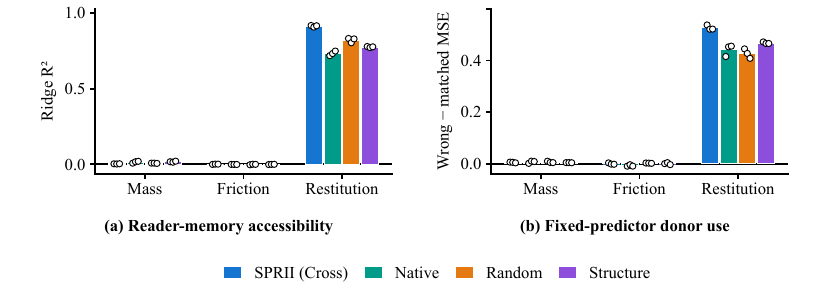}
\caption{\textbf{Restitution dominates reader-memory accessibility and donor effects.}
(a) Training-fitted ridge $R^2$ on RSSM reader memory.
(b) Wrong-one-factor minus matched MSE on 2,000 recipients through fixed
weights. Bars average three source/reader fits; symbols show individual fits.
SPRII uses Cross.}
\label{fig:v3-balls}
\end{figure}

For the third SPRII (Cross) replica, wrong restitution changes MSE from 1.1823 to 1.7033;
mass and friction changes are 0.00331 and $-0.00209$. Its within/between ratio
is 1.059: within-system code variation remains slightly larger than
between-system variation. Repeated-history and S3/S5/S8 inputs pass through the
same fixed S3-trained head, measuring its sensitivity to available histories.

\begin{table}[!htbp]\centering\small
\caption{\textbf{Distinct histories lower error through the fixed S3-trained RSSM reader.}
MSE averages three fits; Repeat averages repetitions of each individual
history. Reader weights remain fixed.}
\label{tab:v3-balls-histories}
\begin{tabular}{@{\hspace{3.5pt}}lrrrr@{\hspace{3.5pt}}}\toprule
\textbf{Recipe} & \textbf{Repeat} & \textbf{S3} & \textbf{S5} & \textbf{S8}\\\tableheadrule
SPRII (Cross) & 1.29 & 1.171 & 1.156 & 1.147\\
Native & 1.532 & 1.286 & 1.241 & 1.206\\
Random & 1.497 & 1.356 & 1.333 & 1.316\\
Structure & 1.517 & 1.288 & 1.246 & 1.218\\\bottomrule\end{tabular}\end{table}

JEPA and CoPhyNet factor substitutions also favor restitution; their
cohorts and replication counts are given with the corresponding results.

\paragraph{Actual-memory channels and fixed-reader substitutions.}
\label{app:records-balls-use}
The CoPhyNet probes below measure actual reader memory $M$; source $U$ and
independent-support $S$ are separate channels. Development-bank coverage is
specified in Appendix~\ref{app:replication-status}.
\begin{table}[!htbp]\centering\small\setlength{\tabcolsep}{5pt}
\caption{\textbf{Reader-memory probes, RSSM geometry and separate donor assays.}
A: training-fitted CoPhyNet memory probes, including the learned projection.
B: RSSM same-physics/wrong-physics code-distance ratios.
C: separate single-source evaluations, reporting wrong-factor minus matched
MSE. These panels measure distinct channels and cohorts.}
\label{tab:cophy-new5-probes}\label{tab:cophy-physical-assays}\label{tab:balls-existing-factor-pairs}
\textbf{A. CoPhyNet reader-memory probes ($R^2$)}\par\smallskip
\begin{tabular}{@{\hspace{3.5pt}}llrrr@{\hspace{3.5pt}}}\toprule
\textbf{Recipe} & \textbf{Factor} & \textbf{Source 1} & \textbf{Source 2} & \textbf{Source 3}\\\tableheadrule
SPRII (A+C) & Mass & 0.0509 & 0.0546 & 0.0678\\
 & Friction & 0.0027 & $-0.0020$ & 0.0002\\
 & Restitution & 0.8297 & 0.8266 & 0.8210\\
\addlinespace[3pt]
Native & Mass & 0.0283 & 0.0271 & 0.0411\\
 & Friction & $-0.0014$ & 0.0024 & 0.0009\\
 & Restitution & 0.7581 & 0.7599 & 0.7684\\
\addlinespace[3pt]
Random & Mass & 0.0306 & 0.0262 & 0.0295\\
 & Friction & $-0.0025$ & $-0.0043$ & 0.0007\\
 & Restitution & 0.7611 & 0.7681 & 0.7598\\
\bottomrule\end{tabular}
\par\medskip\textbf{B. RSSM within/between geometry}\par\smallskip
\begin{tabular}{@{\hspace{3.5pt}}lrrr@{\hspace{3.5pt}}}\toprule
\textbf{Recipe} & \textbf{Source 1} & \textbf{Source 2} & \textbf{Source 3}\\\tableheadrule
SPRII (Cross) & 1.182 & 1.373 & 1.059\\
Native & 3.904 & 1.769 & 2.493\\
Random & 2.370 & 2.429 & 2.334\\
Structure & 2.479 & 2.415 & 2.496\\\bottomrule\end{tabular}
\par\medskip\textbf{C. Separate source100 donor assay}\par\smallskip
\begin{tabular}{@{\hspace{3.5pt}}lrrrr@{\hspace{3.5pt}}}\toprule
\textbf{Learner} & \textbf{Matched MSE} & \textbf{$\Delta m$} & \textbf{$\Delta$ friction} & \textbf{$\Delta$ restitution}\\\tableheadrule
JEPA & 1.240 & $+0.000490$ & $+0.001173$ & 0.5140\\
CoPhyNet & 1.006 & $+0.004422$ & $-0.001568$ & 0.5184\\
\bottomrule\end{tabular}
\begin{papertablenotes}
Source $U$, independent support $S$ and actual reader memory $M$ are distinct
channels. A+C denotes Align + Cross.
Table~\ref{tab:v3-balls-histories} gives history-count means.
\end{papertablenotes}
\end{table}

\FloatBarrier
\subsection{Recipient-specific updates through frozen weights: Swimmer}\label{app:fixed-weight}
Recipient-specific persistent updates reduce error for three frozen predictors. The 512
prospective systems are disjoint from training and selection; averages cover
three adapter checkpoints, then 36 query/candidate rows per system.
Four thousand paired-system draws quantify uncertainty.

\begin{table}[H]
\centering\small
\setlength{\tabcolsep}{4pt}
\caption{\textbf{Recipient-specific updates improve all three frozen Swimmer predictors.}
Gain is wrong-system minus own-system update error in $10^{-5}$ units.
Intervals are paired-system 95\% bootstrap intervals on 512 new systems.}
\label{tab:v3-swimmer}\label{tab:ctp-fixed-weight}
\begin{tabular}{lrrr}\toprule
\textbf{Frozen base} & \textbf{Gain} & \textbf{95\% interval} & \textbf{Positive systems}\\
\tableheadrule
Persistent-JEPA & 5.329 & $[3.930, 6.724]$ & 65.04\% \\
Masked-GRU (64101) & 6.930 & $[5.415, 8.621]$ & 67.97\% \\
Masked-GRU (64103) & 6.661 & $[4.878, 8.486]$ & 68.36\% \\
\bottomrule\end{tabular}
\end{table}

Each base retains its architecture and training normalization. The
within-base comparison measures the value of its recipient-specific update
through fixed predictor weights.

\label{app:records-swimmer-use}\label{app:records-swimmer}

Task benefit under different targets and fitted readouts is examined in
Appendix~\ref{app:conditional-value}.

\FloatBarrier
\section{Value: Conditional Task Benefit and Reader Realization}\label{app:records-value-group}
\label{app:conditional-value}\label{app:value-details}
This section expands Figure~\ref{fig:information-use}:
SpringWorld tests horizon and reader effects, D-Clean contrasts direct context
with decoded parameters, and PokeWorld separates aggregate from component-level
benefit. Population and replication details follow Appendix~\ref{app:replication-status}.

\FloatBarrier
\subsection{Prediction horizon and recipient evidence: SpringWorld}
\label{app:horizon-conditions}
History benefit grows toward longer horizons for both cold and moving queries
in the 178-system common-adapter grid. Both query types have zero observed
recipient transitions and differ in initial motion. The comparison spans four
strata, three sources and three readers per source.

\begin{table}[!htbp]\centering\small
\caption{\textbf{History benefit increases with prediction horizon.}
Percentage error reduction of separately fitted Persistent versus Null readers,
with pointwise stratified system-bootstrap 95\% intervals.}
\label{tab:v3-horizon}
\begin{tabular}{@{\hspace{3.5pt}}lrr@{\hspace{3.5pt}}}\toprule
\textbf{Horizon} & \textbf{Cold: gain [95\% interval]} & \textbf{Moving: gain [95\% interval]}\\\tableheadrule
1 & $1.95\,[-8.47, 11.01]$ & $1.15\,[-0.83, 3.01]$\\
2 & $13.39\,[3.93, 21.49]$ & $1.98\,[-0.22, 4.02]$\\
4 & $23.19\,[14.37, 30.35]$ & $3.23\,[0.91, 5.77]$\\
8 & $35.20\,[26.74, 41.92]$ & $11.62\,[7.17, 16.57]$\\
16 & $35.68\,[28.22, 42.27]$ & $29.73\,[22.68, 36.45]$\\\bottomrule
\end{tabular}\end{table}

The 36.10\% cold-start and 7.11\% full-mixture summaries use different
populations of 100 and 178 systems, respectively. The following temporal-reader
comparison varies recipient evidence under a separate protocol.

\paragraph{Recipient-prefix manipulation.}\label{app:spring-recipient-evidence}
A separate temporal reader with $r\in\{0,1,2,4,8\}$ does not materially
change the recipient-only baseline. At h16, r0/r8 errors are 0.469305/.467214
for Null, 0.334363/.333367 for Persistent and 0.164647/.156172 for Oracle.
The observed endpoint changes are small under this temporal-reader protocol.

\label{app:records-value}
The recipient-prefix reader uses 100 validation systems and three sources by
three readers. Its endpoint intervals use 4,000 system draws and generally
include zero.

\FloatBarrier
\subsection{Direct context versus physical bottlenecks}
\label{app:horizon}\label{app:spring-bottleneck}\label{app:spring-same-donor}
\takeaway{Direct context outperforms a decoded-parameter interface in D-Clean; the same comparison favors decoded parameters in SpringWorld.} Four fresh readers share the
frozen Align + Cross query encoder: Null, Persistent, Decode and Oracle.
Persistent and Decode receive exactly the same independent donor history;
the train-fitted parameter decoder freezes before reader fitting. Each arm
uses the same 10,000-step horizon-conditioned recipe, three sources and three
readers, with 32 windows from each of 200 validation systems. This population
is separate from the 100-system TDS comparison.

\begin{figure}[!htbp]\centering
\includegraphics[width=\linewidth]{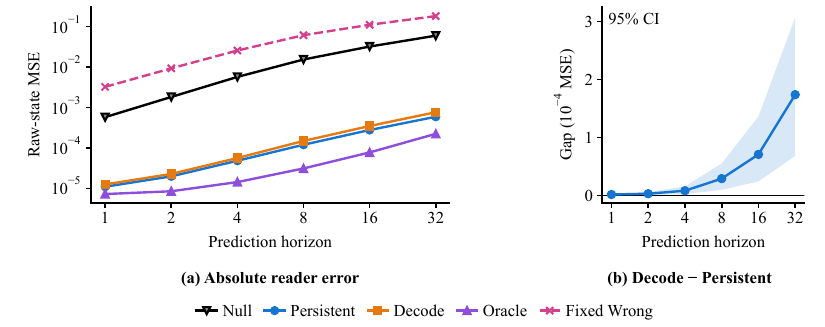}
\caption{\textbf{Direct context outperforms the physical bottleneck across D-Clean horizons.}
(a) Raw-state MSE (log scale); Fixed Wrong substitutes donors through the
Persistent reader. (b) Decode-minus-Persistent MSE with pointwise
system-bootstrap 95\% intervals, conditional on fitted sources and readers.
The primary endpoint is h32.}
\label{fig:v3-dclean-readers}
\end{figure}

At h32, Persistent MSE is 0.000590019 versus 0.000763955 for Decode and
0.000223496 for Oracle. The paired Decode-minus-Persistent gap is 0.000173937
[0.000068237,0.000307406], positive in all nine cells. Independent validation
of the frozen decoder gives $R^2=0.996715,0.997340,0.996993$ on systems
1000--1199. Accurate parameter estimation therefore need not supply a
predictively equivalent learned interface.

\paragraph{SpringWorld's opposite ordering.}
The common-adapter comparison projects the same donor code either directly
or through training-only ridge estimates of three log-parameters. All arms
share query encoder, donor identities, sample order and target normalization;
Decode/Oracle add only the 256-parameter input projection to the common reader.
On the 100-system cold-start bank, Decode-minus-Persistent is $-0.0023623$
[$-0.0045732,-0.0004762$]: explicit physical parameters preserve and slightly
improve realized value here. Persistent reduces Null error by 36.10\%, and
Persistent error is 2.12\% higher than Decode error, using the same
Decode-denominator convention as the D-Clean comparison. Table~\ref{tab:spring-same-donor} gives actual-donor probes on 178 systems
and separate full-mixture results.

\label{app:records-bottlenecks}
\begin{table}[!htbp]
\centering\small\setlength{\tabcolsep}{5pt}
\caption{\textbf{Direct-context and bottleneck readers use the same D-Clean donors.}
(A) Raw-state MSE; (B) pointwise paired-system 95\% intervals conditional
on the fitted source--reader grid; (C) independent validation of the frozen drag decoder.
The primary endpoint is h32.}
\label{tab:v3-bottleneck-endpoints}\label{tab:ctp-horizon}\label{tab:dclean-same-donor}\label{tab:dclean-fixed-wrong}\label{tab:dclean-gamma-validation}
\textbf{A. Absolute reader and intervention error}\par\smallskip
\begin{tabular}{@{\hspace{3.5pt}}rrrrrr@{\hspace{3.5pt}}}\toprule
\textbf{Horizon} & \textbf{Null} & \textbf{Persistent} & \textbf{Decode} & \textbf{Oracle} & \textbf{Fixed Wrong}\\\tableheadrule
1 & 0.000577 & 0.000011 & 0.000013 & 0.000007 & 0.003227\\
2 & 0.001821 & 0.000020 & 0.000023 & 0.000009 & 0.009263\\
4 & 0.005689 & 0.000048 & 0.000056 & 0.000014 & 0.025560\\
8 & 0.015320 & 0.000120 & 0.000149 & 0.000031 & 0.061030\\
16 & 0.031970 & 0.000278 & 0.000349 & 0.000078 & 0.110300\\
32 & 0.059360 & 0.000590 & 0.000764 & 0.000223 & 0.180900\\\bottomrule
\end{tabular}\par\medskip
\textbf{B. Paired contrasts and intervention uncertainty}\par\smallskip
\begin{tabular}{@{\hspace{3.5pt}}rrrr@{\hspace{3.5pt}}}\toprule
\textbf{Horizon} & \textbf{Fixed Wrong 95\% CI} & \textbf{D$-$P ($10^{-4}$)} & \textbf{D$-$P 95\% CI ($10^{-4}$)}\\\tableheadrule
1 & $[0.002261, 0.004323]$ & 0.0154 & $[0.0059, 0.0281]$\\
2 & $[0.006822, 0.01199]$ & 0.0291 & $[0.0076, 0.0586]$\\
4 & $[0.01949, 0.03224]$ & 0.0809 & $[0.0218, 0.1565]$\\
8 & $[0.04752, 0.07559]$ & 0.2898 & $[0.0965, 0.5504]$\\
16 & $[0.08612, 0.1362]$ & 0.7078 & $[0.2409, 1.3620]$\\
32 & $[0.1433, 0.2213]$ & 1.7390 & $[0.6824, 3.0740]$\\\bottomrule
\end{tabular}\par\medskip
\textbf{C. Independent frozen-decoder validation}\par\smallskip
\begin{tabular}{@{\hspace{3.5pt}}rrrr@{\hspace{3.5pt}}}\toprule
\textbf{Sampling seed} & \textbf{Validation $R^2$} & \textbf{$\gamma$ MSE} & \textbf{95\% MSE interval}\\\tableheadrule
0 & 0.9967 & 0.003653 & $[0.0027, 0.0052]$\\
1 & 0.9973 & 0.002957 & $[0.0022, 0.0043]$\\
2 & 0.9970 & 0.003344 & $[0.0025, 0.0046]$\\\bottomrule
\end{tabular}
\begin{papertablenotes}
D$-$P is Decode minus Persistent. Fixed Wrong changes the Persistent input,
without fitting a fifth reader. The grid contains 27 Null/Persistent/Oracle
fits and nine Decode fits; no per-arm source SD is reconstructed.
\end{papertablenotes}
\end{table}

\begin{table}[!htbp]
\centering\small\setlength{\tabcolsep}{5pt}
\caption{\textbf{SpringWorld reader error and donor-factor accessibility.}
(A) Common-adapter MSE across three sources and three readers.
(B) Training-fitted log-parameter probes on 712 donor histories from 178
validation systems, weighted equally by system.}
\label{tab:spring-same-donor}\label{tab:spring-frozen-probes}
\textbf{A. Reader risk: cold-start and full-mixture populations}\par\smallskip
\begin{tabular}{@{\hspace{3.5pt}}lrrrr@{\hspace{3.5pt}}}\toprule
\textbf{Population} & \textbf{Null} & \textbf{Persistent} & \textbf{Decode} & \textbf{Oracle}\\\tableheadrule
Cold start: 100 systems / 600 cases & 0.17780 & 0.11360 & 0.11130 & 0.06188\\
Full mixture: 178 systems & 0.87170 & 0.80970 & 0.80870 & 0.76590\\\bottomrule
\end{tabular}\par\medskip
\textbf{B. Actual-donor accessibility: 178 systems}\par\smallskip
\begin{tabular}{@{\hspace{3.5pt}}llrrrr@{\hspace{3.5pt}}}\toprule
\textbf{Source recipe} & \textbf{Seed} & \textbf{$R^2_m$} & \textbf{$R^2_\gamma$} & \textbf{$R^2_k$} & \textbf{$R^2_{k/m}$}\\\tableheadrule
Align + Cross & 0 & 0.3918 & 0.1754 & 0.2050 & 0.4842\\
 & 1 & 0.3112 & 0.1537 & 0.1421 & 0.3774\\
 & 2 & 0.3569 & 0.1618 & 0.2345 & 0.5115\\\midrule
FCRL-style temporal & 0 & $-0.0000$ & $-0.0019$ & $-0.0180$ & $-0.0091$\\
 & 1 & 0.0107 & $-0.0073$ & $-0.0349$ & $-0.0154$\\
 & 2 & 0.0030 & 0.0013 & $-0.0134$ & $-0.0056$\\\bottomrule
\end{tabular}
\begin{papertablenotes}
Cold-start Decode-minus-Persistent is
$-0.0023623$ $[-0.0045732, -0.0004762]$. The $k/m$ estimate derives from fitted
mass/stiffness coordinates. The probe bank in B is distinct from the 100-system
cold-start risk population in A.
\end{papertablenotes}
\end{table}

\FloatBarrier
\subsection{Aggregate and component-level benefit: PokeWorld}
\label{app:poke-value}\label{app:frozen-head}
Persistent context improves object-position and velocity prediction, with benefits that vary across output components.
The R8 source study fits separate Null, Persistent, Shuffled and Oracle
readers with the same architecture and training schedule. All 36 combinations
use three sources by three reader seeds, 10,000 updates, and the reused
400-system development bank with 16 windows per system. Shuffled fits a
separate reader using wrong-context inputs.
\begin{figure}[!htbp]\centering
\begin{minipage}[t]{\dimexpr\linewidth-2.8in-0.15in\relax}
\vspace{0pt}\raggedright
At the prespecified h16 endpoint, object-position and velocity errors fall,
while finger-position error rises. The aggregate Null-minus-Persistent
interval spans zero; Persistent improves on the separately trained
Shuffled reader.

\smallskip
Actual donor-window drag readout is strong ($R^2=0.747$--$0.783$), compared
with $0.131$--$0.164$ for query-only inputs. The Oracle's $0.200\%$ and
Persistent's $0.057\%$ aggregate improvements describe this learned
reader family.
\end{minipage}\hfill
\begin{minipage}[t]{2.8in}
\vspace{0pt}\centering
\includegraphics[width=2.8in]{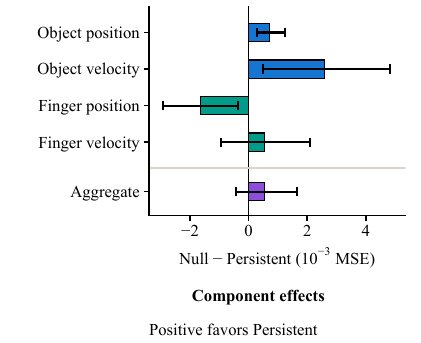}
\end{minipage}
\caption{\textbf{Persistent context improves object-state prediction, with mixed effects on finger state.}
Null-minus-Persistent MSE on the development bank; positive favors Persistent.
Pointwise paired-system 95\% intervals condition on the fitted source--reader
grid; the aggregate interval adjusts for three contrasts. Components are not additive.}
\label{fig:v3-poke-components}
\end{figure}
\begin{table}[!htbp]\centering
\begingroup\makeatletter\def\@captype{table}\makeatother
\centering\small\setlength{\tabcolsep}{6pt}
\caption{\textbf{Aggregate PokeWorld gains are small despite component-level benefits.}
Separately fitted readers on 400 development systems; aggregate 95\%
intervals adjust for three contrasts. Oracle receives true parameters.}
\label{tab:frozen-head}\label{tab:frozen-head-components}
\begin{minipage}[t]{0.30\linewidth}\centering
\textbf{A. Aggregate h16 MSE}\par\smallskip
\begin{tabularx}{\linewidth}{@{\hspace{3.5pt}}Xr@{\hspace{3.5pt}}}\toprule
\textbf{Arm} & \textbf{MSE}\\\tableheadrule
Null & 0.9629\\
\textbf{Persistent} & 0.9624\\
Shuffled & 0.9637\\
Oracle & 0.961\\\bottomrule
\end{tabularx}
\end{minipage}\hfill
\begin{minipage}[t]{0.68\linewidth}\centering
\textbf{B. Prespecified aggregate contrasts}\par\smallskip
\begin{tabular}{@{\hspace{3.5pt}}lrr@{\hspace{3.5pt}}}\toprule
\textbf{Contrast} & \textbf{Difference} & \textbf{Adjusted 95\% interval}\\\tableheadrule
Null--Persistent & 0.000553 & $[-0.000419, 0.001641]$\\
Null--Oracle & 0.001927 & $[0.000427, 0.003615]$\\
Shuffled--Persistent & 0.001367 & $[0.000274, 0.002498]$\\\bottomrule
\end{tabular}
\end{minipage}\par\medskip
\noindent\makebox[\linewidth][c]{\begin{minipage}{0.68\linewidth}\centering
\textbf{C. Oracle component reference: Null--Oracle}\par\smallskip
\begin{tabular*}{\linewidth}{@{\hspace{3.5pt}\extracolsep{\fill}}lrr@{\hspace{3.5pt}}}\toprule
\textbf{Component} & \textbf{Position} & \textbf{Velocity}\\\tableheadrule
Object & 0.002809 & 0.006168\\
Finger & $-0.003009$ & 0.001740\\\bottomrule
\end{tabular*}\par
\end{minipage}}
\endgroup

\end{table}

\paragraph{Factor-targeted utility.}\label{app:poke-targeted-utility}
The 36-reader $G_1$/$G_2$ development grid pairs task gains with actual-donor
physical probes. Its utility-versus-simulator-sensitivity slope is 0.002810
$[-0.004443, 0.009203]$, with positive slopes in each source. $G_1$ retains higher
mass readout and $G_2$ higher drag readout in 5,913 actual donor windows.
Table~\ref{tab:poke-targeted-utility} gives the slopes and donor probes.
Additional population checks are grouped in Appendix~\ref{app:records-followup}.

\paragraph{Aggregate, component and factor-targeted value.}
\label{app:records-poke-value}
For targeted utility, $\Delta U=(E_N-E_P)_{G_1}-(E_N-E_P)_{G_2}$ is regressed against $d=z_{\rm train}(S_m)-z_{\rm train}(S_\gamma)$. Simulator sensitivities use central $\pm1\%$ perturbations, fixed query/actions and training factor/target scales. Actual-donor probes cover 5,913 distinct windows, equally weighting systems, with ridge coefficient 1.
\begin{table}[H]
\centering\small
\caption{\textbf{PokeWorld factor-targeted utility and donor accessibility.}
(A) The $G_1$-minus-$G_2$ history-benefit slope against simulator sensitivity
has a system-bootstrap interval spanning zero. (B) Probe $R^2$ on actual donor
windows, averaged across source seeds. The development grid includes the initial fits.}
\label{tab:poke-targeted-utility}
\begin{tabular}{lrrr}
\toprule
\multicolumn{4}{l}{\textbf{A. Continuous slope of $\Delta U$ against sensitivity dominance}}\\
\textbf{Source seed} & \textbf{0} & \textbf{1} & \textbf{2}\\
\tableheadrule
Expanded-grid slope & 0.003838 & 0.003307 & 0.001284\\
\multicolumn{4}{l}{Mean: $0.002810$; 95\% system-bootstrap interval: $[-0.004443, 0.009203]$}\\
\bottomrule
\end{tabular}
\par\medskip
\begin{tabular}{lrrrr}
\toprule
\multicolumn{5}{l}{\textbf{B. Train-fitted donor-code probe, validation $R^2$}}\\
\textbf{Recipe} & \textbf{$\log m$} & \textbf{$\log\gamma$} & \textbf{$\log k$} & \textbf{$\log(k/m)$}\\
\tableheadrule
$G_1$ & 0.20700 & 0.12460 & 0.08187 & 0.24440\\
$G_2$ & 0.07075 & 0.78890 & 0.07491 & 0.14640\\
\bottomrule
\end{tabular}
\end{table}

\FloatBarrier
\subsection{Realizing value through frozen-source readouts}
\label{app:realization-slot}\label{app:downstream-realization}
Changing the consuming route recovers additional value from frozen SpringWorld
context, while the PokeWorld comparisons show that this recovery is not automatic.

\paragraph{Reader recovery across source families.}
The $M_1/M_2$ formulas are defined in Appendix~\ref{app:reader-interfaces}.
Nine Align + Cross pairs cover three sources by three readers; Structure has
one source and three readers. Poke $G_1$/$G_2$ use source 2 with two/three readers.
Checkpoints are selected on development data. Align + Cross favors $M_2$ in
all nine standard and physically weighted pairs; Poke favors $M_1$ in all
five. Table~\ref{tab:realization-complete-grid} reports descriptive fitted-grid
means. The routes share their architecture, but their active
capacity is not exactly matched (Appendix~\ref{app:reader-interfaces}).

\begin{figure}[!htbp]\centering
\includegraphics[width=\linewidth]{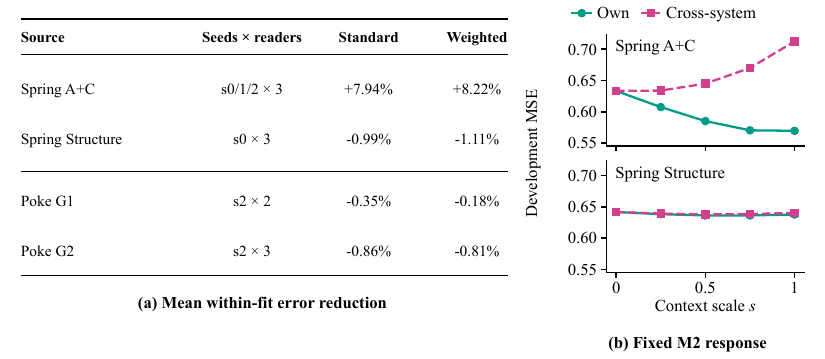}
\caption{\textbf{Reader recovery depends on source family and donor context.}
(a) Mean within-fit error reduction, $100[E(M_1)-E(M_2)]/E(M_1)$;
positive favors $M_2$. Reader percentages are averaged within source, then
across sources. (b) SpringWorld raw MSE under own- and cross-system donors
through fixed $M_2$ readers; $s=0$ gives the recipient-only base $R_0$.}
\label{fig:v3-m1m2}
\end{figure}
\begin{table}[!htbp]\centering
\begingroup\makeatletter\def\@captype{table}\makeatother
\centering\small\setlength{\tabcolsep}{4.5pt}
\caption{\textbf{Reader recovery differs across frozen source families.}
Descriptive mean errors and gains $E(M_1)-E(M_2)$; positive favors $M_2$.
Weighted gains use the separate physically weighted route.
These fitted-run means do not estimate uncertainty.}
\label{tab:realization-complete-grid}\label{tab:realization-routes}
\begin{tabular}{@{\hspace{3.5pt}}llrrrr@{\hspace{3.5pt}}}\toprule
\textbf{Family} & \textbf{Sources $\times$ readers} & \textbf{$M_1$} & \textbf{$M_2$} & \textbf{Gain} & \textbf{Weighted gain}\\\tableheadrule
Align + Cross & $3\times3$ & 0.6185 & 0.5695 & 0.04901 & 0.05059\\
Structure & $1\times3$ & 0.6316 & 0.6378 & $-0.006213$ & $-0.006971$\\
Poke $G_1$ & $1\times2$ & 0.8099 & 0.8127 & $-0.002805$ & $-0.001495$\\
Poke $G_2$ & $1\times3$ & 0.7919 & 0.7987 & $-0.006841$ & $-0.006443$\\\bottomrule
\end{tabular}
\begin{papertablenotes}
All nine Align + Cross pairs favor $M_2$
in both route variants; all five Poke pairs favor $M_1$. Figure~\ref{fig:v3-m1m2}
shows within-fit percentages, while this table retains absolute differences.
\end{papertablenotes}
\endgroup

\end{table}

\paragraph{Matched-architecture donor control.}
\label{app:reader-capacity-control}
Matched context lowers error relative to a separately fitted mismatched
reader with the same $M_2$ architecture. The comparison reuses the complete
three-source by three-reader $M_1/M_2$ grid. Each mismatched reader receives
1,000 AdamW updates with batch size 256 and learning rate $10^{-3}$;
checkpoints are selected every 100 updates on the original development
population. The source and its fitted recipient-only base $R_0$ remain
frozen. A fixed different-system donor map applies during training,
checkpoint selection and evaluation, with recipient inputs and targets
unchanged. The original $M_1/M_2$ predictions are recomputed on the same
records. The physical probes in Table~\ref{tab:spring-frozen-probes} describe
these same frozen source checkpoints.

\begin{figure}[!htbp]\centering
\includegraphics[width=\linewidth]{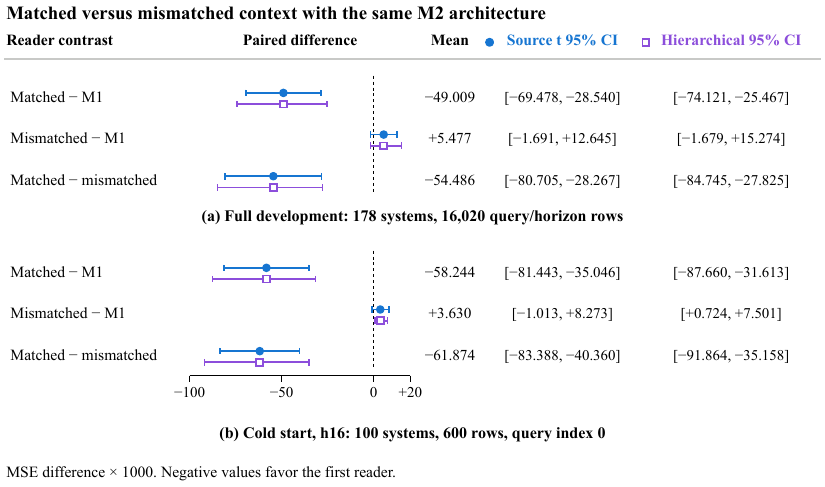}
\caption{\textbf{Paired error differences for the matched-architecture control.}
(a) Full development mixture; (b) cold-start h16. Negative values favor the
first reader ($10^{-3}$ MSE). Source-paired 95\% $t$ intervals use three
source means; hierarchical 95\% intervals resample sources, readers and
systems. Both condition on development-selected fits; matched and mismatched
$M_2$ readers are fitted separately.}
\label{fig:reader-capacity-contrasts}
\end{figure}

Table~\ref{tab:reader-capacity-errors} separates the full development
mixture from its cold-start h16 endpoint. Matched $M_2$ has lower error than
mismatched $M_2$ in all nine source--reader pairs at both endpoints, by
8.7\% and 33.7\% in the respective means. Mismatched $M_2$ improves on
$M_1$ in none of the nine pairs. Figure~\ref{fig:reader-capacity-contrasts}
retains both source-paired and hierarchical intervals: for mismatched
$M_2-M_1$, the source-paired intervals include zero at both endpoints,
whereas the cold-start hierarchical interval is positive. These
results identify a benefit of correctly associated context within this
architecture and development protocol.

\begin{table}[!htbp]\centering\small
\caption{\textbf{Matched context lowers error with identical $M_2$ architecture.}
Task MSE, mean $\pm$ sample SD across three sources, each averaging three
readers. Both populations are development data: 178 systems for the full
mixture and 100 for cold-start h16.}
\label{tab:reader-capacity-errors}
\setlength{\tabcolsep}{5pt}
\begin{tabular}{@{\hspace{3.5pt}}lccc@{\hspace{3.5pt}}}\toprule
\textbf{Endpoint} & \textbf{$M_1$} & \textbf{$M_2$ matched} & \textbf{$M_2$ mismatched}\\\tableheadrule
Full development & \msd{0.6185}{0.009762} & \msd{0.5695}{0.01796} & \msd{0.6240}{0.007561}\\
Cold start, h16 & \msd{0.1799}{0.006272} & \msd{0.1216}{0.01436} & \msd{0.1835}{0.008084}\\
\bottomrule\end{tabular}\end{table}

\paragraph{Stronger formation does not ensure reader gains.}
\label{app:alpha-m2}
Increasing relation reliability strengthens organization and drag accessibility,
but it does not produce a positive mean $M_2$ gain at the 5,000-update h16
endpoint under this reader protocol (Table~\ref{tab:external-alpha}).
Increasing the fitting budget reduces absolute error. Each
$\alpha\in\{0,0.5,1\}$ uses three frozen 20,000-update PokeWorld sources and
three readers per source. Unlike post-fitting scale $s$, $\alpha$ changes
source training relations. Each paired contrast compares readers within a
fixed source; query encoders vary across $\alpha$.

The reader predicts train-standardized eight-state displacements on 400
systems from the existing validation bank. The recipient-only base $R_0$ and each route
fit for 1,000 or 5,000 updates, with batch size 128 and learning rate 0.001,
using the same 5,913 windows and development selection rule. Sources and
cached observations are identical between budgets; geometry is unchanged
and refitted probe differences are below $10^{-9}$. Uncertainty uses the
paired source/system bootstrap in Table~\ref{tab:external-alpha}.
At 5,000 updates, all three wrong-minus-matched intervals include zero.
Zero-context and wrong-donor interventions measure channel dependence and
content specificity, respectively.
\begin{table}[!htbp]\centering\small\setlength{\tabcolsep}{5pt}
\caption{\textbf{Stronger formation does not ensure greater reader value.}
h16 gain is $E(M_1)-E(M_2)$ in $10^{-3}$ normalized MSE; positive favors
$M_2$. Budgets share frozen sources and development systems, averaging
three readers per source. B/W uses training-standardized latent statistics.}
\label{tab:external-alpha}\label{tab:external-alpha-records}
\textbf{A. Organization, accessibility and pooled reader value}\par\smallskip
\begin{tabular}{@{\hspace{3.5pt}}rrrrrr@{\hspace{3.5pt}}}\toprule
\textbf{$\alpha$} & \textbf{B/W} & \textbf{$R^2_{\log\gamma}$} & \textbf{1k gain} & \textbf{5k gain} & \textbf{5k 95\% interval}\\\tableheadrule
0 & 0.320 & 0.101 & $-2.512$ & $-1.890$ & $[-8.366, +3.116]$\\
0.5 & 1.042 & 0.254 & $+0.061$ & $-2.942$ & $[-8.255, +1.948]$\\
1 & 3.796 & 0.797 & $-4.469$ & $-6.291$ & $[-12.010, -0.328]$\\\bottomrule
\end{tabular}\par\medskip
\textbf{B. Source-specific reader value}\par\smallskip
\begin{tabular}{@{\hspace{3.5pt}}rrrrr@{\hspace{3.5pt}}}\toprule
\textbf{$\alpha$} & \textbf{Reader updates} & \textbf{Source 0} & \textbf{Source 1} & \textbf{Source 2}\\\tableheadrule
0 & 1000 & $-3.463$ & $-3.017$ & $-1.057$\\
0.5 & 1000 & $+1.222$ & $-0.015$ & $-1.023$\\
1 & 1000 & $-3.310$ & $-4.186$ & $-5.910$\\\midrule
0 & 5000 & $-4.739$ & $-1.617$ & $+0.686$\\
0.5 & 5000 & $-2.123$ & $-3.436$ & $-3.267$\\
1 & 5000 & $-5.288$ & $-11.270$ & $-2.314$\\\bottomrule
\end{tabular}
\begin{papertablenotes}
Intervals use 5,000 paired source/system bootstrap draws and are descriptive
with three source seeds.
\end{papertablenotes}
\end{table}

\FloatBarrier
\paragraph{Donor-specific responses through fixed readers.}
\label{app:records-realization}
SpringWorld donor substitutions keep the fitted source and reader weights
fixed while varying post-fitting context scale $s$. Own-system context lowers
Align + Cross error as its scale increases, whereas whole-system cross donors
raise it (Figure~\ref{fig:v3-m1m2}(b); Table~\ref{tab:realization-replacements}).
This difference supports content-specific use through the fixed reader.
The single-source Structure comparison has smaller, nonmonotone own-context
effects. At $s=0$, $M_2$ equals the recipient-only base $R_0$, not $M_1$;
standard and physically weighted routes are evaluated separately.
\begin{table}[!htbp]\centering\small\setlength{\tabcolsep}{5pt}
\caption{\textbf{Donor identity changes the response to context scale.}
Development MSE as persistent inputs are multiplied by $s$, with donor
substitutions through fixed readers. At $s=0$, the route equals the
recipient-only base $R_0$, not $M_1$.}
\label{tab:realization-replacements}
\begin{tabular}{@{\hspace{3.5pt}}llrrrrr@{\hspace{3.5pt}}}\toprule
\multicolumn{7}{@{}l}{\textbf{A. Standard $M_2$; displayed in Figure~\ref{fig:v3-m1m2}}}\\[2pt]
\textbf{Source} & \textbf{Donor} & \textbf{$s=0$} & \textbf{$0.25$} & \textbf{$0.5$} & \textbf{$0.75$} & \textbf{$1$}\\\tableheadrule
Align + Cross & Own & 0.6333 & 0.6074 & 0.5851 & 0.5703 & 0.5695\\
 & Cross-system & 0.6333 & 0.6339 & 0.6450 & 0.6700 & 0.7125\\
Structure & Own & 0.6419 & 0.6384 & 0.6363 & 0.6364 & 0.6378\\
 & Cross-system & 0.6419 & 0.6394 & 0.6381 & 0.6386 & 0.6404\\\midrule
\multicolumn{7}{@{}l}{\textbf{B. Physically weighted $M_{2,\mathrm{phys}}$}}\\[2pt]
\textbf{Source} & \textbf{Donor} & \textbf{$s=0$} & \textbf{$0.25$} & \textbf{$0.5$} & \textbf{$0.75$} & \textbf{$1$}\\\tableheadrule
Align + Cross & Own & 0.6333 & 0.6090 & 0.5859 & 0.5684 & 0.5660\\
 & Cross-system & 0.6333 & 0.6337 & 0.6434 & 0.6671 & 0.7102\\
Structure & Own & 0.6419 & 0.6387 & 0.6365 & 0.6361 & 0.6372\\
 & Cross-system & 0.6419 & 0.6394 & 0.6376 & 0.6376 & 0.6388\\\bottomrule
\end{tabular}
\begin{papertablenotes}
Route-selection and fixed-intervention evaluators remain separate.
\end{papertablenotes}
\end{table}

\FloatBarrier
\par\noindent\begin{minipage}{\linewidth}
\paragraph{Fresh-population consistency check.}
\label{app:records-poke-followup}\label{app:records-followup}
The preliminary Poke study uses 100 new systems with 16 windows each, six frozen source
exports and twelve paired readers. Source slopes
$+0.009202, -0.168619, +0.131394$ vary in sign. The study therefore ended at the
prespecified consistency check, before the subsequent intervention. These
fresh systems form a separate population from the earlier reader comparisons.
\end{minipage}\par

\paragraph{Object-only PokeWorld readouts.}
A separate single-source comparison restricts the target to four-dimensional
object state. Its R8 $M_2-M_1$ interval spans zero, while $G_2$ has higher $M_2$
mean error (Table~\ref{tab:poke-object-only-config}). These independently
selected readers do not establish additional recovery on this development
population.
\begin{table}[!htbp]\centering\small
\caption{\textbf{Object-only PokeWorld reader comparisons.}
h16 object-state MSE on 400 development systems for one source--reader pair.
$M_1$ and $M_2$ are selected independently; the paired interval is conditional
on these fits and evaluation systems.}
\label{tab:poke-object-only-config}
\begin{tabular}{@{\hspace{3.5pt}}lrrr@{\hspace{3.5pt}}}\toprule
\textbf{Source} & \textbf{$M_2$} & \textbf{$M_1$} & \textbf{Recipient-only}\\\tableheadrule
R8 & 0.9011 & 0.9026 & 0.9118\\
$G_2$ & 0.9132 & 0.9085 & 0.9080\\\bottomrule
\end{tabular}
\begin{papertablenotes}
R8 $M_2-M_1$: $-0.001497$ $[-0.011855, 0.008400]$.
\end{papertablenotes}
\end{table}

\FloatBarrier
\subsection{Closed-loop control with a development-selected recipe: Pendulum}
\label{app:pendulum-control}
The complete R7 recipe is selected on ID control development with relation
weight 0.003 and variance target 0.1, then held fixed across ID and the four
OOD groups. Three training seeds each contribute ten 200-step episodes per
group. Table~\ref{tab:pendulum-r7} reports the return and success rate from
those same trajectories. Success means that all of the final 100 post-action
states lie within $60^\circ$ of upright; this criterion is unchanged.
\begin{table}[!htbp]\centering\small
\caption{\textbf{Pendulum control with one ID-selected R7 recipe.}
Mean return $\pm$ sample SD across three seeds, with success percentage in
parentheses; higher is better. The same recipe is used for all OOD groups.}
\label{tab:pendulum-r7}
\setlength{\tabcolsep}{4pt}
\begin{tabular}{@{\hspace{3.5pt}}lccc@{\hspace{3.5pt}}}\toprule
\textbf{Group} & \textbf{CaDM} & \textbf{Random relation} & \textbf{SPRII (R7)}\\\tableheadrule
ID & $-326.5\sdev{46.5}$ (96.7\%) & $-321.4\sdev{33.2}$ (100.0\%) & $-323.9\sdev{26.2}$ (100.0\%)\\
OOD\_c0 & $-376.6\sdev{63.4}$ (93.3\%) & $-346.8\sdev{30.6}$ (100.0\%) & $-322.2\sdev{29.4}$ (100.0\%)\\
OOD\_c1 & $-387.0\sdev{11.9}$ (93.3\%) & $-400.1\sdev{24.6}$ (93.3\%) & $-381.0\sdev{13.6}$ (100.0\%)\\
OOD\_c2 & $-355.5\sdev{33.9}$ (90.0\%) & $-479.6\sdev{126.7}$ (76.7\%) & $-370.2\sdev{51.8}$ (90.0\%)\\
OOD\_c3 & $-1100\sdev{273.5}$ (20.0\%) & $-1185\sdev{131.2}$ (6.7\%) & $-1060\sdev{384.4}$ (30.0\%)\\
\bottomrule\end{tabular}\end{table}
\FloatBarrier

R7 has higher mean return than CaDM in four of five groups, but lower return
in OOD c2; the direction is not shared by all seeds in every group. The
return and success endpoints remain separate. The earlier complete relation
recipe is not combined columnwise with R7; R7 did not improve every group
over that earlier recipe.

Same-checkpoint ID probes give mass $R^2$ of 0.027, 0.012 and 0.050 for CaDM,
Random and SPRII, respectively, and length $R^2$ of 0.670, 0.683 and 0.669.
The corresponding common-bank own-donor H10 MSEs are 0.0181, 0.0226 and
0.0190; wrong-minus-matched errors are 0.1132, 0.0956 and 0.1018.
These logged-action, fixed-model interventions test context use separately
from online control. The weak mass probes do not establish recovery of all
physical parameters.

\FloatBarrier

\section{Generality Across Learners, Partners, and Fields}\label{app:extensions}\label{app:broader-results}\label{app:records-generality}
CoPhy varies learner and scene;
Overcooked separates behavioral readout from cooperation outcomes; and FHN
varies the number of independent histories for field prediction. The suite map
in Appendix~\ref{app:suite} links each application to its implementation.
The corresponding Formation, Use and Value analyses appear in
Appendices~\ref{app:formation-complete}, \ref{app:information-use}
and~\ref{app:conditional-value}.

\FloatBarrier
\subsection{Learner and scene transfer: CoPhy}\label{app:cophy}\label{app:cophy-new5}
The CoPhy comparisons show both gains and adverse effects across learner
families and scenes. We distinguish the source-0 development matrix,
additional-source replication, and sensitivity to the fitted reader.
The three-seed held-out Collision comparison is reported in
Appendix~\ref{app:cophy-collision-test}.

\begin{figure}[!htbp]
\centering
\includegraphics[width=\linewidth]{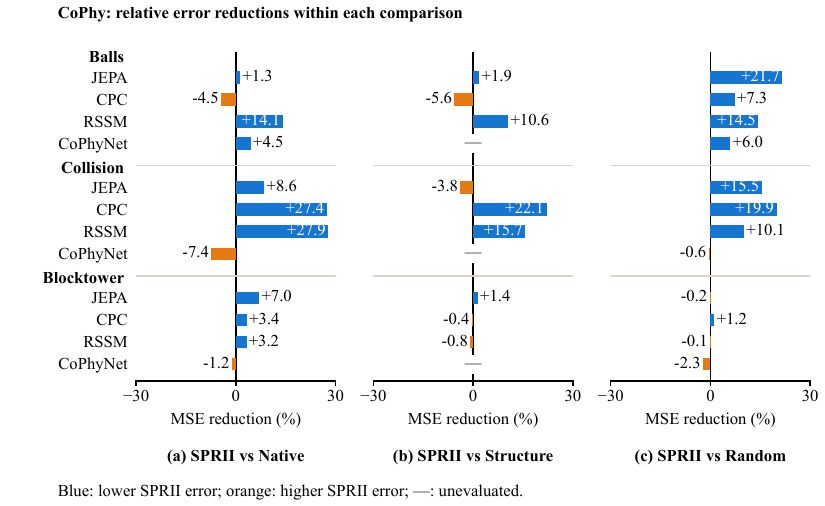}
\caption{\textbf{CoPhy gains depend on learner, scene and comparator.}
(a)--(c) Percentage MSE reduction against Native, Structure and Random:
$100(\mathrm{control}-\mathrm{SPRII})/\mathrm{control}$; positive favors SPRII.
These are single-source development comparisons. CoPhyNet Structure was
not evaluated in this matrix; held-out Collision results appear separately
in Table~\ref{tab:v3-collision}.}
\label{fig:v3-cophy-breadth}\label{fig:cophy-interface}
\end{figure}

\paragraph{Breadth across learners and scenes.}
The source-0 matrix contains 45 evaluated learner--scene--recipe cells across
JEPA, CPC, RSSM and supervised CoPhyNet in Balls, Collision and Blocktower.
Figure~\ref{fig:v3-cophy-breadth} gives relative error reductions against each
available control, and Table~\ref{tab:cophy-complete} gives absolute errors.
CoPhyNet has Native and Random controls; its Structure arm was not evaluated.

\label{app:records-cophy}

\begin{table}[H]
\centering\small
\setlength{\tabcolsep}{4pt}
\caption{\textbf{CoPhy prediction error by learner and scene.}
Source-0 development MSE under S3/query3 and source100/head100.
Dashes mark unevaluated Structure arms; bold marks the lowest MSE per row.}
\label{tab:cophy-complete}
\begin{tabular*}{\linewidth}{@{\hspace{3.5pt}\extracolsep{\fill}}llrr>{\columncolor{KeyRowTint}}rr@{\hspace{3.5pt}}}
\toprule
\textbf{Scene} & \textbf{Family} & \textbf{Native} & \textbf{Structure} & \textbf{SPRII} & \textbf{Random}\\
\tableheadrule
\multicolumn{6}{@{}l}{\textbf{Cross: self-supervised dynamics objectives}}\\[3pt]
Balls & JEPA & 1.2563 & 1.2630 & \textbf{1.2396} & 1.5827 \\
 & CPC & 1.4399 & \textbf{1.4239} & 1.5042 & 1.6228 \\
 & RSSM & 1.3105 & 1.2582 & \textbf{1.1254} & 1.3155 \\
\addlinespace[2pt]
Collision & JEPA & 0.2512 & \textbf{0.2212} & 0.2297 & 0.2717 \\
 & CPC & 0.2448 & 0.2283 & \textbf{0.1778} & 0.2220 \\
 & RSSM & 0.2731 & 0.2338 & \textbf{0.1970} & 0.2192 \\
\addlinespace[2pt]
Blocktower & JEPA & 0.0931 & 0.0878 & 0.0865 & \textbf{0.0863} \\
 & CPC & 0.1030 & \textbf{0.0991} & 0.0995 & 0.1007 \\
 & RSSM & 0.0974 & \textbf{0.0935} & 0.0943 & 0.0942 \\
\midrule
\multicolumn{6}{@{}l}{\textbf{Align + Cross: supervised CoPhyNet objective}}\\[3pt]
Balls & CoPhyNet & 1.0534 & -- & \textbf{1.0056} & 1.0698 \\
Collision & CoPhyNet & \textbf{0.2033} & -- & 0.2184 & 0.2171 \\
Blocktower & CoPhyNet & 0.0871 & -- & 0.0882 & \textbf{0.0862} \\
\bottomrule
\end{tabular*}
\end{table}

\paragraph{Source information and consumed memory.}
Source $U$, independent support $S$ and trained reader memory $M$ are
separate measurements (Appendices~\ref{app:method-interfaces} and~\ref{app:reader-interfaces}). The complete
source-$U$ grid covers all 45 source roles at source100, seed 0; it uses
train-fitted ridge probes before the supervised reader projection.
In Balls, SPRII restitution $R^2_U$ is 0.7875/0.7672/0.8133/0.6170 for
JEPA/CPC/RSSM/CoPhyNet, while mass/friction accessibility stays weak.
Accessibility and task error dissociate here as well: CPC Balls has
$R^2_U=0.7672$ versus Native 0.5959 but MSE 1.5042 versus 1.4399.
Collision mass access is lower for SPRII than Native in JEPA
(0.3683 versus 0.3799) and CoPhyNet (0.1757 versus 0.2004).

\FloatBarrier
\paragraph{Replication across sources and in Balls.}
Additional sources preserve the comparison grid: JEPA Blocktower has higher
error than Random at both sources, while JEPA Collision changes ordering
against Structure. Appendix~\ref{app:replication-status} specifies source coverage and development selection.

\begin{table}[!htbp]\centering\small
\setlength{\tabcolsep}{2.2pt}\renewcommand{\arraystretch}{1.04}
\caption{\textbf{CoPhy prediction and accessibility across source seeds.}
Development results include reader-selection examples. $S$ denotes independent
support and $M$ fitted reader memory. Probe targets are restitution (Balls),
mass (Collision) and vertical gravity (Blocktower). Dashes mark unevaluated cells.}
\label{tab:ctp-replication-complete}\label{tab:rssm-collision-replication}
\begin{tabular}{@{\hspace{3.5pt}}lrrrrrrrrr@{\hspace{3.5pt}}}\toprule
 & \multicolumn{3}{c}{\textbf{Prediction MSE}} & \multicolumn{3}{c}{\textbf{Support $R^2_S$}} & \multicolumn{3}{c}{\textbf{Reader memory $R^2_M$}}\\
\cmidrule(lr){2-4}\cmidrule(lr){5-7}\cmidrule(lr){8-10}
\textbf{Arm} & \textbf{Src 0} & \textbf{Src 1} & \textbf{Src 2} & \textbf{Src 0} & \textbf{Src 1} & \textbf{Src 2} & \textbf{Src 0} & \textbf{Src 1} & \textbf{Src 2}\\\tableheadrule
\multicolumn{10}{@{}l}{\textbf{JEPA / Balls}}\\
Native & 1.256 & 1.319 & --- & 0.7977 & 0.7981 & --- & 0.7865 & 0.7778 & ---\\
Structure & 1.263 & 1.294 & --- & 0.7855 & 0.7866 & --- & 0.7832 & 0.7788 & ---\\
Cross & 1.240 & 1.305 & --- & 0.9162 & 0.9185 & --- & 0.9051 & 0.9116 & ---\\
Random & 1.583 & 1.644 & --- & 0.6777 & 0.6918 & --- & 0.6884 & 0.6823 & ---\\
\addlinespace[3pt]
\multicolumn{10}{@{}l}{\textbf{JEPA / Collision}}\\
Native & 0.2512 & 0.2444 & --- & 0.4431 & 0.4366 & --- & 0.5657 & 0.5619 & ---\\
Structure & 0.2212 & 0.2238 & --- & 0.5976 & 0.6246 & --- & 0.6659 & 0.6706 & ---\\
Cross & 0.2297 & 0.2154 & --- & 0.6422 & 0.6718 & --- & 0.6451 & 0.6713 & ---\\
Random & 0.2717 & 0.2645 & --- & 0.5005 & 0.5250 & --- & 0.5286 & 0.5502 & ---\\
\addlinespace[3pt]
\multicolumn{10}{@{}l}{\textbf{JEPA / Blocktower}}\\
Native & 0.0931 & 0.0895 & --- & 0.1028 & 0.0902 & --- & 0.0430 & 0.0467 & ---\\
Structure & 0.0878 & 0.0876 & --- & 0.1252 & 0.1301 & --- & 0.0714 & 0.0879 & ---\\
Cross & 0.0865 & 0.0876 & --- & 0.1761 & 0.1564 & --- & 0.1242 & 0.0860 & ---\\
Random & 0.0863 & 0.0853 & --- & 0.1673 & 0.1678 & --- & 0.1256 & 0.1154 & ---\\
\addlinespace[3pt]
\multicolumn{10}{@{}l}{\textbf{CPC / Collision}}\\
Native & 0.2448 & 0.2517 & --- & 0.4372 & 0.4213 & --- & 0.5604 & 0.5484 & ---\\
Structure & 0.2283 & 0.2323 & --- & 0.6126 & 0.5929 & --- & 0.6476 & 0.6367 & ---\\
Cross & 0.1778 & 0.1841 & --- & 0.7495 & 0.7411 & --- & 0.7541 & 0.7419 & ---\\
Random & 0.2220 & 0.2212 & --- & 0.6657 & 0.6588 & --- & 0.6717 & 0.6711 & ---\\
\addlinespace[3pt]
\multicolumn{10}{@{}l}{\textbf{RSSM / Collision}}\\
Native & 0.2731 & 0.2674 & 0.2729 & 0.334 & 0.240 & 0.344 & 0.497 & 0.494 & 0.477\\
Structure & 0.2338 & 0.2332 & 0.2273 & 0.564 & 0.565 & 0.570 & 0.627 & 0.642 & 0.634\\
Cross & 0.1970 & 0.1997 & 0.2112 & 0.738 & 0.723 & 0.733 & 0.736 & 0.719 & 0.724\\
Random & 0.2192 & 0.2199 & 0.2284 & 0.633 & 0.643 & 0.634 & 0.678 & 0.669 & 0.664\\
\bottomrule\end{tabular}
\par\smallskip\textbf{RSSM Collision: three-source summary}\par\smallskip
\begin{tabular}{@{\hspace{3.5pt}}lrrr@{\hspace{3.5pt}}}\toprule
\textbf{Arm} & \textbf{MSE: mean $\pm$ SD} & \textbf{Support $R^2_S$} & \textbf{Reader memory $R^2_M$}\\\tableheadrule
Native & \msd{0.2711}{0.0032} & 0.306 & 0.490\\
Structure & \msd{0.2314}{0.0036} & 0.566 & 0.635\\
Cross & \msd{0.2026}{0.0075} & 0.731 & 0.727\\
Random & \msd{0.2225}{0.0051} & 0.636 & 0.671\\\bottomrule
\end{tabular}\end{table}

The three-source Balls extension uses 2,000 development recipients with
exact pairing. Its relation-trained arm improves Matched error against
Native/Random/Structure by 8.95\%/13.68\%/9.13\% for RSSM, and against
Native/Random by 3.91\%/3.19\% for CoPhyNet. All five recipient-bootstrap
family-adjusted intervals are positive. Table~\ref{tab:cophy-new5-summary}
reports Null, Matched and Wrong inputs and all five contrasts; these intervals
condition on the fitted models and exclude selection uncertainty.
Actual-memory use is analyzed in Appendix~\ref{app:balls-use}; the source-$U$
probes above measure a different channel.

\begin{table}[!htbp]\centering\small
\setlength{\tabcolsep}{4pt}
\caption{\textbf{All five Balls contrasts favor relation training.}
Three-source means on 2,000 development recipients. Recipient-bootstrap
95\% intervals are Bonferroni-adjusted over five contrasts, conditional on
fitted models and excluding selection uncertainty.}
\label{tab:cophy-new5-summary}\label{tab:cophy-three-seed-contrasts}
\begin{tabular}{@{\hspace{3.5pt}}llrrr@{\hspace{3.5pt}}}\toprule
\multicolumn{5}{l}{\textbf{A. Absolute risks under recorded context conditions}}\\
\textbf{Family} & \textbf{Arm} & \textbf{Matched} & \textbf{Null} & \textbf{Wrong}\\\tableheadrule
RSSM & SPRII (C) & 1.171 & 1.892 & 2.296\\
 & Native & 1.286 & 2.485 & 2.293\\
 & Random & 1.356 & 2.004 & 2.300\\
 & Structure & 1.288 & 1.937 & 2.335\\
CoPhyNet & SPRII (A+C) & 1.044 & 1.791 & 2.228\\
 & Native & 1.087 & 1.773 & 2.244\\
 & Random & 1.079 & 1.759 & 2.216\\\midrule
\multicolumn{5}{l}{\textbf{B. Comparator minus SPRII, paired before source averaging}}\\
\textbf{Family} & \textbf{Comparator} & \textbf{Difference} & \multicolumn{2}{c}{\textbf{Adjusted 95\% interval}}\\\tableheadrule
RSSM & Native & 0.1151 & \multicolumn{2}{c}{$[0.0902, 0.1395]$}\\
 & Random & 0.1856 & \multicolumn{2}{c}{$[0.1607, 0.2096]$}\\
 & Structure & 0.1177 & \multicolumn{2}{c}{$[0.0927, 0.1433]$}\\
CoPhyNet & Native & 0.0424 & \multicolumn{2}{c}{$[0.0276, 0.0574]$}\\
 & Random & 0.0344 & \multicolumn{2}{c}{$[0.0193, 0.0497]$}\\\bottomrule
\end{tabular}
\end{table}

\paragraph{Sensitivity to the fitted reader.}\label{app:cophy-reader-design}
With frozen sources, the CoPhyNet reader recipe changes comparison sign and
magnitude across scenes. The source-50 factorial in
Table~\ref{tab:v3-cophy-reader} varies initialization, dropout and delivery;
the source-100 endpoint below uses a separate budget.

\begin{table}[!htbp]
\centering\small
\caption{\textbf{CoPhyNet gains depend on the reader recipe.}
SPRII versus Native error reduction (\%) for one frozen source on
validation data (source-50/head-100).}
\label{tab:v3-cophy-reader}\label{tab:cophy-reader}
\begin{tabular}{lrrr}
\toprule
\textbf{Reader recipe} & \textbf{Balls} & \textbf{Collision} & \textbf{Blocktower}\\
\tableheadrule
Main initialization, dropout 0.1, initial & $-5.448$ & $-4.089$ & $-1.224$ \\
Alternative initialization, dropout 0 & $+7.506$ & $-2.104$ & $-1.477$ \\
Alternative initialization, dropout 0.1 & $+6.074$ & $-4.426$ & $+2.090$ \\
New initialization, dropout 0 & $+0.617$ & $-7.591$ & $+0.621$ \\
New initialization, dropout 0.1, every step & $+6.646$ & $-2.925$ & $+1.134$ \\
\bottomrule
\end{tabular}
\end{table}

At source-100, the every-step reader gives Native/SPRII/Random MSE
1.037115/0.956198/1.124851 on Balls and 0.208457/0.222786/0.218758 on Collision.
These use the source-100 budget.

\paragraph{Operation and configuration comparisons.}
Table~\ref{tab:followup-complete-summary} reports the CoPhy configuration
comparisons with their evaluation populations and fitted-model coverage.
The single-source development comparisons complement the replicated results;
their paired intervals condition on the fitted models and evaluation units.
\begin{table}[!htbp]\centering\small
\caption{\textbf{CoPhy operation and configuration comparisons.}
Errors are lower-is-better MSE. Development rows use one fitted source/reader
seed; paired intervals condition on the reported models and evaluation units.}
\label{tab:followup-complete-summary}
\setlength{\tabcolsep}{3pt}\renewcommand{\arraystretch}{1.14}
\begin{tabularx}{\linewidth}{@{\hspace{3.5pt}}>{\raggedright\arraybackslash}p{0.17\linewidth}>{\raggedright\arraybackslash}p{0.19\linewidth}>{\raggedright\arraybackslash}p{0.33\linewidth}>{\raggedright\arraybackslash}X@{\hspace{3.5pt}}}\toprule
\textbf{Configuration} & \textbf{Coverage} & \textbf{Numerical comparison} & \textbf{Finding / scope}\\\tableheadrule
JEPA Collision: J2 & 1,994 test episodes; 3 seeds per arm &
J2 \msd{0.184}{0.005};\newline Structure \msd{0.215}{0.007} &
14.2\% lower than Structure (all arms source-50); all four arms in Table~\ref{tab:jepa-collision-followup}.\\\tableentryrule
CoPhyNet Collision: Cross & 4,000 development recipients; seed 0 &
Cross 0.2032;\newline SPRII S1 reference 0.2107 &
Near Native; paired interval against Native includes zero.\\\tableentryrule
CPC Balls: Align & 2,000 development recipients; seed 0 &
Align 1.393;\newline SPRII C2 reference 1.415;\newline Structure 1.388 &
Lower than the SPRII C2 reference, higher than Structure.\\\tableentryrule
CoPhyNet Blocktower & 8,088 development recipients; seed 0 &
SPRII 0.088180;\newline Native 0.087114;\newline later, separately evaluated Structure 0.087114;\newline Random 0.086230.\newline SPRII$-$Random: 0.001951 $[0.000668, 0.003184]$ &
Paired interval versus the later Structure control includes zero; Random has lower mean error.\\\bottomrule
\end{tabularx}
\begin{papertablenotes}
Each row uses its stated population and fitted-model coverage.
\end{papertablenotes}\end{table}

\FloatBarrier

\subsection{Recurring-partner behavior and cooperation: Overcooked}\label{app:overcooked}\label{app:overcooked-external}
Relative to H+VC, relation alignment lowers behavioral-probe error across
the three history budgets. At full history, familiar-partner returns also improve in each of
three training seeds. All comparisons use one layout and fixed 20,000-update
endpoints; Appendix~\ref{app:overcooked-profile} gives the shared evaluation protocol.

\begin{table}[!htbp]\centering\small
\caption{\textbf{Full-history alignment improves familiar-partner returns in all three seeds.}
Paired Align $-$ H+VC differences, mean $\pm$ sample SD across three training
seeds. Lower probe MSE and higher return are better. Familiar and held-out
development partners are reported separately.}
\label{tab:v3-overcooked}\label{tab:overcooked-intervals}
\setlength{\tabcolsep}{5pt}
\begin{tabular}{@{\hspace{3.5pt}}lrrr@{\hspace{3.5pt}}}\toprule
\textbf{History} & \textbf{Familiar probe MSE} & \textbf{Familiar return} & \textbf{Held-out dev. return}\\\tableheadrule
25\% & \msd{-0.00567}{0.00810} & \msd{1.489}{2.335} & \msd{-2.222}{15.21}\\
50\% & \msd{-0.01344}{0.00728} & \msd{0.389}{1.110} & \msd{18.44}{20.66}\\
\rowcolor{KeyRowTint}
\textbf{100\%} & \msd{\mathbf{-0.02120}}{0.00610} & \msd{\mathbf{1.889}}{2.175} & \msd{\mathbf{-1.111}}{11.44}\\
\bottomrule\end{tabular}
\end{table}

For full history, the familiar return differences are $+4.400$, $+0.667$ and
$+0.600$ across seeds, whereas held-out development differences are
$+9.333$, $+0.667$ and $-13.333$. The latter population therefore shows mixed
return effects. At 25\% and 50\% history, seeds also differ in familiar-return signs.

\label{app:records-behavior}
Table~\ref{tab:overcooked-complete} compares behavioral readout, representation
distances and cooperation returns at the same checkpoints. At full history, Align lowers
probe MSE relative to Random in all three seeds (paired difference
$\msd{-0.02583}{0.00342}$), and the Align--H+VC probe difference is
$\msd{-0.02120}{0.00610}$. The Align mean, 0.05395, remains above the
constant-mean reference, 0.04333; held-out development probe means also
remain above their reference, 0.07062.

\begin{table}[!htbp]\centering\small
\caption{\textbf{Overcooked behavioral readout and cooperation returns.}
Mean $\pm$ sample SD across three training seeds. A denotes Align; VC denotes
H+VC. The distance ratio is different-/same-partner distance on the familiar
behavioral-signature bank, computed per seed before averaging.
Random has no return evaluation.}
\label{tab:overcooked-complete}
\setlength{\tabcolsep}{5pt}
\textbf{A. Behavioral readout and representation}\par\smallskip
\begin{tabular}{@{\hspace{3.5pt}}llrrr@{\hspace{3.5pt}}}\toprule
\textbf{History} & \textbf{Recipe} & \textbf{Familiar MSE $\downarrow$} & \textbf{Held-out dev. MSE $\downarrow$} & \textbf{Distance ratio}\\\tableheadrule
25\% & VC & \msd{0.06788}{0.00667} & \msd{0.13880}{0.04230} & \msd{1.343}{0.061}\\
25\% & A & \msd{0.06220}{0.00974} & \msd{0.09733}{0.01384} & \msd{1.917}{0.071}\\
\addlinespace[2pt]
50\% & VC & \msd{0.06934}{0.00811} & \msd{0.14090}{0.02324} & \msd{1.408}{0.019}\\
50\% & A & \msd{0.05590}{0.00736} & \msd{0.08966}{0.01240} & \msd{2.329}{0.326}\\
\addlinespace[2pt]
100\% & VC & \msd{0.07516}{0.00791} & \msd{0.12400}{0.02857} & \msd{1.458}{0.036}\\
100\% & A & \msd{0.05395}{0.01106} & \msd{0.10630}{0.01046} & \msd{2.269}{0.206}\\
\midrule
100\% & Random & \msd{0.07978}{0.01423} & \msd{0.14840}{0.03203} & \msd{1.460}{0.066}\\
\bottomrule\end{tabular}
\par\smallskip\textbf{B. Cooperation returns}\par\smallskip
\begin{tabular}{@{\hspace{3.5pt}}llrr@{\hspace{3.5pt}}}\toprule
\textbf{History} & \textbf{Recipe} & \textbf{Familiar partners} & \textbf{Held-out dev. partners}\\\tableheadrule
25\% & VC & \msd{15.78}{1.262} & \msd{-1.333}{4.372}\\
25\% & A & \msd{17.27}{1.768} & \msd{-3.556}{14.850}\\
\addlinespace[2pt]
50\% & VC & \msd{15.20}{0.200} & \msd{-19.56}{2.341}\\
50\% & A & \msd{15.59}{0.916} & \msd{-1.111}{18.410}\\
\addlinespace[2pt]
100\% & VC & \msd{14.04}{1.500} & \msd{-19.11}{8.572}\\
100\% & A & \msd{15.93}{0.742} & \msd{-20.22}{3.672}\\
\bottomrule\end{tabular}\end{table}

\FloatBarrier
\subsection{History-count scaling in reaction-diffusion fields: FHN}
\label{app:external-fhn}
Additional independent histories reduce h50 field-prediction error in both
methods. SPRII has larger relative reductions on all three splits, while its
mean error remains higher than NOD at matched $K=1$ and $K=4$. From $K=1$ to $K=4$, SPRII error decreases by 16.9\%, 18.0\% and 10.9\%
on ID, OOD-Intra and OOD-Extra, compared with 8.4\%, 9.2\% and 7.1\% for NOD.
The released NOD-Hier architecture runs on the audited Python-generated bank
(Appendix~\ref{app:nod-fhn-profile}). Initial condition 5 selects configurations;
reporting uses 15, 24 and 45, with three fixed sources 42/43/44. Source 42
is the development seed; 43/44 repeat the selected recipe.
NOD uses 45,000 updates and SPRII 50,000; these history-count comparisons
therefore do not isolate alignment at equal training compute. The training
schedules are specified in Appendix~\ref{app:nod-fhn-profile}.
\begin{figure}[!htbp]\centering
\includegraphics[width=\linewidth]{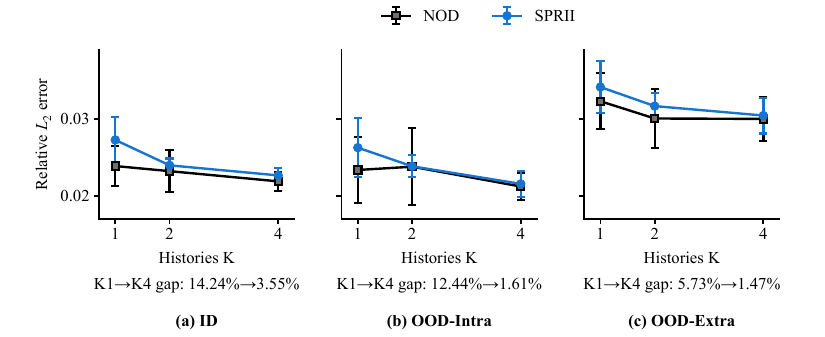}
\caption{\textbf{Additional histories narrow the FHN endpoint gap on all three splits.}
H50 relative-$L_2$ error: (a) ID, (b) OOD-Intra, (c) OOD-Extra.
Points and whiskers show mean $\pm$ sample SD across three sources.
Annotations give $100(\mathrm{SPRII}-\mathrm{NOD})/\mathrm{NOD}$ at $K=1,4$.}
\label{fig:visual-fhn-history}
\end{figure}

\begin{table}[H]\centering\small\setlength{\tabcolsep}{4pt}
\caption{\textbf{Additional histories narrow the FHN prediction gap.}
H50 relative-$L_2$ error averaged across three sources; lower is better.
Both methods receive the same $K$ independent histories, with different
training budgets as described above.}\label{tab:fhn-breadth}
\begin{tabular}{@{\hspace{3.5pt}}lrrrrrr@{\hspace{3.5pt}}}\toprule
 & \multicolumn{2}{c}{\textbf{$K=1\to4$ reduction}} & \multicolumn{2}{c}{\textbf{NOD error}} & \multicolumn{2}{c}{\textbf{SPRII error}}\\
\cmidrule(lr){2-3}\cmidrule(lr){4-5}\cmidrule(lr){6-7}
\textbf{Split} & \textbf{SPRII} & \textbf{NOD} & \textbf{$K=1$} & \textbf{$K=4$} & \textbf{$K=1$} & \textbf{$K=4$}\\\tableheadrule
ID & 16.9\% & 8.4\% & 0.02388 & 0.02188 & 0.02728 & 0.02266\\
OOD-Intra & 18.0\% & 9.2\% & 0.02336 & 0.02121 & 0.02627 & 0.02155\\
OOD-Extra & 10.9\% & 7.1\% & 0.03233 & 0.03002 & 0.03418 & 0.03046\\
\bottomrule\end{tabular}\end{table}

At $K=1$, SPRII's H50 mean error is 14.24\%, 12.44\% and 5.73\% higher
on ID, OOD-Intra and OOD-Extra. At the matched $K=4$ observation budget,
these gaps shrink to 3.55\%, 1.61\% and 1.47\%. Factor accessibility varies
by factor, split and source, as shown in the probe table below.

\paragraph{FHN factor accessibility.}
\label{app:records-fhn}
The bank contains nine training, four OOD-Intra and twelve OOD-Extra systems;
$K=1$ uses initial condition 36. All six models share 13,500 report-case keys.

\begin{table}[!htbp]\centering\small
\caption{\textbf{FHN accessibility varies by factor, split and source.}
Training-fitted probe $R^2$ with one history ($K=1$).}
\label{tab:external-fhn-probes}
\begin{minipage}[t]{0.48\linewidth}\centering
\textbf{A. $k$}\par\smallskip
\setlength{\tabcolsep}{3pt}
\begin{tabular}{@{\hspace{3.5pt}}llrrr@{\hspace{3.5pt}}}\toprule
\textbf{Split} & \textbf{Method} & \textbf{Src 42} & \textbf{Src 43} & \textbf{Src 44}\\\tableheadrule
ID & NOD & 0.9698 & 0.9301 & 0.9836\\
ID & SPRII & 0.9444 & 0.3775 & 0.8952\\
Intra & NOD & 0.9269 & 0.8098 & 0.9377\\
Intra & SPRII & 0.8602 & 0.5513 & 0.6923\\
Extra & NOD & 0.9351 & 0.9139 & 0.9284\\
Extra & SPRII & 0.8587 & 0.0949 & 0.8730\\
\bottomrule\end{tabular}\end{minipage}
\hfill
\begin{minipage}[t]{0.48\linewidth}\centering
\textbf{B. $\beta$}\par\smallskip
\setlength{\tabcolsep}{3pt}
\begin{tabular}{@{\hspace{3.5pt}}llrrr@{\hspace{3.5pt}}}\toprule
\textbf{Split} & \textbf{Method} & \textbf{Src 42} & \textbf{Src 43} & \textbf{Src 44}\\\tableheadrule
ID & NOD & 0.9627 & 0.9496 & 0.9394\\
ID & SPRII & 0.9766 & 0.9449 & 0.9097\\
Intra & NOD & 0.9877 & 0.9829 & 0.9765\\
Intra & SPRII & 0.9600 & 0.9269 & 0.8990\\
Extra & NOD & 0.9521 & 0.9409 & 0.9299\\
Extra & SPRII & 0.9848 & 0.9518 & 0.9154\\
\bottomrule\end{tabular}\end{minipage}
\end{table}

The additional $K=2,4$ probes use their four-recipient population.

\FloatBarrier
\paragraph{Additional-budget configuration comparison.}
A separate source-42 configuration is compared with extra-50k NOD on report
initial conditions 15/24/45, excluding selection initial condition 5.
At ID $K=4,H=50$, SPRII has lower mean error, but the paired-system interval
includes zero: relative $L_2$ error is 0.02013 for SPRII and 0.02152 for
extra-50k NOD. The interval conditions on the reported fitted models and
evaluation systems; the other eight split/horizon cells have higher SPRII error.

\FloatBarrier
\section{Theoretical Connections}
\label{sec:analysis}
\label{app:theory}

We analyze four reference models connecting relations to persistent-context
learning. Pairing determines which variations agreement penalizes; predictive
relevance selects among stable directions; fixed-route interventions measure
context dependence; and conditional risk characterizes the value available
to a downstream reader. Each proposition states its required assumptions.

\label{app:candidate-structure}
\label{app:relation-structure}

\FloatBarrier
\subsection{Setup and lifecycle}
\label{app:theory-setup}
\paragraph{Setup.}
Let $G$ be the grouping variable specified by a relation, $H$ an interaction
history, and $Z=\phi(H)\in\mathbb R^d$ a fixed representation with finite
second moments. A group may denote a whole system, selected attributes, a
partner, or a recorded task. Write $\mu_G=\mathbb E[Z\mid G]$,
$\Sigma_{\rm between}(G)=\operatorname{Cov}(\mu_G)$, and
$\Sigma_{\rm within}(G)=\mathbb E[\operatorname{Cov}(Z\mid G)]$.
The sampling law is $\Pi_G(A,B)=P(B)Q_G(A\mid B)$, as in
Eq.~(\plaineqref{eq:pairing-kernel}). Conditional independence is imposed
explicitly for the iid results below; the general pair law permits dependent
interaction histories.

\paragraph{Remark: available information, reliance and realized value.}
The lifecycle distinguishes information present in a context, reliance by a
specified predictor, and task value realized by that predictor. These are
operationally assessed with geometry and probes, fixed-predictor substitutions,
and task readers, respectively; failure of a particular probe is not evidence
that the information is absent. For a fixed joint law of recipient inputs $X$,
donor code $D$ and target $Y$, conditional independence of $Y$ and $D$ given $X$
implies zero \emph{Bayes-optimal incremental value}
$\Delta_\ell^\star$ in Proposition 4. This statement concerns additional target
information beyond $X$, not the difference between arbitrary finite fitted
readers: their excess-risk terms can differ even when $\Delta_\ell^\star=0$.
Conversely, accessible information need not be relevant beyond $X$, and positive
available value need not be realized by a fitted reader. Donor sensitivity
establishes reliance under the specified substitution, while the residual
cross term in Appendix~\ref{app:specificity-theory} determines whether that
reliance reduces risk. A newly fitted reader's benefit concerns its own use of
context; it does not establish use by the original predictor.

\subsection{Formation: pairing, reliability and predictive selection}
\label{app:formation-theory}
\paragraph{Pairing structure.}
\paragraph{Proposition 1 (structure exposed by pairing).}
For a pair law with finite second moments, means $\mu_A,\mu_B$, marginal covariances
$\Sigma_A,\Sigma_B$, and cross covariance $\Sigma_{AB}$,
\begin{equation}
\mathbb E_{\Pi_G}\|Z^A-Z^B\|_2^2
=\operatorname{tr}\Sigma_A+\operatorname{tr}\Sigma_B
-2\operatorname{tr}\Sigma_{AB}+\|\mu_A-\mu_B\|_2^2.
\label{eq:app-general-pair}
\end{equation}
If a matched pair is instead specified by drawing $G$ from its population
law and then drawing $Z^A,Z^B$ conditionally iid from $P(Z\mid G)$, then
\begin{equation}
\operatorname{Cov}(Z^A,Z^B)=\Sigma_{\rm between}(G),\qquad
\mathbb E\|Z^A-Z^B\|_2^2=2\operatorname{tr}\Sigma_{\rm within}(G).
\label{eq:conditional-pair}
\end{equation}

\paragraph{Proof.}
Expand the squared difference and use
$\mathbb E\|Z^A\|^2=\operatorname{tr}\Sigma_A+\|\mu_A\|^2$
and $\mathbb E[(Z^A)^\top Z^B]=\operatorname{tr}\Sigma_{AB}+\mu_A^\top\mu_B$.
This proves Eq.~(\plaineqref{eq:app-general-pair}). For the ideal matched
pair, write $Z^i=\mu_G+\epsilon_i$. Conditional centering gives
$\mathbb E[\epsilon_i\mid G]=0$, and conditional independence gives
$\mathbb E[\epsilon_A\epsilon_B^\top\mid G]=0$. Thus the cross covariance
is $\operatorname{Cov}(\mu_G)$, while the covariance of
$\epsilon_A-\epsilon_B$ given $G$ is $2\operatorname{Cov}(Z\mid G)$.
Taking its trace and averaging proves Eq.~(\plaineqref{eq:conditional-pair}).
The same centering argument also yields the total-covariance identity
$\operatorname{Cov}(Z)=\Sigma_{\rm between}(G)+\Sigma_{\rm within}(G)$.
\unskip\nobreak\hfill\mbox{$\square$}

\paragraph{What changes under an actual sampler?}
Conditional centering gives, without the iid premise,
\begin{equation*}
\Sigma_{AB}
=\operatorname{Cov}(\mathbb E[Z^A\mid G],\mathbb E[Z^B\mid G])
+\mathbb E[\operatorname{Cov}(Z^A,Z^B\mid G)].
\end{equation*}
The two mixed mean--residual terms vanish by iterated expectation. Overlap,
shared episode dynamics, balancing, and forced mismatches can leave the
second term nonzero or alter the conditional means. The actual pairing law
therefore determines which dependence alignment preserves. Under ideal
pairing, alignment penalizes within-group differences and leaves group-mean
variation as a candidate for retention. A constant code also has zero
alignment loss; predictive and non-collapse terms provide the additional
constraints needed to learn informative codes.

\FloatBarrier
\paragraph{Reliability and independent repetitions.}
\label{app:reliability-refinement}
\label{app:pairing-theory}

Reliability changes whether shared variation is exempt from the disagreement
penalty. Let $Q_c$ select correct pairs and $Q_r$ random pairs with the same
donor and recipient marginals. For $\alpha\in[0,1]$, the mixture
$Q_\alpha=\alpha Q_c+(1-\alpha)Q_r$ implies
$\mathbb E_{Q_\alpha}\ell=\alpha\mathbb E_{Q_c}\ell+(1-\alpha)\mathbb E_{Q_r}\ell$
for any integrable pair loss at fixed parameters.

\paragraph{Proposition 2 (alignment pressure under corrupted relations).}
Let $H=L\theta+N$, where $\theta$ has covariance $\Sigma_\theta$
and independent zero-mean noise $N$ has covariance $\Sigma_N$.
Correct pairs share $\theta$ and independently resample $N$; random
pairs independently resample both. For a fixed encoder $Z=WH$,
\begin{equation}
\mathbb E_{Q_\alpha}\|W(H^A-H^B)\|_2^2
=2\operatorname{tr}(W\Sigma_NW^\top)
+2(1-\alpha)\operatorname{tr}(WL\Sigma_\theta L^\top W^\top).
\label{eq:fidelity-pressure}
\end{equation}

\paragraph{Proof.}
Under $Q_c$, $H^A-H^B=N^A-N^B$ has zero mean and covariance $2\Sigma_N$.
Under $Q_r$, independence gives covariance
$2L\Sigma_\theta L^\top+2\Sigma_N$ for the difference. In each case,
the expected squared norm after applying $W$ is the trace of the transformed
covariance. Mixing these two expressions proves the result.
\unskip\nobreak\hfill\mbox{$\square$}

Correct pairing leaves persistent variation outside this penalty; random
pairing contracts it. The coefficient $1-\alpha$ describes the objective in this
reference model; Figure~\ref{fig:relation-mechanism}(a) reports the learned response.
If only alignment assignments change, per-view variance/covariance
regularizers remain common. If Cross consumes the corrupted assignment,
its task loss changes too. The calculation therefore isolates alignment
pressure; the monotone geometry in Figure~\ref{fig:relation-mechanism}(a) is
an empirical response of separately trained nonlinear models.

\paragraph{Corollary (factor-specific alignment pressure).}
In Proposition 2, let $\theta=(m,\gamma,k)$ have independent components,
$L=[L_m,L_\gamma,L_k]$, and $\operatorname{Var}(\gamma)=\sigma_\gamma^2$.
Correct pairs share all three factors. In the corrupted component, retain
$m,k$ and independently redraw only $\gamma$ from its marginal; independently
redraw the observation noise in both components. For a fixed $W$,
\begin{equation}
\mathbb E\|W(H^A-H^B)\|^2
=2\operatorname{tr}(W\Sigma_NW^\top)
+2(1-\alpha)\sigma_\gamma^2
\operatorname{tr}(WL_\gamma L_\gamma^\top W^\top).
\label{eq:structured-fidelity-pressure}
\end{equation}
\paragraph{Proof.}
The correct-pair difference is $N^A-N^B$. In the corrupted component it is
$L_\gamma(\gamma^A-\gamma^B)+N^A-N^B$; independence and centering remove
cross terms and give second moment
$2\sigma_\gamma^2L_\gamma L_\gamma^\top+2\Sigma_N$.
Taking the trace after applying $W$ and mixing the two components yields
Eq.~(\plaineqref{eq:structured-fidelity-pressure}).
\unskip\nobreak\hfill\mbox{$\square$}

The additional penalty acts along $L_\gamma$; the shared $m,k$ variations
cancel from the pair difference. This fixed-encoder statement isolates the
pressure exerted by alignment. It does not imply that retraining a nonlinear
model leaves all other factors' geometry or probe accuracy unchanged:
predictive losses, capacity and variance constraints also select the learned
representation. Factor-selective degradation is therefore an empirical
hypothesis for Appendix~\ref{app:structured-mismatch}.
For a finite bank whose sampler \emph{forces} unequal drag values, independent
redrawing is not the sampling law. Replace $2\sigma_\gamma^2$ by the actual
$\mathbb E_Q[(\gamma^A-\gamma^B)^2]$; for uniform distinct draws from $n$ bank
entries, this is $2n\sigma_{\gamma,\mathrm{bank}}^2/(n-1)$ when bank variance
uses denominator $n$. More general balancing and unequal supports use the
actual pair second moment in Proposition 1. The qualitative cancellation of
exactly shared factors remains, while the coefficient must match the sampler.

\paragraph{Why separate realizations can add evidence.}
For conditionally iid codes $Z_1,\ldots,Z_R\mid G$ with mean $\mu_G$
and covariance $\Sigma(G)$, the average $\bar Z_R=R^{-1}\sum_r Z_r$
has conditional covariance $\Sigma(G)/R$:
independence removes the off-diagonal covariance terms in the sum.
Consequently,
$\mathbb E\|\bar Z_R-\mu_G\|^2=\operatorname{tr}\Sigma_{\rm within}(G)/R$.
For dependent repetitions, the remaining cross-covariance terms determine
the reduction. This estimation reference describes the information available
from repeated histories; \sprii{} learns its encoder from those histories.
The R2/R4/R8 bank comparison in Table~\ref{tab:pokeworld-geometry} tests the
learning effect of providing more interaction realizations.
\FloatBarrier
\paragraph{Semantic refinement and predictive selection.}

A relation specifies which variation may persist across a pair; the training
objective determines which eligible factor the representation emphasizes.
We first analyze a fixed code, then give a learning example in which the
dominant factor changes.

\paragraph{Semantic refinement for a fixed representation.}
Let $G_2$ refine $G_1$, so that $G_1$ is a measurable function of $G_2$.
For the same fixed $Z$ under the same population law, total covariance gives
\begin{equation*}
\Sigma_{\rm between}^{Z}(G_2)-\Sigma_{\rm between}^{Z}(G_1)
=\mathbb E[\operatorname{Cov}(\mathbb E[Z\mid G_2]\mid G_1)]\succeq0.
\end{equation*}
To see this, write $M_i=\mathbb E[Z\mid G_i]$. The tower property gives
$\mathbb E[M_2\mid G_1]=M_1$; total covariance applied to $M_2$ yields the
identity. Positive semidefiniteness follows because the last term's quadratic
form is $\mathbb E[\operatorname{Var}(v^\top M_2\mid G_1)]\geq0$ for every
$v$. Since the total covariance of $Z$ is unchanged, within-group covariance
decreases by the same matrix.

Thus $G_1=m$, $G_2=(m,\gamma)$, and $G_3=(m,\gamma,k)$ successively make
drag and stiffness eligible as group-stable variation. This identity compares
\emph{one representation under different groupings}. Experimental donor pools
may include forced mismatches and have non-nested supports; their effects
enter through the actual pair law.

\paragraph{Predictive selection under relation constraints.}
\label{app:predictive-selection}
\label{app:semantic-theory}
\label{app:learned-selection}
\label{app:spectral-theory}
The preceding identities keep the representation fixed while changing its
grouping. We now allow the representation to change and examine the joint
influence of predictive reward and relation disagreement in a joint spectral
selection problem.
\paragraph{A constrained predictive reference.}
Let $U$ collect the inputs supplied outside the history code, such as current
state, actions, and horizon. To model history information beyond $U$,
assume finite second moments and consider the residuals
$\widetilde H=H-\mathbb E[H\mid U]$ and
$\widetilde Y=Y-\mathbb E[Y\mid U]$, with a centered linear model
$\widetilde Y=K\widetilde H+\eta$ and
$\mathbb E[\eta\widetilde H^\top]=0$. Assume positive-definite history
covariance on the modeled subspace and whiten that subspace. Below, $H,Y$
denote these transformed variables, with $\operatorname{Cov}(H)=I_p$
and $Y=KH+\eta$ (absorbing the whitening into $K$).
Residualization and whitening define this analytical reference model.

\paragraph{Proposition 3 (joint predictive and relation selection).}
For $Z=WH$, $W\in\mathbb R^{d\times p}$ with $1\leq d<p$, impose
$WW^\top=I_d$ and let $\beta\geq0$. Consider
\begin{equation}
\mathcal J_G(W)
=\min_C\mathbb E\|Y-CWH\|_2^2
+\beta\mathbb E_{\Pi_G}\|W(H^A-H^B)\|_2^2.
\label{eq:app-selection-objective}
\end{equation}
Write $D_G=\mathbb E_{\Pi_G}[(H^A-H^B)(H^A-H^B)^\top]$ and
$M_G=K^\top K-\beta D_G$. The optimal decoder is $C^\star=KW^\top$,
and the optimal row spaces of $W$ are top-$d$ eigenspaces of $M_G$:
\begin{equation}
W_G^\star\in\arg\max_{WW^\top=I_d}\operatorname{tr}(WM_GW^\top).
\label{eq:spectral-selection}
\end{equation}
The selected subspace is unique if
$\lambda_d(M_G)>\lambda_{d+1}(M_G)$; its orthonormal basis need not be.

\paragraph{Proof.}
Since $\operatorname{Cov}(WH)=I_d$ and
$\operatorname{Cov}(Y,WH)=KW^\top$, completing the square gives
\begin{align*}
\mathbb E\|Y-CWH\|_2^2
&=\operatorname{tr}\operatorname{Cov}(Y)
-2\operatorname{tr}(CWK^\top)+\operatorname{tr}(CC^\top)\\
&=\operatorname{tr}\operatorname{Cov}(Y)
-\|KW^\top\|_F^2+\|C-KW^\top\|_F^2.
\end{align*}
Minimization sets $C=KW^\top$ and leaves predictive reward
$\operatorname{tr}(WK^\top KW^\top)$. The alignment term equals
$\operatorname{tr}(WD_GW^\top)$, including when the pair difference has a
nonzero mean. Removing the constant target variance therefore yields
Eq.~(\plaineqref{eq:spectral-selection}).

To characterize the maximum, diagonalize the symmetric matrix
$M_G=V\operatorname{diag}(\lambda_1,\ldots,\lambda_p)V^\top$ with decreasing
eigenvalues. Since $W^\top W$ is a rank-$d$ orthogonal projector,
$a_i=\|Wv_i\|_2^2$ satisfies $0\leq a_i\leq1$ and $\sum_i a_i=d$. Hence
\begin{equation*}
\operatorname{tr}(WM_GW^\top)=\sum_{i=1}^p\lambda_i a_i
\leq\sum_{i=1}^d\lambda_i.
\end{equation*}
Choosing the first $d$ eigenvectors as rows of $W$ attains the bound.
A strict eigengap requires projection onto precisely that subspace; ties
permit alternative choices within the tied eigenspace.
\unskip\nobreak\hfill\mbox{$\square$}

\paragraph{A realizable three-factor crossing.}
Let $H=(M,\Gamma,K_0)$ have independent standard-normal coordinates, use
$d=1$, $\beta=1$, and set $K^\top K=\operatorname{diag}(3,4,2)$.
Under $G_1=M$, pairs share mass and independently resample the other two
coordinates. Under $G_2=(M,\Gamma)$, they share mass and drag. Both laws
have the same standard-normal marginals. A shared coordinate cancels in the
pair difference; two independent standard normals have difference variance
$2$. Therefore
\begin{equation}
\begin{aligned}
D_{G_1}&=\operatorname{diag}(0,2,2),&
M_{G_1}&=\operatorname{diag}(3,2,0),\\
D_{G_2}&=\operatorname{diag}(0,0,2),&
M_{G_2}&=\operatorname{diag}(3,4,0).
\end{aligned}
\label{eq:app-crossing-example}
\end{equation}
Proposition 3 selects $W_{G_1}^\star=e_m^\top$ but
$W_{G_2}^\star=e_\gamma^\top$. Mass remains shared, yet removing drag's
penalty makes drag the leading direction. Broader semantic eligibility is
therefore compatible with a different, non-nested learned subspace.
This reproduces the qualitative factor crossover in
Figure~\ref{fig:relation-mechanism}(b) and Appendix~\ref{app:selectivity}: the
preferred direction changes while the original factor remains shared.

The argument also presupposes available evidence: for a fixed encoder, a
factor independent of its input remains independent of its code. The actual-input estimators in Table~\ref{tab:v3-poke-input-diversity}
therefore check for factor evidence in histories before asking which
factors the representation emphasizes.

\FloatBarrier
\subsection{Use: fixed-predictor reliance}
\label{app:specificity-theory}

A fixed-weight intervention holds the predictor and recipient inputs fixed
and changes supplied persistent context. It tests functional use
without refitting readers for each donor condition.

\paragraph{Risk change under a donor substitution.}
Fix a predictor $f$ and a joint evaluation law for $(X,Y,D_m,D_s)$,
where $D_m$ is matched and $D_s$ is the specified replacement. Assume
finite second moments for the target and both predictions. Write
$\hat Y_m=f(X,D_m)$, $\hat Y_s=f(X,D_s)$, and let
$\Delta_{\rm spec}=\mathbb E\|Y-\hat Y_s\|^2-\mathbb E\|Y-\hat Y_m\|^2$.
Expanding $Y-\hat Y_s=(Y-\hat Y_m)+(\hat Y_m-\hat Y_s)$ yields
\begin{equation}
\Delta_{\rm spec}
=\mathbb E\|\hat Y_s-\hat Y_m\|^2
+2\mathbb E\langle Y-\hat Y_m,\hat Y_m-\hat Y_s\rangle.
\label{eq:app-specificity-decomposition}
\end{equation}
The risk change includes both prediction sensitivity and its interaction
with the original residual. If the matched prediction is
$\mathbb E[Y\mid X,D_m]$ and the replacement adds no conditional target
information, so $\mathbb E[Y\mid X,D_m,D_s]=\mathbb E[Y\mid X,D_m]$,
conditioning makes the cross term vanish. Then specificity equals the
mean-squared prediction change. The general decomposition applies to any
trained predictor with finite risk.

Positive specificity establishes lower risk with the matched donor under
the specified replacement law. Near-zero specificity can arise from
cancellation of the two terms despite different predictions. The donor and
recipient protocols in RH20T, SpringWorld and Swimmer specify that law.
Zero-context replacement measures channel dependence, while wrong-donor
replacement measures content specificity. Absolute-risk comparisons and fresh-reader evaluations quantify task benefit.

\FloatBarrier
\subsection{Value: conditional risk and reader realization}
\label{app:utility-theory}
\label{app:cross-task-theory}

The same relation-conditioned interface supports numerical prediction and
categorical action learning, but their native losses measure different
quantities. We apply standard conditional-risk and information-theoretic
identities \citep{cover2006elements,gneiting2007proper} to separate available
task value from its realization by a trained reader.

\paragraph{Common setup.}
Fix the encoders, target construction, and pairing law
$\Pi_G(A,B)=P(B)Q_G(A\mid B)$. They induce a joint law of the permitted
recipient inputs $X=X_B$, donor code $D=D_A$, and target $Y=Y_B$.
Write $R_\ell(f)=\mathbb E\ell(f,Y)$. Risk is evaluated under this fixed
joint law, allowing statistical dependence between donor and recipient.
For Cross it is the recipient-conditioned risk in
Eq.~(\plaineqref{eq:cross-prediction}); the same analysis applies to a
native history-conditioned reader in an Align-only instance.

\paragraph{Proposition 4 (task value and reader error).}
Let $\Delta_\ell^\star$ be the optimal recipient-only risk minus the optimal
risk with $(X,D)$, allowing unrestricted measurable readers.

\emph{(a) Numerical targets.} For square-integrable $Y$ and squared loss,
the optimal readers are $m_0=\mathbb E[Y\mid X]$ and
$m_D=\mathbb E[Y\mid X,D]$, and
\begin{equation}
\Delta_{\rm sq}^\star
=\mathbb E\|Y-m_0\|^2-\mathbb E\|Y-m_D\|^2
=\mathbb E\|m_D-m_0\|^2\geq0.
\label{eq:app-bayes-task-value}
\end{equation}

\emph{(b) Categorical targets.} For $Y$ in a finite alphabet and log loss
$\ell(p,Y)=-\log p(Y)$, write $p_0=P(Y=\cdot\mid X)$ and
$p_D=P(Y=\cdot\mid X,D)$. These distributions attain the respective
optimal risks $H(Y\mid X)$ and $H(Y\mid X,D)$, so
\begin{equation}
\Delta_{\log}^\star=H(Y\mid X)-H(Y\mid X,D)
=I(Y;D\mid X)=\mathbb E\operatorname{KL}(p_D\|p_0)\geq0.
\label{eq:app-log-loss-task-value}
\end{equation}
Logarithms are natural. We assume conditional distributions exist;
$X$ and $D$ may be continuous.

For either loss, two learned readers $f_0$ and $f_D$ with finite risk obey
\begin{equation}
R_\ell(f_0)-R_\ell(f_D)
=\Delta_\ell^\star+\mathcal E_0^\ell-\mathcal E_D^\ell,
\label{eq:app-reader-excess-risk}
\end{equation}
where the excess risks are
\begin{equation*}
\begin{aligned}
\mathcal E_0^{\rm sq}&=\mathbb E\|f_0-m_0\|^2,&
\mathcal E_D^{\rm sq}&=\mathbb E\|f_D-m_D\|^2,\\
\mathcal E_0^{\log}&=\mathbb E\operatorname{KL}(p_0\|\widehat p_0),&
\mathcal E_D^{\log}&=\mathbb E\operatorname{KL}(p_D\|\widehat p_D).
\end{aligned}
\end{equation*}
Here the learned log-loss readers are probability vectors
$\widehat p_0,\widehat p_D$. The KL comparisons use the same conditioning
inputs; assigning zero probability to a positive-probability event is
excluded by the finite-risk assumption.

\paragraph{Proof.}
For squared loss, $r=Y-m_D$ has zero conditional mean given $(X,D)$ and is
orthogonal to $v=m_D-m_0$. Expanding $Y-m_0=r+v$ proves
Eq.~(\plaineqref{eq:app-bayes-task-value}). More generally, for a reader
$f$ with its corresponding conditional mean $m$,
\begin{equation*}
\mathbb E\|Y-f\|^2=\mathbb E\|Y-m\|^2+\mathbb E\|m-f\|^2,
\end{equation*}
by the same orthogonality, establishing optimality and the squared-loss
excess risks.

For log loss, adding and subtracting the true log probabilities gives
\begin{equation*}
\mathbb E[-\log\widehat p_D(Y)\mid X,D]
=-\sum_y p_D(y)\log\widehat p_D(y)
=H(p_D)+\operatorname{KL}(p_D\|\widehat p_D).
\end{equation*}
The nonnegativity of KL divergence shows that $p_D$ is optimal. Average
this equality over $(X,D)$ and repeat it conditioned on $X$ to obtain both
optimal risks and excess-risk terms. Finally, iterated expectation gives
\begin{equation*}
\mathbb E\operatorname{KL}(p_D\|p_0)
=\mathbb E\log\frac{p_D(Y)}{p_0(Y)}
=H(Y\mid X)-H(Y\mid X,D)=I(Y;D\mid X).
\end{equation*}
Subtracting the two reader-risk decompositions for each loss proves
Eq.~(\plaineqref{eq:app-reader-excess-risk}).
\unskip\nobreak\hfill\mbox{$\square$}

\paragraph{What this separates.}
A factor can be perfectly accessible in $D$ while irrelevant to $Y$:
if $F_1,F_2$ are independent, $D=F_1$, and $Y=F_2$ with no recipient
inputs, the code's task value is zero. Redundancy with $X$ can also remove
its incremental value. For squared loss, zero value means that the
conditional \emph{mean} is unchanged; for categorical log loss it means
that the conditional \emph{distribution} is unchanged. The criterion thus
depends on the chosen loss.
Equation~(\plaineqref{eq:app-reader-excess-risk}) further separates
attainable value from how well the particular readers realize it.
The PokeWorld frozen-source study estimates realized risk for its fitted
readers. Together with the accessibility probes, it separates readable
information from the benefit those readers obtain.

\paragraph{An explicit cold-start example.}
Let $\theta\sim\mathcal N(0,\sigma_\theta^2)$,
$X=\theta+\epsilon_x$, and $D=\theta+\epsilon_d$, with centered
Gaussian noises independent of each other and of $\theta$, and positive
prior and noise variances.
For $Y=a_h\theta+\epsilon_y$, assume centered target noise of finite
variance independent of $\theta,\epsilon_x,\epsilon_d$.
Completing the square in the Gaussian conditional
density gives recipient posterior variance
$v_x=(\sigma_\theta^{-2}+\sigma_x^{-2})^{-1}$ and, after observing $D$,
variance $(v_x^{-1}+\sigma_d^{-2})^{-1}$. Target noise cancels from the risk
difference, giving
\begin{equation}
\Delta_{\rm sq}^\star
=a_h^2\left[v_x-(v_x^{-1}+\sigma_d^{-2})^{-1}\right]
=a_h^2\frac{v_x^2}{v_x+\sigma_d^2}.
\label{eq:app-cold-start-benefit}
\end{equation}
Within this model, larger remaining recipient uncertainty $v_x$ makes the
same history more valuable. Averaging $R$ independent history measurements
replaces $\sigma_d^2$ by $\sigma_d^2/R$. Target sensitivity enters through
$a_h$; a longer horizon helps only insofar as it changes the target's
dependence on persistent information or the available evidence.
The model explains the role of recipient uncertainty in SpringWorld cold
start. A controlled change in recipient evidence holds population, target
and fitted route fixed; horizon effects additionally depend on $a_h$.

\paragraph{Connection to training and generalization beyond JEPA.}
For fixed encoders, targets, and pairing law, the isolated Cross objective
asks the reader for the recipient target's conditional mean under squared
loss, or its conditional distribution under categorical log loss.
Sharing an object, partner, or task constructs potentially informative pairs;
the donor still needs to add target-relevant information beyond $X$.
A numerical target can be a future state or a latent embedding held fixed
for this reference; a categorical target can be an observed action.
The complete \sprii{} objective fits these components jointly through a
shared predictor.

A training law and an evaluation law define separate risks; each identity
holds within its specified law and fixed representation. Action log loss
also differs from rollout return, which depends on the policy-induced state
distribution and environment response. Thus the generalization is a common
relation-conditioned learning interface, with native targets and losses
specifying what the reader learns to use. Native-task evaluation tests its
realized benefit.

\paragraph{Connection to the downstream intervention.}
The $M_1/M_2$ study changes the consuming interface while keeping the source
and prescribed query/base path fixed. The gain
$\Delta_{\rm realize}=\operatorname{Err}(M_1)-\operatorname{Err}(M_2)$
measures a difference in realized risk, with positive values favoring $M_2$
(Section~\ref{sec:value-results}).
Equation~(\plaineqref{eq:app-reader-excess-risk}) attributes this difference
to conditional value and reader error jointly. A positive adapter gain
demonstrates additional realization under the compared routes; the
unrestricted attainable value remains a separate quantity.

\FloatBarrier
\section{Closest Methodological Connections}
\label{app:extended-related-work}\label{app:nearest-methods}
The closest precedents already use related experience, separate context from
state, or test dependence on supplied context. Table~\ref{tab:nearest-methods}
locates the additional questions studied here.

\begin{table}[!htbp]
\centering
\small
\setlength{\tabcolsep}{5pt}
\renewcommand{\arraystretch}{1.16}
\caption{\textbf{Closest methodological connections.}}
\label{tab:nearest-methods}
\begin{tabular}{>{\raggedright\arraybackslash}p{0.15\linewidth}>{\raggedright\arraybackslash}p{0.36\linewidth}>{\raggedright\arraybackslash}p{0.40\linewidth}}
\toprule
\rowcolor{TableHeaderTint}
\textbf{Work} & \textbf{Relevant construction} & \textbf{Connection developed by \sprii{}}\\
\tableheadrule
FCRL & Same-function observation sets; contrastive encoding; new decoders on a
fixed encoder. & Relation training within predictive/action learners; selected
shared factors; separate measurement of code access and predictor reliance. \\\tableentryrule
NOD & A separate same-system trajectory supplies a low-dimensional code for
state evolution. & Embedding, outcome, and action targets; Align/Cross
components; controlled relation semantics and information conditions. \\\tableentryrule
RIA & Predictive interventions estimate relations used to organize dynamics
context. & Observed relations provide a controlled variable whose reliability
and semantics can be changed independently. \\\tableentryrule
GG-ODE & Initial-state/environment separation; same-environment contrastive
regularization. & An interface instantiated across base objectives, with
partial-factor sharing and task/use diagnostics. \\\tableentryrule
Policy embeddings & Cross-episode imitation and identity objectives;
downstream policies and wrong-embedding tests. & Relationship to physical and
JEPA persistence; history-conditioned action applications and controlled
information-use studies. \\
\bottomrule
\end{tabular}
\end{table}
\paragraph{Relations and predictive context.}
FCRL learns from same-function observation sets and transfers fixed encoders
\citep{gondal2021fcrl}. NOD conditions an operator on a separate same-system
trajectory, directly preceding the Cross-type construction \citep{chen2026nod}.
RIA uses an interventional prediction module to estimate whether two inferred
contexts come from the same environment, then regularizes their similarity
with a relational head. GG-ODE separates initial state from environmental context \citep{guo2022ria,huang2023ggode}.
Agent embeddings also use cross-episode imitation and incorrect-context tests
\citep{grover2018policy}; our Overcooked instance adds alignment to a
history-conditioned action learner \citep{laskin2023ad,jing2026icrl4aht}.
Here, controlled reliability and partial-factor relations make the selected
shared content an experimental variable, while Align/Cross controls separate
operations within each native predictive objective.

\paragraph{Context inference and test-time adaptation.}
CaDM learns dynamics context through forward/backward prediction
\citep{lee2020cadm}; DALI learns latent dynamics context for adaptation
\citep{dali2025}. Our D-Clean experiment adapts the context-learning component
to a common frozen-code reader and reports source-training budgets explicitly;
it does not rank the original control agents. CoDA and GEPS adapt physical
models with context codes \citep{kirchmeyer2022coda,geps2024}. Their completed
Burgers references use 50 support-code optimization steps per query, whereas
NOD and the SPRII endpoint use amortized code inference. The frozen-decoder
CoDA extension in Appendix~\ref{app:coda-sprii} asks the separate question of
whether relation supervision improves an amortized history encoder on this
host. Its four control arms distinguish encoder learning, relation assignments
and additional reconstruction computation.

\paragraph{Prediction, accessibility and reliance.}
\label{app:predictive-related}\label{app:adaptation-related}
Latent world models and joint-embedding predictors provide the predictive
setting \citep{hafner2019planet,hafner2020dreamer,assran2023ijepa,bardes2024vjepa}.
Physical-information studies distinguish input evidence from learned access
\citep{tan2026latent}, while paired-view learning constrains shared information
\citep{bardes2022vicreg,tian2020goodviews}.
Our grouping supervision supplies relation assignments between histories.
A frozen-code probe tests accessibility; a fitted reader tests realized task
value; substitution through fixed weights tests model reliance under the stated
donor law \citep{fisher2019modelreliance,molnar2023pfi,watson2021cpi}.
These distinctions motivate the separate Formation, Use and Value analyses.

\FloatBarrier
\FloatBarrier
\end{document}